%% file: 0_main.tex
\documentclass[runningheads,a4paper]{llncs}
\def\BibTeX{{\rm B\kern-.05em{\sc i\kern-.025em b}\kern-.08emT\kern-.1667em\lower.7ex\hbox{E}\kern-.125emX}}

\usepackage{geometry}
\usepackage{graphicx}
\usepackage{amsmath}
\usepackage{amssymb}
\usepackage{amsfonts}
\usepackage{mathrsfs}
\usepackage{tikz}
\usepackage{booktabs} % For formal tables
\usepackage[algo2e,ruled,linesnumbered]{algorithm2e} % For algorithms
\usepackage{multicol}
\usepackage{multirow}
\usepackage{subfig}
\usepackage{xcolor,colortbl}
\usepackage{hyperref}
\usepackage{makecell}
\usepackage{academicons}
\usepackage{dirtytalk}

\newcommand{\shap}{{\scshape shap}}
\newcommand{\deepshap}{{\scshape deep-shap}}
\newcommand{\gradshap}{{\scshape grad-shap}}
\newcommand{\lore}{{\scshape lore}}
\newcommand{\lime}{{\scshape lime}}
\newcommand{\dalex}{{\scshape dalex}}
\newcommand{\anchor}{{\scshape anchor}}
\newcommand{\maple}{{\scshape maple}}
\newcommand{\ebm}{{\scshape ebm}}
\newcommand{\cem}{{\scshape cem}}
\newcommand{\ceml}{{\scshape ceml}}

\newcommand{\face}{{\scshape face}}

\newcommand{\deeplift}{{\scshape deeplift}}
\newcommand{\elrp}{{\scshape $\epsilon$-lrp}}
\newcommand{\gradcam}{{\scshape gradcam}}
\newcommand{\gradcamplus}{{\scshape gradcam++}}
\newcommand{\rise}{{\scshape rise}}
\newcommand{\xrai}{{\scshape xrai}}
\newcommand{\smoothgrad}{{\scshape smoothgrad}}
\newcommand{\intgrad}{{\scshape intgrad}}
\newcommand{\tcav}{{\scshape tcav}}
\newcommand{\mmdcritic}{{\scshape mmd-critic}}
\newcommand{\ace}{{\scshape ace}}
\newcommand{\abele}{{\scshape abele}}

\newcommand{\adult}{\texttt{adult}}
\newcommand{\german}{\texttt{german}}
\newcommand{\mnist}{\texttt{mnist}}
\newcommand{\cifar}{\texttt{cifar}}
\newcommand{\imagenet}{\texttt{imagenet}}
\newcommand{\sst}{\texttt{sst}}
\newcommand{\imdb}{\texttt{imdb}}
\newcommand{\yelp}{\texttt{yelp}}
\newcommand{\norm}[1]{\left\lVert#1\right\rVert}
\newcommand\ministrut{\rule{0cm}{0.1cm}}

\begin{document}

\title{Benchmarking and Survey of Explanation Methods for Black Box Models}

\author{Francesco Bodria\inst{1} \and Fosca Giannotti\inst{2} \and Riccardo Guidotti\inst{3} \and Francesca Naretto\inst{1} \and Dino Pedreschi\inst{3} \and Salvatore Rinzivillo\inst{2}}

\authorrunning{F. Bodria, F. Giannotti, R. Guidotti, F. Naretto, D. Pedreschi}

\institute{
Scuola Normale Superiore, Pisa, Italy, \email{\{name.surname\}@sns.it} 
\and 
ISTI-CNR, Pisa, Italy, \email{\{name.surname\}@isti.cnr.it}
\and
Largo Bruno Pontecorvo, Pisa, Italy, \email{\{name.surname\}@unipi.it}}

\maketitle

\begin{abstract}
The widespread adoption of black-box models in Artificial Intelligence has enhanced the need for explanation methods to reveal how these obscure models reach specific decisions. Retrieving explanations is fundamental to unveil possible biases and to resolve practical or ethical issues. Nowadays, the literature is full of methods with different explanations. We provide a categorization of explanation methods based on the type of explanation returned. We present the most recent and widely used explainers, and we show a visual comparison among explanations and a quantitative benchmarking.
\end{abstract}


\input{1_introduction}
\input{2_related}

\input{3_classification}

\input{4_tabular}

\input{5_image}
\input{6_text}

\input{7_Toolboxes}
\input{8_conclusion}

\subsubsection*{Acknowledgements}
This work is partially supported by the European Community H2020 programme under the funding schemes: 
INFRAIA-1-2014-2015 Res. Infr. G.A. 871042 \emph{SoBigData++}, 
G.A. 952026 \emph{HumanE AI Net}, 
G.A. 825619 \emph{AI4EU}, G.A. 834756 \emph{XAI}. 

\bibliographystyle{abbrv}
%\bibliography{references.bib}
\bibliography{references.bib}

\end{document}

%% file: 1_introduction.tex
\section{Introduction}
\label{sec:introduction}
Today AI is one of the most important scientific and technological areas, with a tremendous socio-economic impact and a pervasive adoption in many fields of modern society.
The impressive performance of AI systems in prediction, recommendation, and decision making support is generally reached by adopting complex Machine Learning (ML) models that ``hide'' the logic of their internal processes. 
As a consequence, such models are often referred to as ``black-box models''~\cite{guidotti2018survey,freitas2014comprehensible,pasquale2015black}.
Examples of black-box models used within current AI systems include deep learning models and ensemble such as bagging and boosting models.
The high performance of such models in terms of accuracy has fostered the adoption of non-interpretable ML models even if the opaqueness of black-box models may hide potential issues inherited by training on biased or unfair data ~\cite{kurenkov2020lessons}.
Thus there is a substantial risk that relying on opaque models may lead to adopting decisions that we do not fully understand or, even worse, violate ethical principles.
Companies are increasingly embedding ML models in their AI products and applications, incurring a potential loss of safety and trust~\cite{chouldechova2017fair}.
These risks are particularly relevant in high-stakes decision making scenarios, such as medicine, finance, automation. 
In 2018, the European Parliament introduced in the GDPR\footnote{ \url{https://ec.europa.eu/justice/smedataprotect/}} a set of clauses for automated decision-making in terms of \textit{a right of explanation} for all individuals to obtain ``meaningful explanations of the logic involved'' when automated decision making takes place. 
Also, in 2019, the High-Level Expert Group on AI presented the ethics guidelines for trustworthy AI\footnote{\url{https://ec.europa.eu/digital-single-market/en/news/ethics-guidelines-trustworthy-ai}}.
Despite divergent opinions among legals regarding these clauses~\cite {goodman2016eu,wachter2017right,comande2017regulating}, everybody agrees that the need for the implementation of such a principle is urgent and that it is a huge open scientific challenge. 

As a reaction to these practical and theoretical ethical issues, in the last years, we have witnessed the rise of a plethora of explanation methods for black-box models~\cite{guidotti2018survey,adadi2018peeking,arrieta2020explainable} both from academia and from industries.
Thus, eXplainable Artificial Intelligence (XAI)~\cite{miller2019explanation} emerged as investigating methods to produce or complement AI to make accessible and interpretable the internal logic and the outcome of the model, making such process human understandable.

This work aims to provide a fresh account of the ideas and tools supported by the current explanation methods or explainers from the different explanations offered.\footnote{This work extends and complete ``A Survey Of Methods For Explaining Black-Box Models'' appeared in ACM computing surveys (CSUR), 51(5), 1-42~\cite{guidotti2018survey}.}. 
We categorize explanations w.r.t. the nature of the explanations providing a comprehensive ontology  of the explanation provided by available explainers taking into account the three most popular data formats: tabular data, images, and text.
We also report extensive examples of various explanations and qualitative and quantitative comparisons to assess the faithfulness, stability, robustness, and running time of the explainers.
Furthermore, we include a \textit{quantitative numerical comparison} of some of the explanation methods aimed at testing their faithfulness, stability, robustness, and running time.

\smallskip
The rest of the paper is organized as follows.
Section~\ref{sec:related} summarizes existing surveys on explainability in AI and interpretability in ML and highlights the differences between this work and previous ones.
Then, Section~\ref{sec:categorization} presents the proposed categorization based on the type of explanation returned by the explainer and on the data format under analysis.
Sections \ref{sec:tabular}, \ref{sec:image}, \ref{sec:text} present the details of the most recent and widely adopted explanation methods together with a qualitative and quantitative comparison.
Finally, Section~\ref{sec:conclusion} summarizes the crucial aspects that emerged from the analysis of the state of the art and future research directions.

%% file: 2_related.tex
\section{Related Works}
\label{sec:related}
The widespread need for XAI in the last years caused an explosion of interest in the design of explanation methods~\cite{goebel2018explainable}.
For instance, in the books~\cite{molnar2020interpretable,samek2019explainable} are presented in details the most well-known methodologies to make general machine learning models interpretable~\cite{molnar2020interpretable} and to explain the outcomes of deep neural networks~\cite{samek2019explainable}.

In~\cite{guidotti2018survey}, the classification is based on four categories of problems, and the explanation methods are classified according to the problem they are able to solve.
The first distinction is between \textit{explanation by design} (also named \textit{intrinsic} interpretability and \textit{black-box} explanation (also named \textit{post-hoc} interpretability~\cite{adadi2018peeking,murdoch2019definitions,carvalho2019machine}).
The second distinction in~\cite{guidotti2018survey}, further classify the black-box explanation problem into model explanation, outcome explanation and black-box inspection.
Model explanation, achieved by \textit{global} explainers~\cite{craven1996extracting}, aims at explaining the whole logic of a model.
Outcome explanation, achieved by \textit{local} explainers~\cite{ribeiro2016should,lundberg2017unified}, understand the reasons for a specific outcome.
Finally, the aim of black-box inspection, is to retrieve a visual representation for understanding how the black-box works. 
Another crucial distinction highlighted in~\cite{martens2007comprehensible,guidotti2018survey,adadi2018peeking,dovsilovic2018explainable,carvalho2019machine} is between \textit{model-specific} and \textit{model-agnostic} explanation methods.
This classification depends on whether the technique adopted to explain can work only on a specific black-box model or can be adopted on any black-box. 

In~\cite{gilpin2018explaining}, the focus is to propose a unified taxonomy to classify the existing literature. 
The following key terms are defined: \textit{explanation}, \textit{interpretability} and \textit{explainability}.
An explanation answers a ``why question'' justifying an event. 
Interpretability consists of describing the internals of a system in a way that is understandable to humans. 
A system is called interpretable if it produces descriptions that are simple enough for a person to understand using a vocabulary that is meaningful to the user.
An alternative, but similar, classification of definitions is presented in~\cite{arrieta2020explainable}, with a specific taxonomy for explainers of deep learning models.
The leading concept of the classification is Responsible Artificial Intelligence, i.e.,~a methodology for the large-scale implementation of AI methods in real organizations with fairness, model explainability, and accountability at its core. 
Similarly to~\cite{guidotti2018survey}, in~\cite{arrieta2020explainable} the term interpretability (or transparency) is used to refer to a passive characteristic of a model that makes sense for a human observer.
On the other hand, explainability is an active characteristic of a model, denoting any action taken with the intent of clarifying or detailing its internal functions.
Further taxonomies and definitions are presented in~\cite{murdoch2019definitions,carvalho2019machine}.
Another branch of the literature review is focusing on the quantitative and qualitative evaluation of explanation methods~\cite{samek2019explainable,carvalho2019machine}.
Finally, we highlight that the literature reviews related to explainability are focused not just on ML and AI but also on social studies~\cite{miller2019explanation,byrne2019counterfactuals}, recommendation systems~\cite{zhang2018explainable}, model-agents\cite{anjomshoae2019explainable}, and domain-specific applications such as health and medicine~\cite{tjoa2019survey}.

In this survey we decided to rewrite the taxonomy proposed in~\cite{guidotti2018survey} but from a data type perspective. In light of the works mentioned above, we believe that an updated systematic categorization of explanation methods based on the type of explanation returned and comparing the explanations is still missing in the literature.

%% file: 3_classification.tex
\begin{table}[t]
    \centering
    \caption{Examples of explanations divided for different data type and explanation}
    \label{tab:taxonomy_examples}
    \footnotesize
    \renewcommand{\arraystretch}{1.5}
    \begin{tabular}{|>{\centering}m{0.31\linewidth}|>{\centering}m{0.31\linewidth}|>{\centering}m{0.31\linewidth}|}
         \hline
         \rowcolor{gray!15} TABULAR & IMAGE & TEXT \tabularnewline
         \hline
         \begin{minipage}{0.31\textwidth}
             \centering
             \textbf{Rule-Based (RB)} \\ 
             A set of premises that the record must satisfy in order to meet the rule's consequence.\\
             \begin{minipage}[t][1cm][t]{.78\linewidth}\em
             $r$ =  \text{Education} $\leq$ \text{College} $\rightarrow$ $\leq 50k$
             \end{minipage}
             %$r$ \ =&  \{\text{Education} $\leq$ \text{College} $\rightarrow$ $\leq 50k$ \\
         \end{minipage} 
         & 
         \centering
         \textbf{Saliency Maps (SM)} \\
         A map which highlight the contribution of each pixel at the prediction. \\
         \includegraphics[trim={0cm 0 14cm 5cm},clip,width=.6\linewidth]{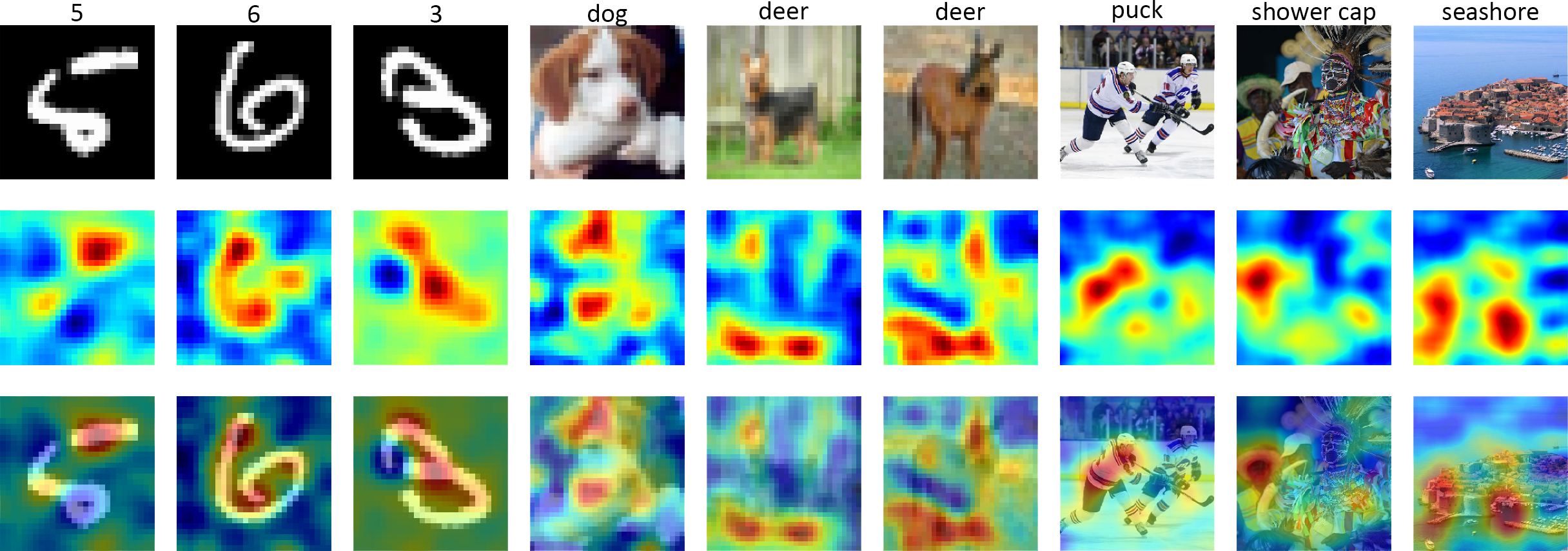} 
         & 
         \textbf{Sentence Highlighting (SH)} \\
         A map which highlight the contribution of each word at the prediction.
         \includegraphics[trim={0 2.2cm 45cm 0.1cm},clip,width=.8\linewidth]{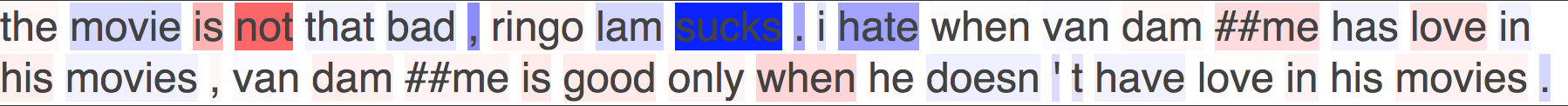} 
         \tabularnewline
         \hline
         \textbf{Feature Importance (FI)} \\
         A vector containing a value for each feature. Each value indicates the importance of the feature for the classification. \\
         \includegraphics[trim={24cm 3cm 2cm 1.65cm},clip,width=0.50\linewidth]{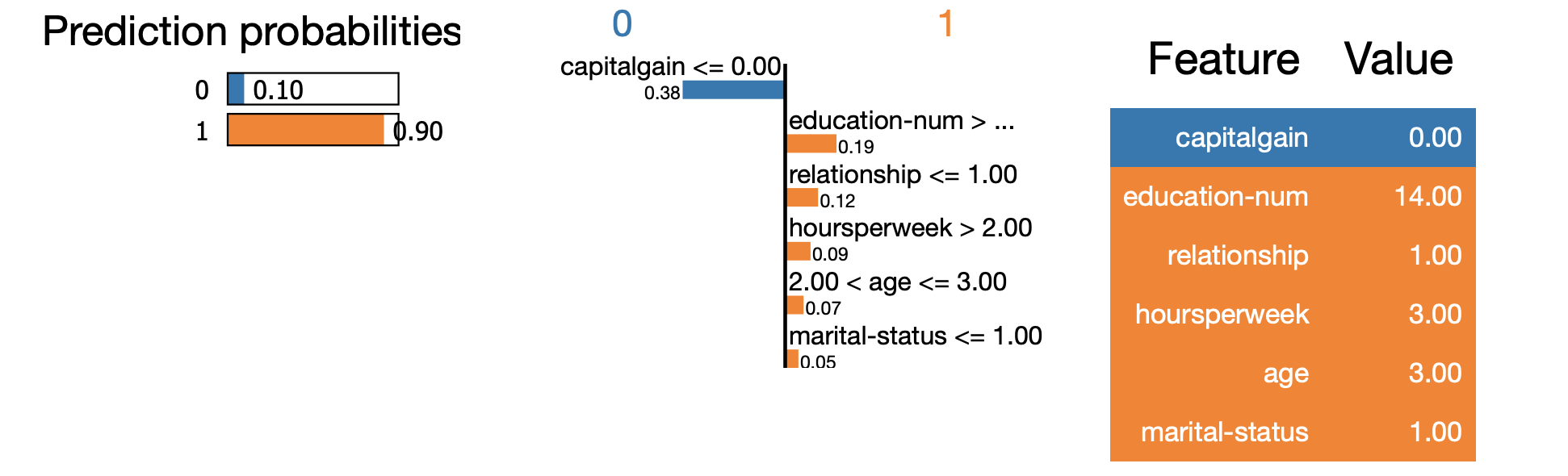} 
         & 
         \textbf{Concept Attribution (CA)} \\
         Compute attribution to a target “concept” given by the user. For example, how sensitive is the output (a prediction of zebra) to a concept (the presence of stripes)? \\
         \ministrut \\
         \includegraphics[width=\linewidth]{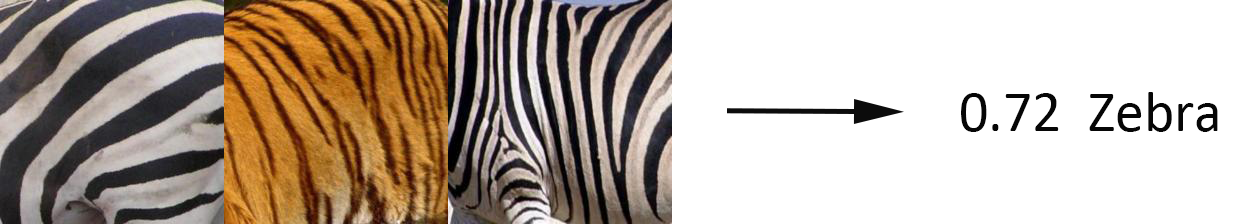}
         & 
         \textbf{Attention Based (AB)} \\
         This type of explanation gives a matrix of scores which reveal how the word in the sentence are related to each other.\\
         \ministrut \\
         \includegraphics[trim={0 0 3cm 0},clip, width=.8\linewidth]{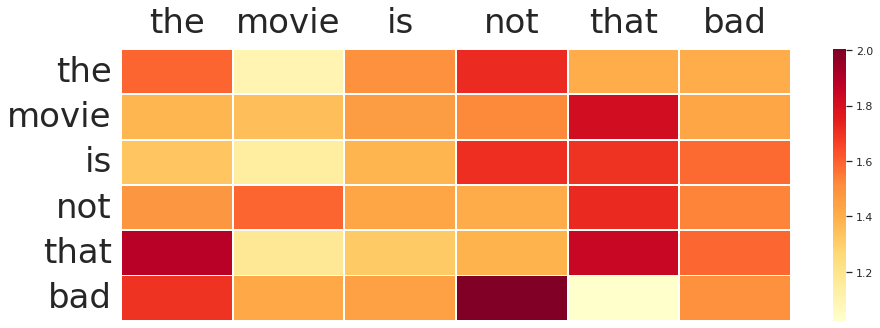}
         \tabularnewline
         \hline
         \multicolumn{3}{|c|}{
         \begin{minipage}{.93\linewidth}
             \centering
             \ministrut \\
             \textbf{Prototypes (PR)} \\
             The user is provided with a series of examples that characterize a class of the black box \\
             \begin{minipage}{.33\linewidth}\centering \em
                 $p$ = \text{Age} $\in [ 35, 60 ]$, \text{Education} $\in [\text{College}, \text{Master}] \rightarrow$``$\geq 50k$'' 
             \end{minipage}
             \begin{minipage}{.33\linewidth}
                \centering
                \ministrut \\
                \begin{minipage}{.1\linewidth}
                    \centering
                    $p = $
                \end{minipage}
                \begin{minipage}{.5\linewidth}
                    \centering
                    \includegraphics[width=.8\linewidth]{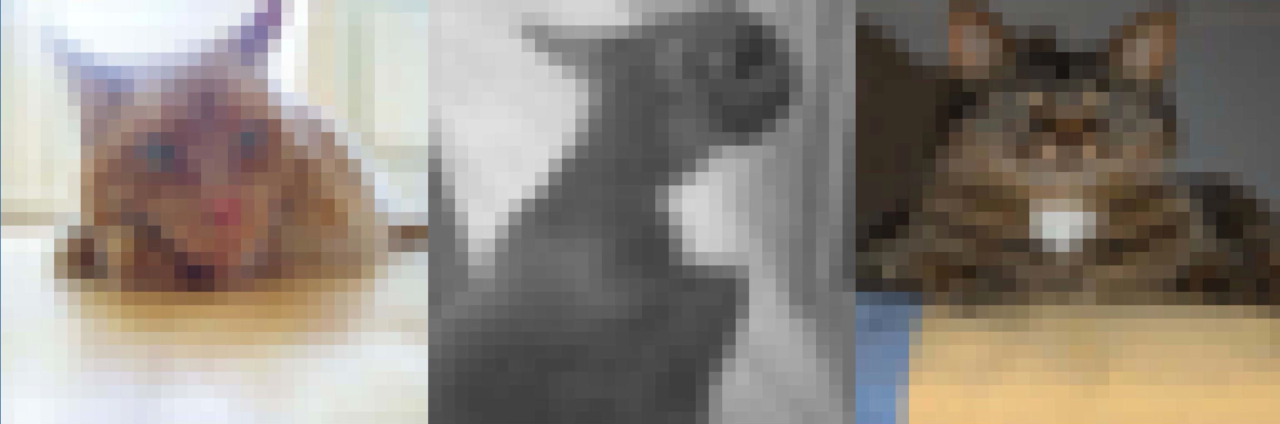}
                \end{minipage}
                \begin{minipage}{.2\linewidth}
                    \centering
                    $\rightarrow$ ``cat''   
                \end{minipage} \\
                \ministrut
             \end{minipage}
             \begin{minipage}{.3\linewidth}
             \centering
                $p$ = ``... not bad ...'' $\rightarrow$ ``positive''
             \end{minipage} \\
             \ministrut
         \end{minipage}}
         \tabularnewline
         \hline
         \multicolumn{3}{|c|}{
         \begin{minipage}{.93\linewidth}
             \centering
             \ministrut \\
             \textbf{Counterfactuals (CF)} \\
              The user is provided with a series of examples similar to the input query but with different class prediction \\
              \ministrut \\
             \begin{minipage}{.33\linewidth}\centering \em
                 $q =  \text{Education} \leq \text{College} \rightarrow$ ``$\leq 50k$'' \\
                 $c =  \text{Education} \geq \text{Master}  \rightarrow$ ``$\geq 50k$'' 
             \end{minipage} 
             \begin{minipage}{.33\linewidth}
                \centering
                \begin{minipage}{.1\linewidth}
                   \centering
                   $q = $
                \end{minipage}
                \begin{minipage}{.23\linewidth}
                   \centering
                   \includegraphics[trim={0 0 3.24cm 0},clip,width=.8\linewidth]{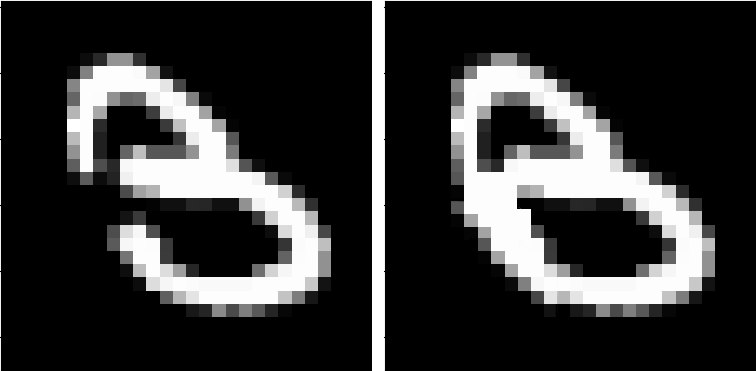}
                \end{minipage}
                \begin{minipage}{.23\linewidth}
                    \centering
                    $\rightarrow$``$3$''$\,\,c = $
                \end{minipage}
                \begin{minipage}{.23\linewidth}
                   \centering
                   \includegraphics[trim={3.24cm 0 0 0},clip,width=.8\linewidth]{Figures/counter_table.png}
                \end{minipage} 
                \begin{minipage}{.1\linewidth}
                   \centering
                   $\rightarrow$``$8$''
                \end{minipage}\\
                \ministrut
             \end{minipage}
             \begin{minipage}{.33\linewidth}
                \centering
                $q =  \text{The movie is not that bad} \rightarrow$``positive'' \\
                $c =  \text{The movie is that bad} \rightarrow$``negative'' \\
             \end{minipage}
         \end{minipage}} 
         \tabularnewline
         \hline
    \end{tabular}
    
\end{table}

\section{Explanation-Based Categorization of Explainers and Evaluation Measures}
\label{sec:categorization}
This paper aims to categorize explanation methods concerning the type of explanation returned and present the most widely adopted quantitative evaluation measures to validate explanations under different aspects and benchmark the explainers adopting these measures.
The objective is to provide to the reader a guide to map a black-box model to a set of compatible explanation methods.
Furthermore, we systematically present a qualitative comparison of the explanations that also help understand how to read these explanations returned by the different methods\footnote{All the experiments in the next sections are performed on a server with GPU: 1xTesla K80 , compute 3.7, having 2496 CUDA cores , 12GB GDDR5 VRAM, CPU: 1xsingle core hyper threaded Xeon Processors @2.3Ghz i.e (1 core, 2 threads) with 16 GB of RAM, or on a server: CPU: 16x Intel(R) Xeon(R) Gold 5120 CPU @ 2.20GHz (64 bits), 63 gb RAM. The code for reproducing the results is available \url{https://github.com/kdd-lab/XAI-Survey}.}.

\subsection{Categorization of Type of Explanations}
\label{sec:cate}
In this survey, we present explanations and explanation methods acting on the three principal data types recognized in the literature: \textit{tabular} data, \textit{images} and \textit{text}~\cite{guidotti2018survey}.
In particular, for every of these data types, we have distinguished different types of explanations illustrated in Table~\ref{tab:taxonomy_examples}.
A Table appearing at the beginning of each subsequent Section summarizes the explanation methods by grouping them accordingly to the classification illustrated in Table~\ref{tab:taxonomy_examples}.
Besides, in every section we present the meaning of each type of explanation. 
The acronyms reported in capital letters in Table~\ref{tab:taxonomy_examples}, in this section and in the following are used in the remainder of the work to quickly categorize the various explanations and explanation methods. % e per risparmiare spazio LOL
We highlight that the nature of this work is tied to test the available libraries and toolkits for XAI.
Therefore, the presentation of the existing methods is focused on the most recent works (specifically from 2018 to the date of writing) and to those papers providing a usable implementation that is nowadays widely adopted.

\subsection{Existing XAI Taxonomy for Explanation Methods}
\label{sec:comparison_theory}
\begin{figure}[t]
    \centering
    \includegraphics[width=0.6\linewidth]{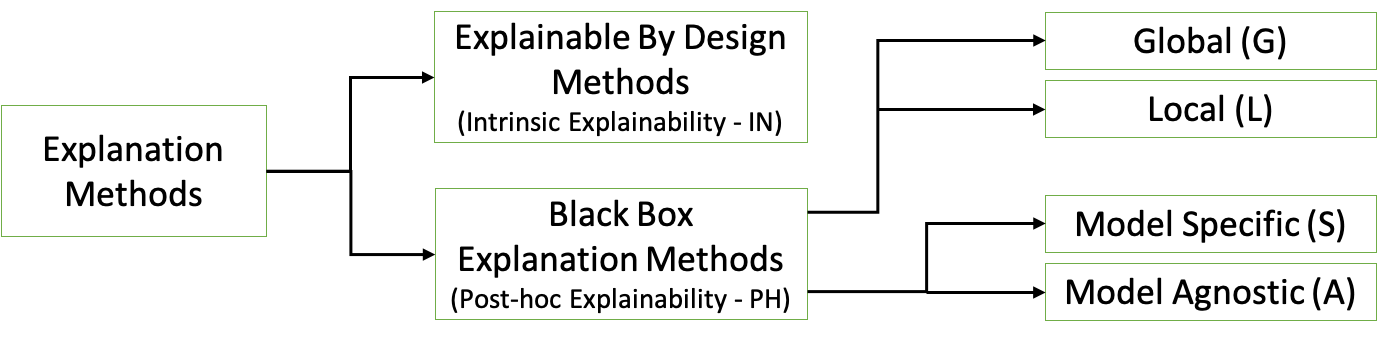}
    \caption{Existing taxonomy for the classification of explanation methods.}
    \label{fig:taxonomy}
\end{figure}

In this section, we synthetically recall the existing taxonomy and classification of XAI methods present in the literature~\cite{guidotti2018survey,adadi2018peeking,gilpin2018explaining,arrieta2020explainable,samek2019explainable,carvalho2019machine} to allow the reader to complete the proposed explanation-based categorization of explanation methods.
We summarize the fundamental distinctions adopted to annotate the methods in Figure~\ref{fig:taxonomy}.

The first distinction separates explainable by design methods from black-box explanation methods:
\begin{itemize}
    \item \textit{\textbf{Explainable by design methods}} are \textit{INtrinsically} \textit{(IN)} explainable methods that returns a decision, and the reasons for the decision are directly accessible because the model is transparent.
    \item \textit{\textbf{Black-box explanation}} are \textit{Post-Hoc} \textit{(PH)} explanation methods that provides explanations for a non interpretable model that takes decisions.
\end{itemize}

The second differentiation distinguishes post-hoc explanation methods in global and local:
\begin{itemize}
    \item \textit{\textbf{Global (G)}} explanation methods aim at explaining the overall logic of a black-box model. Therefore the explanation returned is a global, complete explanation valid for any instance;
    \item \textit{\textbf{Local (L)}} explainers aim at explaining the reasons for the decision of a black-box model for a specific instance. 
\end{itemize}

The third distinction categorizes the methods into model-agnostic and model-specific:
\begin{itemize}
    \item \textit{\textbf{Model-Agnostic (A)}} explanation methods can be used to interpret \textit{any type} of black-box model;
    \item \textit{\textbf{Model-Specific (S)}} explanation methods can be used to interpret \textit{only a specific type} of black-box model.
\end{itemize}

To provide to the reader a self-contained review of XAI methods, we complete this section by rephrasing succinctly and unambiguously the definitions of explanation, interpretability, transparency, and complexity: 
\begin{itemize}
    \item \textit{Explanation}~\cite{arrieta2020explainable,guidotti2018survey} is an \textit{interface} between humans and an AI decision-maker that is both comprehensible to humans and an accurate proxy of the AI. Consequently, explainability is the ability to provide a \textit{valid} explanation.
    \item \textit{Interpretability}\cite{guidotti2018survey}, or comprehensibility~\cite{gleicher2016framework}, is the ability to explain or provide the meaning in understandable terms a human.
    Interpretability and comprehensibility are normally tied to the evaluation of the model complexity.
    \item \textit{Transparency}~\cite{arrieta2020explainable}, or equivalently understandability or intelligibility, is the capacity of a model of being interpretable itself. Thus, the model allows a human to understand its functioning without explaining its internal structure or the algorithmic means by which the model processes data internally.
    \item \textit{Complexity}~\cite{doshi2017towards} is the degree of effort required by a user to comprehend an explanation. The complexity can consider the user background or eventual time limitation necessary for the understanding. 
\end{itemize}

\subsection{Evaluation Measures}
\label{evaluation-metrics}
The validity and the utility of explanations methods should be evaluated in terms of goodness, usefulness, and satisfaction of explanations.
In the following, we describe a selection of established methodologies for the evaluation of explanation methods both from the qualitative and quantitative point of view.
Moreover, depending on the kind of explainers under analysis, additional evaluation criteria may be used. 
\textit{Qualitative evaluation} is important to understand the actual usability of explanations from the point of view of the end-user: they satisfy human curiosity, find meanings, safety, social acceptance and trust.
In~\cite{doshi2017towards} is proposed a systematization of evaluation criteria into three major categories:
\begin{itemize}
    \item[1.] \textbf{Functionally-grounded} metrics aim to evaluate the interpretability by exploiting some formal definitions that are used as proxies. They do not require humans for validation. The challenge is to define the proxy to employ, depending on the context. As an example, we can validate the interpretability of a model by showing the improvements w.r.t. to another model already proven to be interpretable by human-based experiments.
    \item[2.] \textbf{Application-grounded} evaluation methods require human experts able to validate the specific task and explanation under analysis~\cite{Williams2016axis,suissa2016automatic}. They are usually employed in specific settings. For example, if the model is an assistant in the decision making process of doctors, the validation is done by the doctors.
    \item[3.] \textbf{Human-grounded} metrics evaluate the explanations through humans who are not experts. The goal is to measure the overall understandability of the explanation in simplified tasks~\cite{lakkaraju2016interpretable,kim2015inferring}. This validation is most appropriate for general testing notions of the quality of an explanation.
\end{itemize}
Moreover, in~\cite{doshi2017towards,doshi2018considerations} are considered several other aspects: the form of the explanation; the number of elements the explanation contains; the compositionality of the explanation, such as the ordering of FI values; the monotonicity between the different parts of the explanation; uncertainty and stochasticity, which take into account how the explanation was generated, such as the presence of random generation or sampling.

In \textit{quantitative evaluation}, the evaluation focuses on the performance of the explainer and how close the explanation method $f$ is to the black-box model $b$.
Concerning quantitative evaluation we can consider two different types of \textit{criteria}:
\begin{itemize}
    \item[1.] \textbf{Completeness w.r.t. the black-box model}. The metrics aim at evaluating how closely $f$ approximates $b$.
    \item[2.] \textbf{Completeness w.r.t. to specific task}. The evaluation criteria are tailored for a particular task or behavior.
\end{itemize}
In the first criterion, we group the metrics that are often used in the literature~\cite{ribeiro2016should,ribeiro2018anchors,Guidotti2018LocalRE,sundararajan2017axiomatic}.
One of the metric most used in this setting is the \textit{fidelity} that aims to evaluate how good is $f$ at mimicking the black-box decisions. 
There are different specializations of fidelity, depending on the type of explanator under analysis~\cite{Guidotti2018LocalRE}. For example, in methods where there is a creation of a surrogate model $g$ to mimic $b$, fidelity compares the prediction of $b$ and $g$ on the instances $Z$ used to train $g$. 

Another measure of completeness w.r.t.~$b$ is the \textit{stability}, which aims at validating how consistent the explanations are for similar records. 
The higher the value, the better is the model to present similar explanations for similar inputs. Stability can be evaluated by exploiting the \textit{Lipschitz constant}~\cite{alvarez2018towards} as $ L_x = \text{max} \frac{\norm{e_{x} - e_{x'}}}{\norm{x - x'}} , \forall x' \in \mathcal{N}_x$
where $x$ is the explained instance, $e_x$ the explanation and $\mathcal{N}_x$ is a neighborhood of instances $x'$ similar to $x$.

Besides the synthetic ground truth experimentation proposed in~\cite{guidotti2020evaluating}, a strategy to validate the correctness of the explanation $e = f(b,x)$ is to remove the features that the explanation method $f$ found important and see how the performance of $b$ degrades. 
These metrics are called \textit{deletion} and \textit{insertion}~\cite{petsiuk2018rise}. 
The intuition behind deletion is that removing the ``cause'' will force the black-box to change its decision. 
Among the deletion methods, there is the \textit{faithfulness}~\cite{alvarez2018towards} which is tailored for FI explainers. 
It aims to validate if the relevance scores indicate true importance: we expect higher importance values for attributes that greatly influence the final prediction\footnote{An implementation of the faithfulness is available in {\scshape aix360}, presented in Section~\ref{expl-tools}}.
Given a black-box $b$ and the feature importance $e$ extracted from an importance-based explanator $f$, the faithfulness method incrementally removes each of the attributes deemed important by $f$.
At each removal, the effect on the performance of $b$ is evaluated. 
These values are then employed to compute the overall correlation between feature importance and model performance. 
This metrics corresponds to a value between $-1$ and $1$: the higher the value, the better the faithfulness of the explanation. 
In general, a sharp drop and a low area under the probability curve mean a good explanation. 
On the other hand, the insertion metric takes a complementary approach. 
\textit{monotonicity} is an implementation of an insertion method: it evaluates the effect of $b$ by incrementally adding each attribute in order of increasing importance. 
In this case, we expect that the black-box performance increases by adding more and more features, thereby resulting in monotonically increasing model performance. 
Finally, other standard metrics, such as \textit{accuracy}, \textit{precision} and \textit{recall}, are often evaluated to test the performance of the explanation methods.
The \textit{running time} is also an important evaluation.

%% file: 4_tabular.tex
\begin{table}[t]
    \small
    \caption{Summary of methods for explaining black-boxes for tabular data. The methods are sorted by explanation type: Features Importance (FI), Rule-Based (RB), Counterfactuals (CF), Prototypes (PR), and Decision Tree (DT). For every method, there is a data type on which it is possible to apply it: only on tabular (TAB) or any data (ANY). If it is an Intrinsic Model (IN) or a Post-Hoc one (PH), a local method (L) or a global one (G), and finally if it is model agnostic (A) or model-specific (S).}
    \label{tab:tabular_table}
    \centering
    \setlength{\tabcolsep}{1mm}
    %\rowcolors{1}{white}{light-gray}
    \begin{tabular}{cccccccccc}
    \hline
    \rotatebox[origin=c]{0}{\textbf{Type}} & \rotatebox[origin=c]{0}{\textbf{Name}}       & \rotatebox[origin=c]{0}{\textbf{Ref.}}  & \rotatebox[origin=c]{0}{\textbf{Authors}} & \rotatebox[origin=c]{0}{\textbf{Year}} & \rotatebox[origin=c]{0}{\textbf{Data Type}}  & \rotatebox[origin=c]{0}{\textbf{IN/PH}} & \rotatebox[origin=c]{0}{\textbf{G/L}} & \rotatebox[origin=c]{0}{\textbf{A/S}}  & \rotatebox[origin=c]{0}{\textbf{Code}}  \\ 
    
    \hline
     \cellcolor{white}
     &  \shap{} 		        & \cite{lundberg2017unified} 		& Lundberg et al.   	    & 2007  & ANY & PH & G/L & A & \href{https://github.com/slundberg/shap}{link} \\
     \rowcolor{gray!15} 
     \cellcolor{white}
     & \lime{}		            & \cite{ribeiro2016should} 		    & Ribeiro et al. 	        & 2016  & ANY & PH & L & A  & \href{https://github.com/marcotcr/lime}{link} \\
     & {\scshape lrp}           & \cite{bach2015pixel} 	            & Bach et al.               & 2015  & ANY & PH & L & A & \href{https://github.com/sebastian-lapuschkin/lrp_toolbox}{link} \\
     \rowcolor{gray!15} 
     \cellcolor{white}
     &  {\scshape dalex} 	    & \cite{biecekexplanatory}	& Biecek et al.  	    & 2020  & ANY & PH & L/G & A & \href{https://github.com/ModelOriented/DALEX}{link}\\
     \cellcolor{white}
     & {\scshape nam}           & \cite{agarwal2020neural}          & Agarwal et al.            & 2020  & TAB & PH & L & S & \href{https://github.com/nickfrosst/neural_additive_models}{link}\\
     \rowcolor{gray!15}
     \cellcolor{white}
     & {\scshape ciu}           & \cite{anjomshoae2020py}           & Anjomshoae et al.         & 2020  & TAB & PH & L & A & \href{https://github.com/TimKam/py-ciu}{link}\\
     \multirow{-10}{*}{FI} 
     \cellcolor{white} 
     & \maple{} 	            & \cite{plumb2018model} 		    & Plumb et al. 		        & 2018  & TAB & PH/IN & L & A & \href{https://github.com/GDPlumb/MAPLE}{link} \\
     \rowcolor{gray!15}       
     \cellcolor{gray!15}
     &  \anchor{}             	& \cite{ribeiro2018anchors}         & Ribeiro et al.    	    & 2018  & TAB & PH & L/G & A & \href{https://github.com/marcotcr/anchor}{link} \\
     \cellcolor{gray!15}
     & \lore{}                  & \cite{Guidotti2018LocalRE}		& Guidotti et al. 	        & 2018  & TAB & PH & L & A & \href{https://github.com/riccotti/LORE}{link} \\
     \rowcolor{gray!15} 
     \cellcolor{gray!15}
     &  {\scshape slipper}      & \cite{cohen1999simple}		    & Cohen et al.      	    & 1999  & TAB & IN & L & S & \href{https://github.com/scikit-learn-contrib/skope-rules}{link} \\
     \cellcolor{gray!15} 
     \cellcolor{gray!15}
     & {\scshape lri}		    & \cite{weiss2000lightweight}		& Weiss et al.      	    & 2000  & TAB & IN & L & S & -\\
     \rowcolor{gray!15} 
     \cellcolor{gray!15}
     & {\scshape mlrule}	    & \cite{dembczynski2008maximum}		& Domingos et al.   	    & 2008  & TAB & IN & G/L & S & \href{https://github.com/scikit-learn-contrib/skope-rules}{link}     \\
     \cellcolor{gray!15}
     & {\scshape rulefit} 	    & \cite{Friedman2008PREDICTIVELV}   & Friedman et al.    	    & 2008  & TAB & IN & G/L & S & \href{https://github.com/scikit-learn-contrib/skope-rules}{link} \\
     \rowcolor{gray!15} 
     \cellcolor{gray!15}
     & {\scshape scalable-brl}  & \cite{yang2017scalable}           & Yang et al.               & 2017  & TAB & IN & G/L & A & - \\
     \cellcolor{gray!15}
     & {\scshape rulematrix}    &  \cite{ming2018rulematrix}        & Ming et al.               & 2018  & TAB & PH & G/L & A & \href{https://github.com/rulematrix/rule-matrix-py}{link} \\
     \rowcolor{gray!15} 
     \cellcolor{gray!15}
     & {\scshape ids} 			& \cite{lakkaraju2016interpretable} & Lakkaraju et al.  	    & 2016  & TAB & IN & G/L & S & \href{https://github.com/jirifilip/pyIDS}{link} \\
     \cellcolor{gray!15}& {\scshape trepan}         & \cite{craven1996extracting}       & Craven et al.     & 1996  & TAB & PH & G & S & \href{https://github.com/abarthakur/trepan_python}{link} \\
    \rowcolor{gray!15}\cellcolor{gray!15}
    & {\scshape dectext}        & \cite{boz2002extracting}     	    & Boz et al.     	        & 2002  & TAB & PH & G & S & -\\
    \cellcolor{gray!15} 
     & {\scshape msft}              & \cite{chipman1998making}     	    & Chipman et al.     	    & 1998  & TAB & PH & G & S & -\\
    \rowcolor{gray!15}\cellcolor{gray!15} 
    & {\scshape cmm}            & \cite{domingos1998knowledge}     	& Domingos et al. 	        & 1998  & TAB & PH & G & S & -\\
    \cellcolor{gray!15}
     & {\scshape sta}           & \cite{zhou2016interpreting}    	& Zhou et al.     	        & 2016  & TAB & PH & G & S & -\\
    \rowcolor{gray!15} 
     & {\scshape skoperule}           & \cite{Friedman2008PREDICTIVELV}   	& Gardin et al.    & 2020  & TAB & PH & L/G & A & \href{https://github.com/scikit-learn-contrib/skope-rules}{link}\\
     \cellcolor{gray!15} \multirow{-16}{*}{RB}
     & {\scshape glocalx} & \cite{setzu2019global} & Setzu et al. & 2019 & TAB & PH & L/G & A & \href{https://github.com/msetzu/glocalx}{link} \\  
     \rowcolor{gray!15}\cellcolor{white} 
     & \mmdcritic{}             & \cite{mmdcritic} 			        & Kim et al. 		        & 2016  & ANY & IN & G & S & \href{https://github.com/BeenKim/MMD-critic}{link} \\
     \cellcolor{white} 
     &  {\scshape protodash}    & \cite{gurumoorthy2019efficient}   & Gurumoorthy et al.        & 2019  & TAB & IN & G & A & \href{https://github.com/Trusted-AI/AIX360}{link}\\
     \rowcolor{gray!15} \cellcolor{white}
     &   {\scshape tsp}	        & \cite{tan2020tree}                & Tan et al.     	        & 2020  & TAB & PH & L & S & -\\
    \multirow{-4}{*}{PR}  
    \cellcolor{white}
    &     {\scshape ps}	        & \cite{bien2011prototype}   	    & Bien et al.     	        & 2011  & TAB & IN & G/L & S & -\\
    \rowcolor{gray!15}
    & {\scshape cem}            & \cite{dhurandhar2018explanations} & Dhurandhar et al.         & 2018  & ANY & PH & L & S & \href{https://github.com/IBM/Contrastive-Explanation-Method}{link} \\
    \cellcolor{gray!15}
    & {\scshape dice}           & \cite{mothilal2020explaining} 	& Mothilal et al.  	        & 2020  & ANY & PH & L & A & \href{https://github.com/interpretml/DiCE}{link} \\
    \rowcolor{gray!15}

    & {\scshape face}           & \cite{poyiadzi2020face} 	        & Poyiadzi et al.  	        & 2020  & ANY & PH & L & A & -\\
    \multirow{-4}{*}{CF}  \cellcolor{gray!15}
    & {\scshape cfx} & \cite{albini2020relation} & Albini et al. & 2020 & TAB & PH & L & IN & - \\
    \hline
    \end{tabular}%
   % \end{varwidth}
\end{table}

\section{Explanations for Tabular Data}
\label{sec:tabular}
In this Section we present a selection of approaches for explaining decision systems acting on tabular data. 
In particular, we present the following types of explanations based on: \textit{Features Importance} (FI,  Section~\ref{sec:feat_imp_tabular}), \textit{Rules} (RB, Section~\ref{sec:rules}), \textit{Prototype} (PR) and \textit{Counterfactual} (CF) (Section~\ref{sec:proto_tabular}). 
Table~\ref{tab:tabular_table} summarizes and categorizes the explainers.
After the presentation of the explanation methods, we report experiments obtained from the application of them on two datasets\footnote{\adult{}: \url{https://archive.ics.uci.edu/ml/datasets/adult}, \german{}: \url{https://archive.ics.uci.edu/ml/datasets/statlog+(german+credit+data)}}: \adult{} and \german{}.
We trained the following ML models: Logistic Regression (LG), XGBoost (XGB), and Catboost (CAT).

\subsection{Feature Importance}
\label{sec:feat_imp_tabular}
Feature importance is one of the most popular types of explanation returned by local explanation methods. 
The explainer assigns to each feature an importance value which represents how much that particular feature was important for the prediction under analysis. 
Formally, given a record $x$, an explainer $f(\cdot)$ models a feature importance explanation as a vector $e = \{ e_1, e_2, \dots,$ $e_m \}$, in which the value $e_i \in e$ is the importance of the $i^{\mathit{th}}$ feature for the decision made by the black-box model $b(x)$. 
For understanding the contribution of each feature, the sign and the magnitude of each value $e_i$ are considered. 
W.r.t.~the sign, if $e_i < 0$, it means that feature contributes negatively for the outcome $y$; otherwise, if $e_i > 0$, the feature contributes positively. 
The magnitude, instead, represents how great the contribution of the feature is to the final prediction $y$. 
In particular, the greater the value of $|e_i|$, the greater its contribution. 
Hence, when $e_i = 0$ it means that the $i^{\mathit{th}}$ feature is showing no contribution for the output decision. 
An example of a feature based explanation is $e = \{\mathit{age} = 0.8, \mathit{income} = 0.0, \mathit{education} = -0.2\}, y = \mathit{deny}$. In this case, \textit{age} is the most important feature for the decision $\mathit{deny}$, \textit{income} is not affecting the outcome and \textit{education} has a small negative contribution.

\begin{figure}[t]
    \centering
    \includegraphics[width=\linewidth]{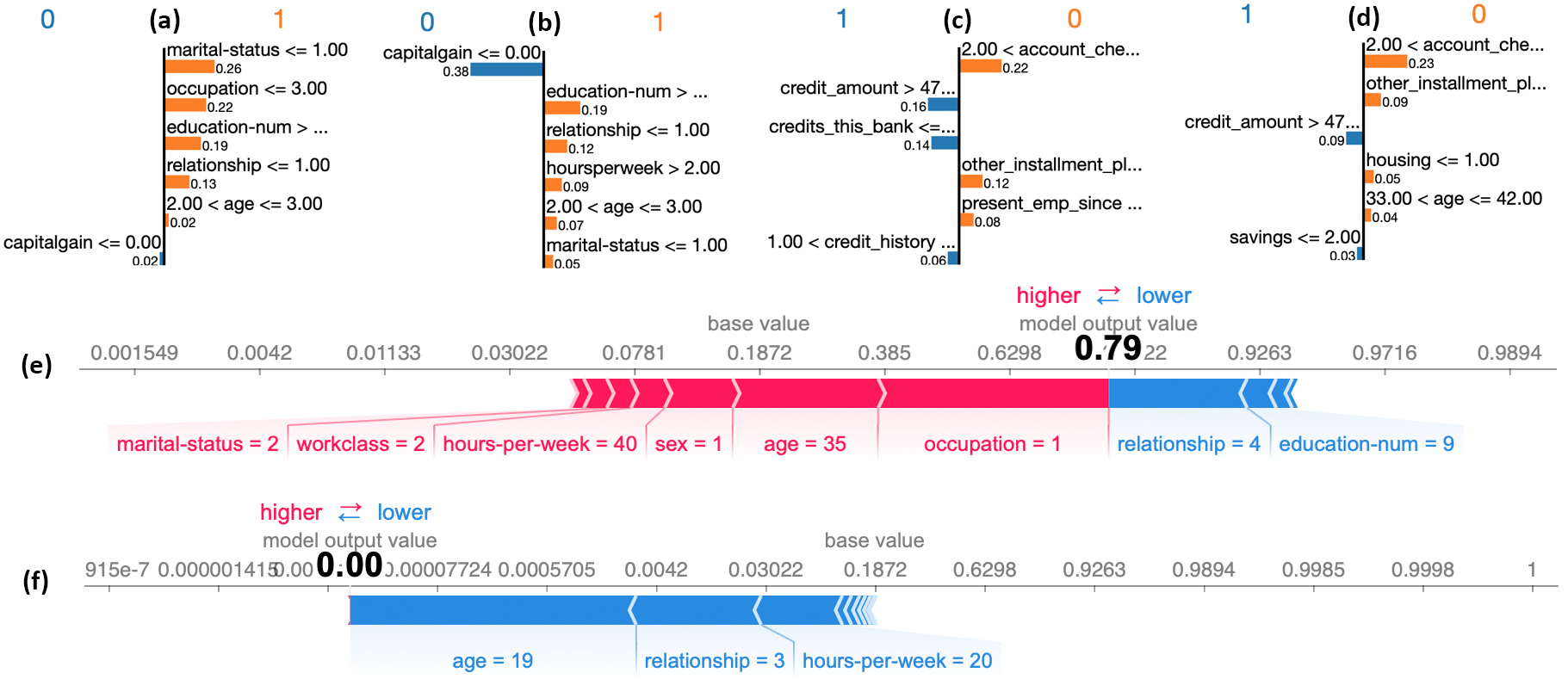}
    \caption{\textit{TOP}: \lime{} application on the same record for \adult{} (a/b), \german{} (c/d): a/c are the LG model explanation and b/d the CAT model explanation. All the models correctly predicted the output class. \textit{BOTTOM}: Force plot returned by \shap{} explaining XGB on two records of \adult{}: (e), labeled as class 1 ($> 50K$)  and, (f), labeled as class 0 ($\leq 50K$).
    Only the features that contributed more (i.e. higher \shap{}'s values) to the classification are reported.}
    \label{fig:shap-lime-local-all}
\end{figure}

\textbf{LIME}, Local Interpretable Model-agnostic Explanations~\cite{ribeiro2016should}, is a model-agnostic explanation approach which returns explanations as features importance vectors. 
The main idea of \lime{} is that the explanation may be derived locally from records generated randomly in the neighborhood of the instance that has to be explained.
The key factor is that it samples instances both in the vicinity of $x$ (which have a high weight) and far away from $x$ (low weight), exploiting $\pi_x$, a proximity measure able to capture the locality. 
We denote $b$ the black-box and $x$ the instance we want to explain. 
To learn the local behavior of $b$, \lime{} draws samples weighted by $\pi_x$. It samples these instances around $x$ by drawing nonzero elements of $x$ uniformly at random. 
This gives to \lime{} a perturbed sample of instances $\{z \in \mathbb{R}^d \}$ to fed to the model $b$ and obtain $b(z)$. 
They are then used to train the explanation model $g(\cdot)$: a sparse linear model on the perturbed samples. The local feature importance explanation consists of the weights of the linear model. 
A number of papers focus on overcoming the limitations of \lime{}, providing several variants of it. 
{\scshape d\lime{}}~\cite{zafar2019dlime} is a deterministic version in which the neighbors are selected from the training data by an agglomerative hierarchical clustering. 
{\scshape i\lime{}}~\cite{elshawi2019ilime} randomly generates the synthetic neighborhood using weighted instances.
{\scshape a\lime{}}~\cite{shankaranarayana2019alime} runs the random data generation only once at ``training time''.
{\scshape kl-lime}~\cite{peltola2018local} adopts a Kullback-Leibler divergence to explain Bayesian predictive models.
{\scshape qlime}~\cite{bramhall2020qlime} also consider nonlinear relationships using a quadratic approximation. 

In Figure~\ref{fig:shap-lime-local-all} are reported examples of \lime{} explanations relative to our experimentation on \adult{} (top) and \german{} (bottom)\footnote{For reproducibility reasons, we fixed the random seed.}.
We fed the same record into two black-boxes, and then we explained it. Interestingly, for \adult{}, \lime{} considers a similar set of features as important (even if with different values of importance) for the two models: on $6$ features, only one differs.
A different scenario is obtained by the application of \lime{} on \german{}: different features are considered necessary by the two models. However, the confidence of the prediction between the two models is quite different: both of them predict the output label correctly, but CAT has a higher value, suggesting that this could be the cause of differences between the two explanations.

\textbf{SHAP}, SHapley Additive exPlanations~\cite{lundberg2017unified}, is a local-agnostic explanation method, which can produce several types of models.
All of them compute \shap{} values: a unified measure of feature importance based on the Shapley values\footnote{We refer the intrested reader to: \url{https://christophm.github.io/interpretable-ml-book/shapley.html}}, a concept from cooperative game theory. 
In particular, the different explanation models proposed by \shap{} differ in how they approximate the computation of the \shap{} values.
All the explanation models provided by \shap{} are called \textit{additive feature attribution methods} and respect the following definition: $g(z') = \phi_{0} + \sum_{i=1}^{M}\phi_{i}z'_{i}$, where $z' \approx x$ as a real number, $z' \in [0,1]$, $\phi_i \in \mathbb{R}$ are effects assigned to each feature, while $M$ is the number of simplified input features.
\shap{} has 3 properties: \textit{(i)} \textit{local accuracy}, meaning that $g(x)$ matches $b(x)$; \textit{(ii)} \textit{missingness}, which allows for features $x_i = 0$ to have no attributed impact on the \shap{} values; \textit{(iii)} \textit{consistency}, meaning that if a model changes so that the marginal contribution of a feature value increases (or stays the same), the \shap{} value also increases (or stays the same).
The construction of the \shap{} values allows to employ them both \textit{locally}, in which each observation gets its own set of \shap{} values; and \textit{globally}, by exploiting collective \shap{} values. 
There are 5 strategies to compute \shap{}'s values: \textit{KernelExplainer}, \textit{LinearExplainer}, \textit{TreeExplainer},  \textit{GradientExplainer}, and \textit{DeepExplainer}.
In particular, the \textit{KernelExplainer} is an agnostic method while the others are specifically designed for different kinds of ML models.

\begin{figure}[t]
    \begin{center}
    \includegraphics[clip,width=0.3\linewidth]{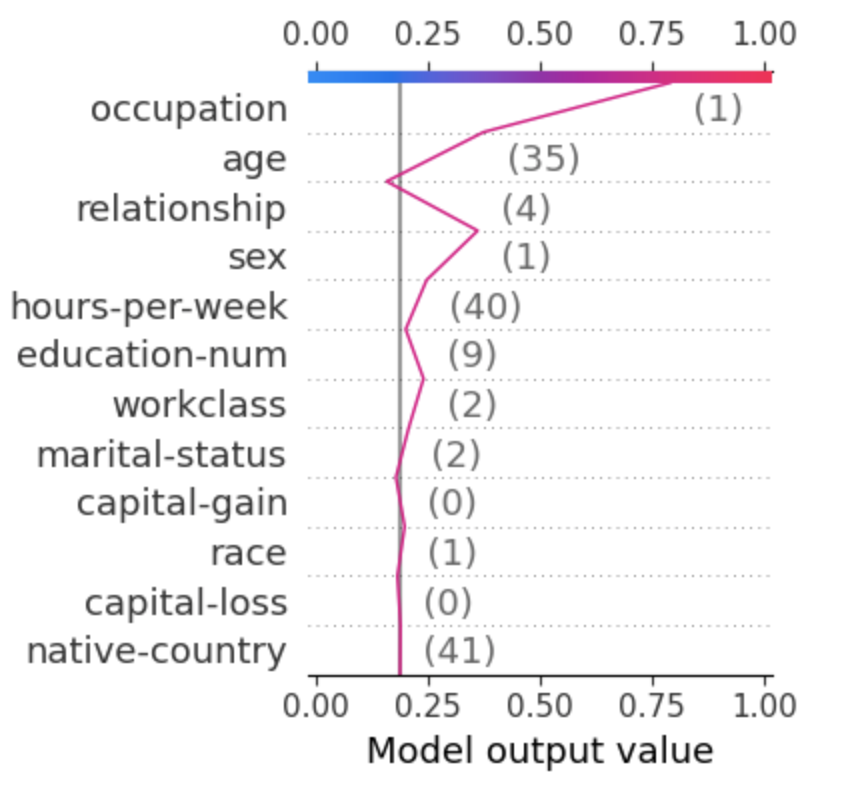}%
    \includegraphics[clip,width=0.3\linewidth]{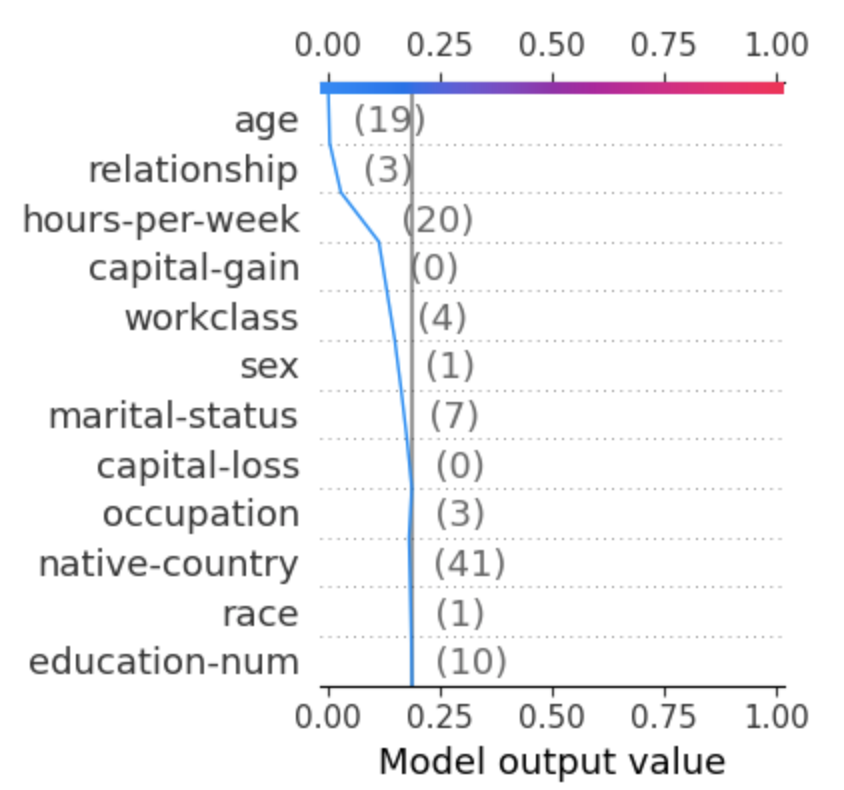}%
    \includegraphics[angle=90,trim=0.6cm 0cm 2cm 0cm,clip,width=.4\textwidth]{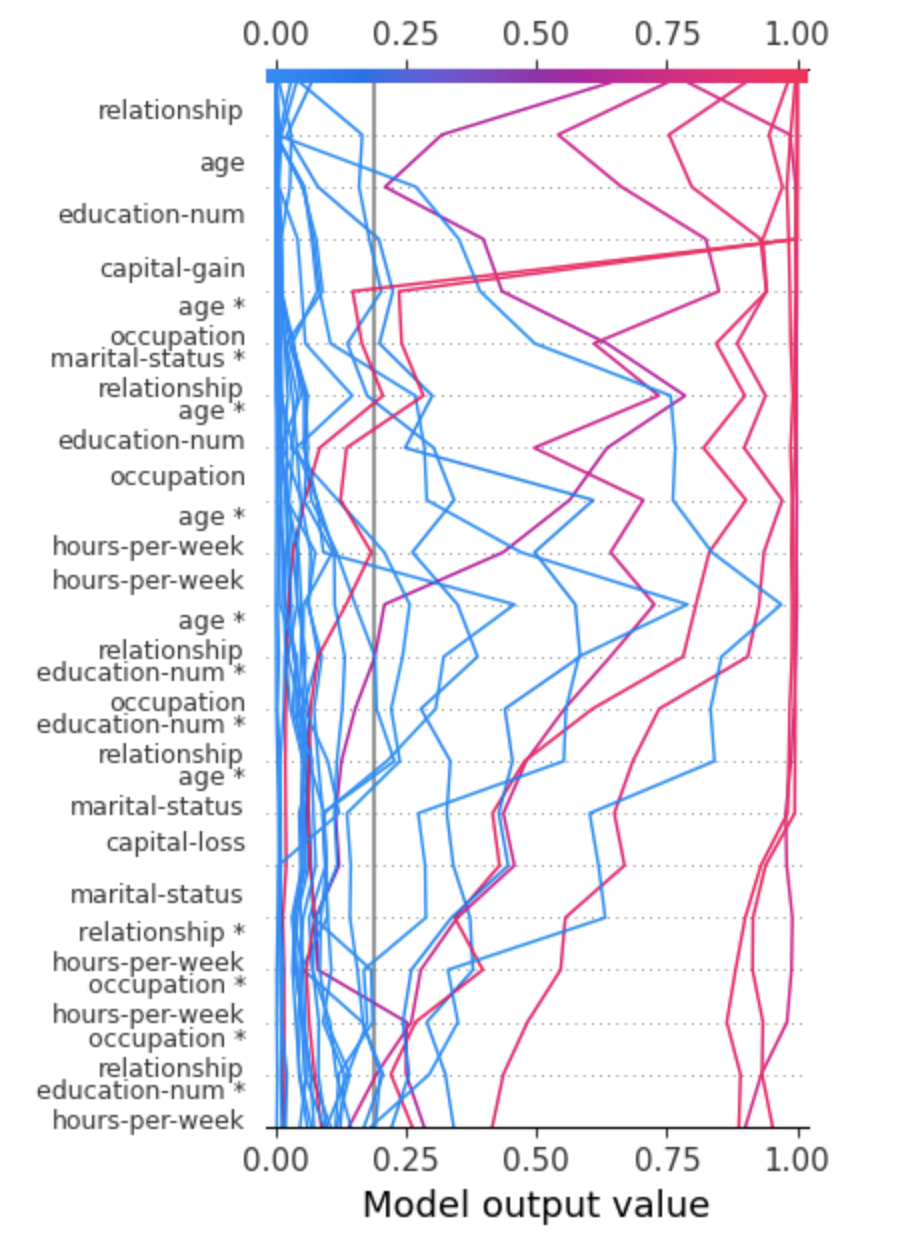}
    \caption{
    \shap{} application on \adult{}: a record labelled $> 50K$ (top-left) and one as $\leq 50K$(top-right). They are obtained applying the TreeExplainer on a XGB model and then the \textit{decision plot}, in which all the input features are shown. At the bottom, the application of \shap{} to explain the outcome of a set of record by XGB on \adult{}. The interaction values among the features are reported.}
  \label{fig:shap_global}
  \end{center}
\end{figure}

In our experiments with \shap{} we applied: \textit{(i)} the \textit{LinearExplainer} to the LG models, \textit{(ii)} the \textit{TreeExplainer} to the XGB and \textit{(iii)} \textit{KernelExplainer} to the CAT models. 
In Figures~\ref{fig:shap-lime-local-all} we report the application of \shap{} on \adult{} through \textit{force plot}. 
The plot shows how each feature contributes to pushing the output model value away from the base value, which is an average among the training dataset's output model values. 
The red features are pushing the output value higher while the ones in blue are pushing it lower. 
For each feature is reported the actual value for the record under analysis. 
Only the features with the highest \shap{} values are shown in this plot. 
In the first force plot, the features that are pushing the value higher are contributing more to the output value, and it is possible to note it by looking at the base value ($0.18$) and the actual output value ($0.79$). 
In the force plot on the right, the output value is $0.0$, and it is interesting to see that only \textit{Age}, \textit{Relationship} and \textit{Hours Per Week} are contributing to pushing it lower. 
Figure~\ref{fig:shap_global} (left and center) depicts the \textit{decision plots}: in this case, we can see the contribution of all the input features in decreasing order of importance. 
In particular, the line represents the feature importance for the record under analysis. 
The line starts at its corresponding observations' predicted value. 
In the first plot, predicted as class $>50k$, the feature \textit{Occupation} is the most important, followed by \textit{Age} and \textit{Relationship}. 
For the second plot, instead, \textit{Age}, \textit{Relationship} and \textit{Hours Per Week} are the most important feature. 
Besides the local explanations, \shap{} also offers a global interpretation of the model-driven by the local interpretations. 
Figure~\ref{fig:shap_global} (right) reports a global decision plot that represents the feature importance of $30$ records of \adult{}. 
Each line represents a record, and the predicted value determines the color of the line.

\begin{figure}[t]
    \centering
    \includegraphics[trim={0 1.5cm 0 2.5cm},clip,width=0.45\linewidth]{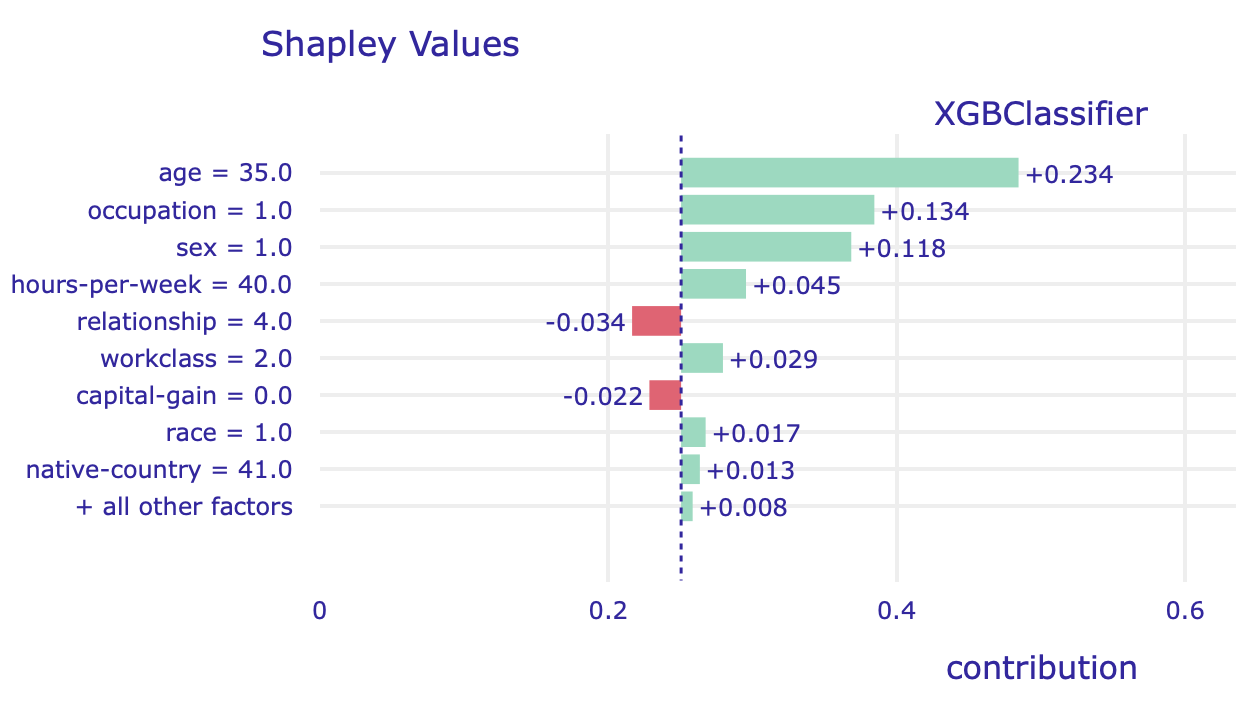}%
    \includegraphics[trim={0 1.5cm 0 2.5cm},clip,width=0.45\linewidth]{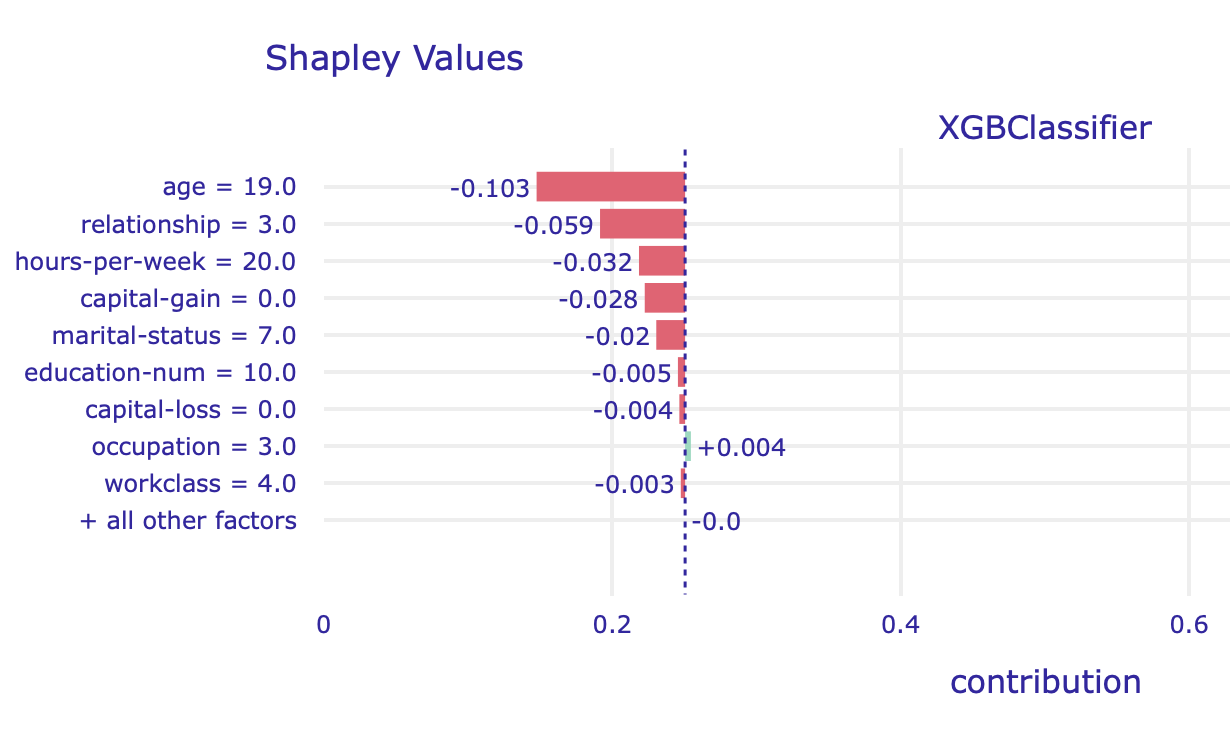}
    \includegraphics[trim={0 1.5cm 0 2.5cm},clip,width=0.45\linewidth]{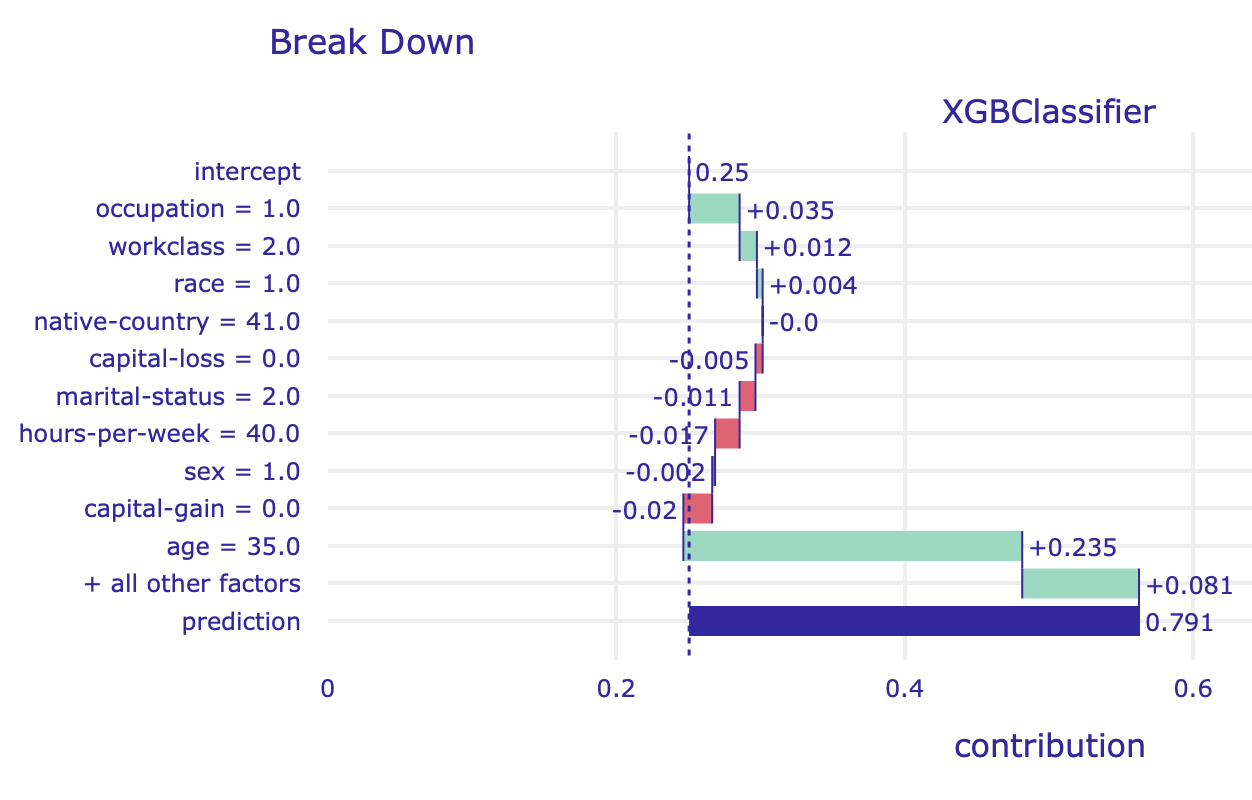}%
    \includegraphics[trim={0 1.5cm 0 2.5cm},clip,width=0.45\linewidth]{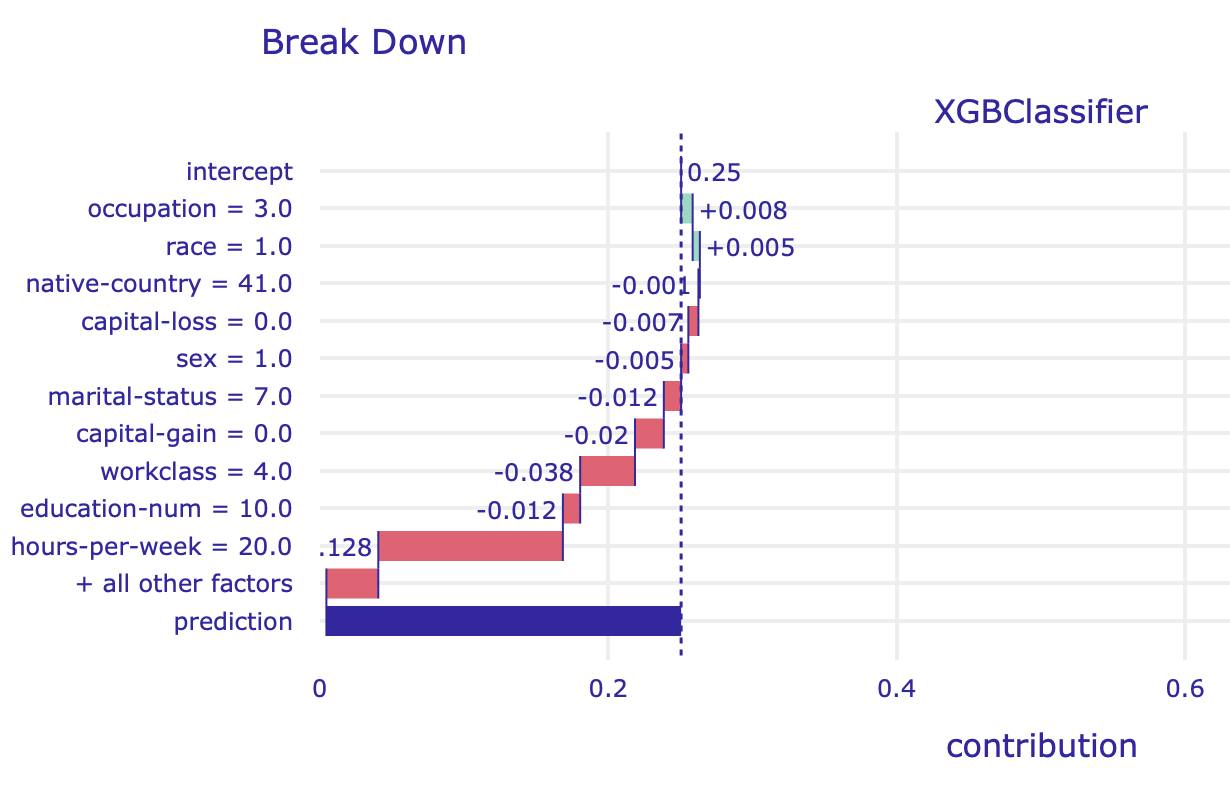}
    \caption{Explanations of \dalex{} for two records of \adult{}: $b(x) = 0$ ($\leq 50$) (left), $b(x) = 1$ ($> 50K$) (right) to explain an XGB in form of Shapely values (top), break down plots (bottom). The y-axis is the features important, the x-axis the positive/negative contribution. }
    \label{fig:dalex-adult-local}
\end{figure} 

\textbf{DALEX}~\cite{biecekexplanatory} is a post-hoc, local and global agnostic explanation method.
Regarding local explanations, \dalex{} contains an implementation of a \textit{variable attribution} approach~\cite{robnik2008explaining}. 
It consists of a decomposition of the model's predictions, in which each decomposition can be seen as a local gradient and used to identify the contribution of each attribute. 
Moreover, \dalex{} contains the \textit{ceteris-paribus} profiles, which allow for a \textit{What-if} analysis by examining the influence of a variable by fixing the others. 
Regarding the global explanations, \dalex{} contains different exploratory tools: model performance measures, variable importance measures, residual diagnoses, and partial dependence plot.  
In Figure~\ref{fig:dalex-adult-local} are reported some local explanations obtained by the application of \dalex{} to an XGB model on \adult{}. 
On the left are reported two explanation plots for a record classified as class $> 50k$. 
On the top, there is a visualization based on Shapely values, which highlights as most important the feature \textit{Age} ($35$ years old), followed by \textit{occupation}. 
At the bottom, there is a \textit{Breakdown} plot, in which the green bars represent positive changes in the mean predictions, while the red ones are negative changes. 
The plot also shows the \textit{intercept}, which is the overall mean value for the predictions.  
It is interesting to see that \textit{Age} and \textit{occupation} are the most important features that positively contributed to the prediction for both the plots. In contrast, \textit{Sex} is positively important for Shapely values but negatively important for the \textit{Breakdown} plot. 
On the right part of Figure~\ref{fig:dalex-adult-local} we report a record classified as $< 50k$. 
In this case, there are important differences in the feature considered most important by the two methods: for the Shapely values, \textit{Age} and \textit{Relationship} are the two most important features, while in the Breakdown plot \textit{Hours Per Week} is the most important one. 

\textbf{CIU}, Contextual Importance and Utility~\cite{anjomshoae2020py}, is a local, agnostic explanation method. 
{\scshape ciu} is based on the idea that the context, i.e.,~the set of input values being tested, is a key factor in generating faithful explanations. 
The authors suggest that a feature that may be important in a context may be irrelevant in another one. 
{\scshape ciu} explains the model's outcome based on the \textit{contextual importance (CI)}, which approximates the overall importance of a feature in the current context, and on the \textit{contextual utility (CU)}, which estimates how good the current feature values are for a given output class. 
Technically, {\scshape ciu} computes the values for CI and CU by exploiting Monte Carlo simulations. 
We highlight that this method does not require creating a simpler model to employ for deriving the explanations. 

\begin{figure}[t]
    \centering
    \includegraphics[width=0.9\linewidth]{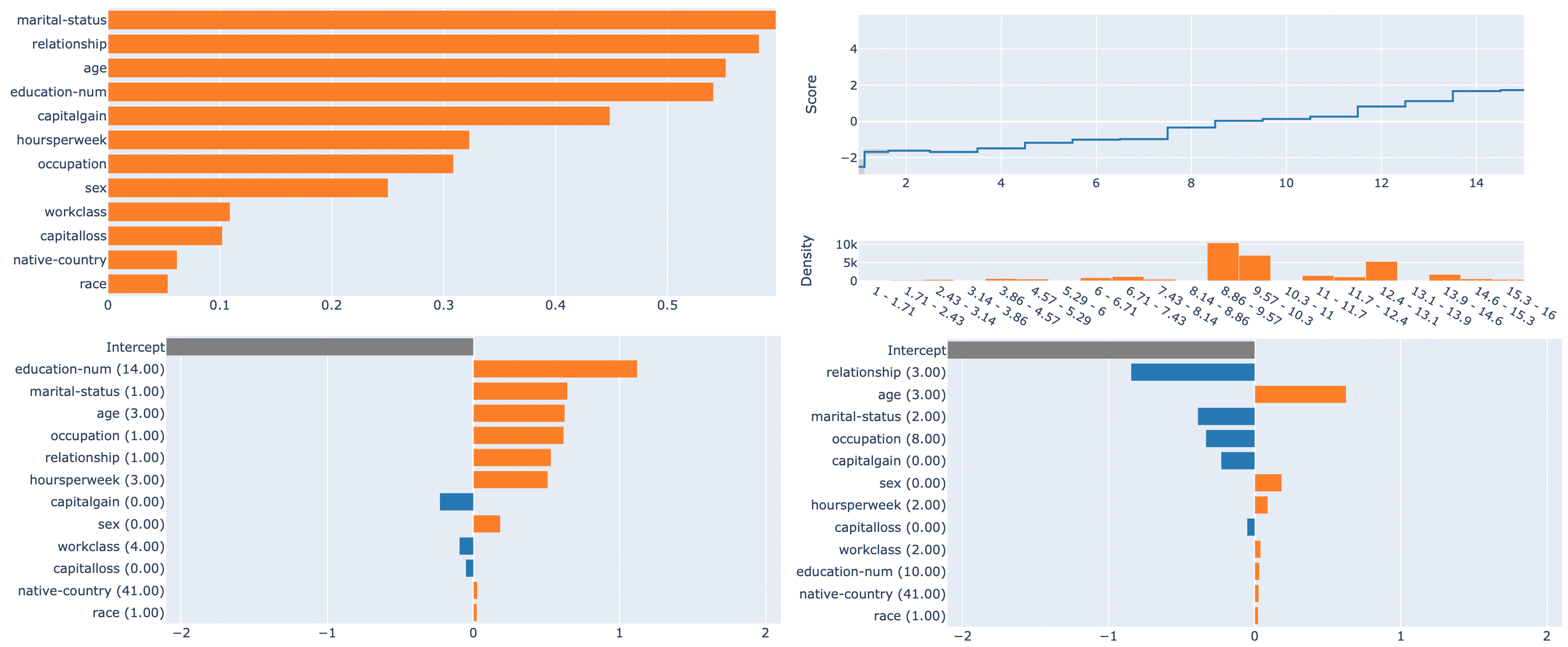}
    \caption{\textit{TOP}: Results of \ebm{} on \adult{}: overall global explanation (left), example of a global explanation for \textit{education number} (right).\\
    \textit{BOTTOM}: Local explanations of \ebm{} on \adult{}: left, a record classified as 1 ($> 50K$); right a record classified as 0 ($\leq 50$).}
    \label{fig:ebm}
\end{figure}

\textbf{NAM}, Neural Additive Models~\cite{agarwal2020neural}, is a different extension of {\scshape gam}. 
This method aims to combine the performance of powerful models, such as deep neural networks, with the inherent intelligibility of generalized additive models. 
The result is a model able to learn graphs that describe how the prediction is computed. {\scshape nam} trains multiple deep neural networks in an additive fashion such that each neural network attend to a single input feature.

\subsection{Rule-based Explanation}
\label{sec:rules}
Decision rules give the end-user an explanation about the reasons that lead to the final prediction.
The majority of explanation methods for tabular data are in this category since decision rules are human-readable.
A decision rule $r$, also called \textit{factual} or \textit{logic} rule~\cite{Guidotti2018LocalRE}, has the form $p \rightarrow y$, in which $p$ is a premise, composed of a Boolean condition on feature values, while $y$ is the consequence of the rule.
In particular, $p$ is a conjunction of split conditions of the form $x_i {\in} [v_i^{(l)}, v_i^{(u)}]$, where $x_i$ is a feature and $v_i^{(l)}, v_i^{(u)}$ are lower and upper bound values in the domain of $x_i$ extended with $\pm \infty$.
An instance $x$ \emph{satisfies} $r$, or $r$ \emph{covers} $x$, if every Boolean conditions of $p$ evaluate to true for $x$.
If the instance $x$ to explain satisfies $p$, the rule $p \rightarrow y$ represents then a candidate explanation of the decision $g(x) = y$. 
Moreover, if the interpretable predictor mimics the behavior of the black-box in the neighborhood of $x$, we further conclude that the rule is a candidate local explanation of $b(x) = g(x) = y$.
We highlight that, in the context of rules we can also find the so-called \textit{counterfactual rules}~\cite{Guidotti2018LocalRE}. 
Counterfactual rules have the same structure of decision rules, with the only difference that the consequence of the rule $\overline{y}$ is different w.r.t.~$b(x) = y$.  
They are important to explain to the end-user what should be changed to obtain a different output. 
An example of a rule explanation is $r = \{\mathit{age} < 40, \mathit{income} < 30k, \mathit{education} \leq \mathit{Bachelor}\}, y = \mathit{deny}$. In this case, the record $\{ \mathit{age} = 18, \mathit{income} = 15k, \mathit{education} = \mathit{High school} \}$ satisfies the rule above. A possible counterfactual rule, instead can be: $r = \{ \mathit{income} > 40k, \mathit{education} \geq \mathit{Bachelor}\}, y = \mathit{allow}$.

\begin{figure}[t]
\footnotesize
\centering
    \begin{minipage}[t][2.7cm][t]{.48\linewidth}\em
        \begin{tabular}{lp{45ex}}
            %class 1
            x\ =& \{ \text{Education} $=$ \text{Bachelors}, %\\ &
            \text{Occupation} $=$ \text{Prof-specialty},  \text{Sex} $=$ \text{Male},
            \textit{NativeCountry} = \text{Vietnam}, \text{Age} $=$ 35, \text{Workclass} $=$ 3, \text{HoursWeek} $=$ 40, \text{Race} $=$ \text{Asian-Pac-Islander}, \text{MaritialStatus} $=$\text{Married-civ},
            \text{Relationship} $=$ \textit{Husband}, \text{CapitalGain} = 0, \text{CapitalLoss} = 0\}, $> 50k$
        \end{tabular}
    \end{minipage}
    \begin{minipage}[t][2.7cm][t]{.45\linewidth}\em
        \begin{tabular}{lp{45ex}}
            %class 0
            x\ =& \{ \text{Education} $=$ \text{College},  %\\ &
            \text{Occupation} $=$ \text{Sales}, \text{Sex} $=$ \text{Male}, \textit{NativeCountry} = \text{US}, \text{Age} $=$ 19, \text{Workclass} $=$ 2, \text{HoursWeek} $=$ 15, \text{Race} $=$ \text{White}, \text{MaritialStatus} $=$ \text{Married-civ}, \text{Relationship} $=$ \textit{Husband}, \text{CapitalGain} = 2880, \text{CapitalLoss} = 0 \}, $\leq 50k$
        \end{tabular}
    \end{minipage}
    \begin{minipage}[t][1.5cm][t]{.48\linewidth}\em
        \begin{tabular}{lp{40ex}}
        $r_{\text{anchor}}$ \ =& \{ \text{EducationNum} $>$ \text{Bachelors}, \\& \text{Occupation} $\leq$ 3.00, \text{HoursWeek} $>$ 20, \text{Relationship} $\leq$ 1.00, 34 $<$ \text{Age} $\leq$ 41 \}  $\rightarrow$ $> 50k$ \\
        %bb(x) = 1
        
        \end{tabular}
    \end{minipage}
    \begin{minipage}[t][1.5cm][t]{.45\linewidth}\em
        \begin{tabular}{lp{45ex}}
            $r_{\text{anchor}}$ \ =&  \{\text{Education} $\leq$ \text{College}, \\& \text{MaritialStatus} $>$ 1.00 \} $\rightarrow$ $\leq 50k$ \\
            \\
            \\
            %bb(x) = 0
        \end{tabular}
    \end{minipage}
    \begin{minipage}[t][1.55cm][t]{.48\linewidth}\em
        \begin{tabular}{lp{45ex}}
            $r_{\text{lore}}$\ =&  \{ \text{Education} $>$ \text{5-6th}, \text{Race} $>$ 0.86, \text{WorkClass} $\leq$ 3.41, \text{CapitalGain} $\leq$ 20000, \text{CapitalLoss} $\leq$ 1306 \} $\rightarrow$ $> 50k $\\
            %b(x)\ =& 1
        \end{tabular}
    \end{minipage}
    \begin{minipage}[t][1.55cm][t]{.45\linewidth}\em
        \begin{tabular}{lp{40ex}}
            $r_{\text{lore}}$\ =&  \{\text{Education} $\leq$ \text{Masters}, \text{Occupation} $>$ -0.34, \text{HoursWeek} $\leq$ 40, \text{WorkClass} $\leq$ 3.50, \text{CapitalGain} $\leq$ 10000, \text{Age} $\leq$ 34\} $\rightarrow$ $\leq 50k$ \\
            %b(x)\ =& 0
        \end{tabular}
    \end{minipage}
    \begin{minipage}[t][1cm][t]{.48\linewidth}\em
        \begin{tabular}{lp{45ex}}
            $c_{\text{lore}}$ \ =&  \{\text{CapitalLoss} $\geq$ 436 \} $\rightarrow$ $\leq 50k$\\
            \\
        \end{tabular}
    \end{minipage}
    \begin{minipage}[t][1cm][t]{.45\linewidth}\em
        \begin{tabular}{lp{45ex}}
            $c_{\text{lore}}$ \ =&  \{\text{Education} $>$ \text{Masters} \} $\rightarrow$ $>50k$ \\ &
                 \{\text{CapitalGain} $>$ 20000 \} $\rightarrow$ $>50k$ \\ &
                 \{\text{Occupation} $\leq$ -0.34 \} $\rightarrow$ $>50k$
        \end{tabular}
    \end{minipage}
    \caption{Explanations of \anchor{} and \lore{} for \adult{} to explain an XGB model.}
    \label{fig:anchor-adult}
\end{figure}

\textbf{ANCHOR}~\cite{ribeiro2018anchors} is a model-agnostic system that outputs rules as explanations. 
This approach's name comes from the output rules, called \textit{anchors}.
The idea is that, for decisions on which the anchor holds, changes in the rest of the instance's feature values do not change the outcome. 
Formally, given a record $x$, $r$ is an anchor if $r(x) = b(x)$.
To obtain the anchors, \anchor{} perturbs the instance $x$ obtaining a set of synthetic records employed to extract anchors with precision above a user-defined threshold.
First, since the synthetic generation of the dataset may lead to a massive number of samples \anchor{} exploits a multi-armed bandit algorithm~\cite{katehakis1987multi}. 
Second, since the number of all possible anchors is exponential \anchor{} uses a bottom-up approach and a beam search.
Figure~\ref{fig:anchor-adult} reports some rules obtained by applying \anchor{} to a XGB model on \adult{}.
The first rule has a high precision ($0.96\%$) but a very low coverage ($0.01\%$). 
It is interesting to note that the first rule contains \textit{Relationship} and \textit{Education Num}, which are the features highlighted by most of the explanation models proposed so far. 
In particular, in this case, for having a classification $>50k$, the \textit{Relationship} should be husband and the \textit{Education Num} at least bachelor degree. 
\textit{Education Num} can also be found in the second rule, in which case has to be less or equal to College, followed by the \textit{Maritial Status}, which can be anything other than married with a civilian. 
This rule has an even better precision ($0.97\%$) and suitable coverage ($0.37\%$).

\textbf{LORE}, LOcal Rule-based Explainer~\cite{Guidotti2018LocalRE}, is a local agnostic method that provides faithful explanations in the form of rules and counterfactual rules. 
\lore{} is tailored explicitly for tabular data. 
It exploits a genetic algorithm for creating the neighborhood of the record to explain.
Such a  neighborhood produces a more faithful and dense representation of the vicinity of $x$ w.r.t.~\lime{}. 
Given a black-box $b$ and an instance $x$, with $b(x) = y$, \lore{} first generates a synthetic set $Z$ of neighbors through a genetic algorithm. 
Then, it trains a decision tree classifier $g$ on this set labeled with the black-box outcome $b(Z)$. 
From $g$, it retrieves an explanation that consists of two components: \emph{(i)} a \textit{factual} decision rule, that corresponds to the path on the decision tree followed by the instance $x$ to reach the decision $y$, and \emph{(ii)} a set of counterfactual rules, which have a different classification w.r.t.~$y$. 
This counterfactual rules set shows the conditions that can be varied on $x$ in order to change the output decision.
In Figure~\ref{fig:anchor-adult} we report the factual and counterfactual rules of \lore{} for the explanation of the same records showed for \anchor{}. 
It is interesting to note that, differently from \anchor{} and the others models proposed above, \lore{} explanations focuses more on the \textit{Education Num}, \textit{Occupation}, \textit{Capital Gain} and \textit{Capital Loss}, while the features about the relationship are not present.

\textbf{RuleMatrix}~\cite{ming2018rulematrix} is a post-hoc agnostic explanator tailored for the visualization of the rules extracted. 
First, given a training dataset and a black-box model, {\scshape rulematrix} executes a rule induction step, in which a \textit{rule list} is extracted by sampling the input data and their predicted label by the black-box. 
Then, the rules extracted are filtered based on thresholds of confidence and support. 
Finally, {\scshape rulematrix} outputs a visual representation of the rules. 
The user interface allows for several analyses based on plots and metrics, such as fidelity. 

One of the most popular ways for generating rules is by extracting them from a decision tree. In particular, due to the method's simplicity and interpretability, decision trees explain black-box models' overall behavior. Many works in this setting are model specific to exploit some structural information of the black-box model under analysis.

\textbf{TREPAN}~\cite{craven1996extracting} is a model-specific global explainer tailored for neural networks. 
Given a neural network $b$, {\scshape trepan} generates a decision tree $g$ that approximates the network by maximizing the gain ratio and the model fidelity.

\textbf{DecText} is a global model-specific explainer tailored for neural networks~\cite{boz2002extracting}. 
The aim of {\scshape dectext} is to find the most relevant features. 
To achieve this goal, {\scshape dectext} resembles {\scshape trepan}, with the difference that it considers four different splitting methods. 
Moreover, it also considers a pruning strategy based on fidelity to reduce the final explanation tree's size. 
In this way, {\scshape dectext} can maximize the fidelity while keeping the model simple.

\textbf{MSFT}~\cite{chipman1998making} is a specific global post-hoc explanation method that outputs decision trees starting from random forests. 
It is based on the observation that, even if random forests contain hundreds of different trees, they are quite similar, differing only for few nodes. 
Hence, the authors propose dissimilarity metrics to summarize the random forest trees using a clustering method. 
Then, for each cluster, an archetype is retrieved as an explanation.   

\textbf{CMM}, Combined Multiple Model procedure~\cite{domingos1998knowledge}, is a specific global post-hoc explanation method for tree ensembles. 
The key point of {\scshape cmm} is the data enrichment. 
In fact, given an input dataset $X$, {\scshape cmm} first modifies it $n$ times. 
On the $n$ variants of the dataset, it learns a black-box. 
Then, random records are generated and labeled using a bagging strategy on the black-boxes. 
In this way, the authors were able to increase the size of the dataset to build the final decision tree.  

\textbf{STA}, Single Tree Approximation~\cite{zhou2016interpreting}, is a specific global post-hoc explanation method tailored for random forests, in which the decision tree, used as an explanation, is constructed by exploiting test hypothesis to find the best splits. 

\begin{figure}[t]
\footnotesize
\centering
    \begin{minipage}[c]{.50\linewidth}\em
        \begin{tabular}{lp{50ex}}
            r\ =&  \{\text{Age} $>$ 34, \text{HoursPerWeek} $>$ 20, \\ &
            \text{Education} $>$ 12.5, \text{Occupation} $\leq$ 3.5, \\ & 
            \text{Relationship} $\leq$ 2.5, \text{CapitalGain} $\leq$ 20000,
            \text{CapitalLoss} $\leq$ 223\} $\rightarrow$ 1 \\&
            \{ \text{Prec} = 0.79 \%, \text{Rec} = 0.15 \%, \text{Cov} = 1 \}
        \end{tabular}
    \end{minipage}
    \begin{minipage}[c]{.40\linewidth}\em
        \begin{tabular}{lp{50ex}}
            r\ =&  \{ \text{Age} $\leq$ 19, \text{WorkClass} $\leq$ 4.5, \\ &
            \text{HoursPerWeek} $\leq$ 30, \text{Education} $\leq$ 13.5, \\ &
            \text{Education} $>$ 3.5, \text{MaritialStatus} $>$ 2.5, \\& 
            \text{Relationship} $>$ 2.5, \text{CapitalLoss} $\leq$ 1306 \} $\rightarrow$ 0 \\&
            \{ \text{Prec} = 0.99 \%, \text{Rec} = 0.19 \%, \text{Cov} = 1 \}    
        \end{tabular}
    \end{minipage}
    \caption{{\scshape skoperule} global explanations of XGB on \adult{}. On the left, a rule for class $> 50k$, on the right for class $<50k$.}
    \label{fig:skope-adult}
\end{figure}

\textbf{SkopeRules} is a post-hoc, agnostic model, both global and local \footnote{\url{https://skope-rules.readthedocs.io/en/latest/skope_rules.html}}, based on the {\scshape rulefit}~\cite{Friedman2008PREDICTIVELV} idea to define an ensemble method and then extract the rules from it. 
{\scshape skope-rules} employs fast algorithms such as bagging or gradient boosting decision tress. 
After extracting all the possible rules, {\scshape skope-rules} removes rules redundant or too similar by a similarity threshold. 
Differently from {\scshape rulefit}, the scoring method does not solve the L1 regularization. 
Instead, the weights are given depending on the precision score of the rule.
We can employ {\scshape skoperules} in two ways: \textit{(i)} as an explanation method for the input dataset, which describes, by rules, the characteristics of the dataset; \textit{(ii)} as a transparent method 
by outputting the rules employed for the prediction.
In Figure~\ref{fig:skope-adult}, we report the rule extracted by {\scshape rulefit} with highest precision and recall for each class for \adult{}. 
Similarly to the models analyzed so far, we can find \textit{Relationship} and \textit{Education} among the features in the rules. 
In particular, for the first rule, for $>50k$, the \textit{Education} has to be at least a Bachelor degree, while for the other class, it has to be at least fifth or sixth. 
Interestingly, it is also mentioned the \textit{Capital Gain} and \textit{Capital Loss} which were considered as important by few models, such as \lore{}.
We also tested {\scshape skoperules} to create a rule-based classifier obtaining a precision of $0.68$ on \adult{}.

Moreover, with {\scshape skoperules}, it is possible to explain, using rules, the entire dataset without considering the output labels; or obtain a set of rules for each output class. We tested both of them, but we report only the case of rules for each class. In particular, we report the rule with the highest precision and recall for each class for \adult{} in Figure~\ref{fig:skope-adult}.

\textbf{Scalable-BRL}~\cite{yang2017scalable} is an interpretable probabilistic rule-based classifier that optimizes the posterior probability of a Bayesian hierarchical model over the rule lists. 
The theoretical part of this approach is based on~\cite{letham2015interpretable}. 
The particularity of {\scshape scalable-brl} is that it is \textit{scalable}, due to a specific bit vector manipulation.

\textbf{GLocalX}~\cite{SetzuGlocalX} is a rule-based explanation method which exploits a novel approach: the \textit{local to global} paradigm. The idea is to derive a global explanation by subsuming local logical rules. 
{\scshape GLocalX} start from an array of factual rules and following a hierarchical bottom up fashion merges rules covering similar records and expressing the same conditions.
{\scshape GLocalX} finds the smallest possible set of rules that is: \textit{(i)} general, meaning that the rules should apply to a large subset of the dataset; \textit{(ii)} has high accuracy. 
The final explanation proposed to end-user is a set of rules. 
In~\cite{SetzuGlocalX} the authors validated the model in constrained settings: limited or no access to data or local explanations.
A simpler version of {\scshape GLocalX} is presented~\cite{setzu2019global}: here, the final set of rules is selected through a scoring system based on rules generality, coverage, and accuracy.

\subsection{Prototypes}
\label{sec:proto_tabular}
A prototype, also called archetype or artifact, is an object representing a set of similar records. 
It can be \emph{(i)} a record from the training dataset close to the input data $x$; \emph{(ii)} a centroid of a cluster to which the input $x$ belongs to.
Alternatively, \emph{(iii)} even a synthetic record, generating following some ad-hoc process.
Depending on the explanation method considered, different definitions and requirements to find a prototype are considered. 
Prototypes serve as examples: the user understands the model's reasoning by looking at records similar to his/hers.

\textbf{MMD-CRITIC}~\cite{mmdcritic} is a ``before the model'' methodology, in the sense that it only analyses the distribution of the dataset under analysis. 
It produces prototypes and criticisms as explanations for a dataset using \textit{Maximum Mean Discrepancy (MMD)}. 
The first ones explain the dataset's general behavior, while the latter represent points that are not well explained by the prototypes.
\mmdcritic{} selects prototypes by measuring the difference between the distribution of the instances and the instances in the whole dataset. 
The set of instances nearer to the data distribution are called prototypes, and the farthest are called criticisms.
\mmdcritic{} shows only minority data points that differ substantially from the prototype but belong in the same category. 
For criticism, \mmdcritic{} selects criticisms from parts of the dataset underrepresented by the prototypes, with an additional constraint to ensure the criticisms are diverse.

\textbf{ProtoDash}~\cite{gurumoorthy2019efficient} is a variant of \mmdcritic{}. 
It is an explainer that employs prototypical examples and criticisms to explain the input dataset. 
Differently, w.r.t.~\mmdcritic{}, {\scshape protodash} associates non-negative weights, which indicate the importance of each prototype. 
In this way, it can reflect even some complicated structures. 

\textbf{Privacy-Preserving Explanations}~\cite{josef-microaggregation} is a local post-hoc agnostic explanability method which outputs prototypes and shallow trees as explanations. 
It is the first approach that considers the concept of \textit{privacy in explainability} by producing \textit{privacy protected explanations}. 
To achieve a good trade-off between privacy and comprehensibility of the explanation, the authors construct the explainer by employing \textit{micro aggregation} to preserve privacy. 
In this way, the authors obtained a set of clusters, each with a representative record $c_i$, where $i$ is the $i-th$ cluster. From each cluster, a \textit{shallow decision trees} is extracted to provide an exhaustive explanation while having good comprehensibility due to the limited depth of the trees. 
When a new record $x$ arrives, a representative record and its associated shallow tree are selected. In particular, from $g$ the representative $c_i$ closer to $x$ is selected, depending on the decision of the black-box. 

\textbf{PS}, Prototype Selection ({\scshape ps})~\cite{bien2011prototype} is an interpretable model, composed by two parts. First, the {\scshape ps} seeks a set of prototypes that better represent the data under analysis. It uses a set cover optimization problem with some constraints on the properties the prototypes should have. 
Each record in the original input dataset $D$ is then assigned to a representative prototype.
Then, the prototypes are employed to learn a nearest neighbor rule classifier.

\textbf{TSP}, Tree Space Prototype~\cite{tan2020tree}, is a local, post-hoc and model-specific approach, tailored for explaining random forests and gradient boosted trees.
The goal is to find prototypes in the tree space of the tree ensemble $b$. 
Given a notion of proximity between trees, with variants depending on the kind of ensemble, {\scshape tsp} is able to extract prototypes for each class.
%Exploiting this proximity measure, they were able to extract the prototypes for each class. %, used as explanations. 
Different variants are proposed for allowing for the selection of a different number of prototypes for each class.

\subsection{Counterfactuals}
\label{sec:counter_tabular}
Counterfactuals describe a dependency on the external facts that led to a particular decision made by the black-box model.
It focuses on the differences to obtain the opposite prediction w.r.t.~$b(x) = y$.
Counterfactuals are often addressed as the prototypes' opposite.
In~\cite{wachter2017counterfactual} is formalized the general form a counterfactual explanation should have: $b(x) = y$ was returned because variables of $x$ has values $x_1, x_2,...,x_n$. Instead, if $x$ had values $x_1^1, x_2^1,...,x_n^1$ and all the other variables has remained constant, $b(\overline{x}) = \neg y$ would have been returned, where $\overline{x}$ is the record $x$ with the suggested changes. An ideal counterfactual should alter the values of the variables as little as possible to find the closest setting under which $y$ is returned instead of $\neg y$. 
Regarding the counterfactual explainers, we can divide them into three categories: \textit{exogenous}, which generates the counterfactuals synthetically; \textit{endogenous}, in which the counterfactuals are drawn from a reference population, and hence they can produce more realistic instances w.r.t. the exogenous ones; or \textit{instance-based}, which exploits a distance function to detect the decision boundary of the black-box. 
There are several desiderata in this context: efficiency, robustness, diversity, actionability, and plausibility, among others~\cite{wachter2017counterfactual,karimi2020model,kanamori2020dace}.
To better understand the complex context and the many available possibilities, we refer the interested reader to ~\cite{artelt2019computation,verma2020counterfactual,byrne2020if}. In~\cite{byrne2020if} is presented a study that evaluates the understandability of factual and counterfactual explanations. The authors analyzed the mental model theory, which stated that people construct models that simulate the assertions described. They conducted experiments on a group of people highlighting that people prefer reasoning using mental models and find it challenging to consider probability, calculus, and logic.
There are many works in this area of research; hence, we briefly present only the most representative methods in this category.

\textbf{MAPLE}~\cite{plumb2018model} is a post-hoc local agnostic explanation method that can also be used as a transparent model due to its internal structure.
It combines random forests with feature selection methods to return feature importance based explanations.  
\maple{} is based on two methods: \textit{SILO} and \textit{DStump}. 
\textit{SILO} is employed for obtaining a local training distribution, based on the random forest leaves'. 
\textit{DStump}, instead, ranks the features by importance. 
\maple{} considers the best $k$ features from \textit{DStump} to solve a weighted linear regression problem. 
In this case, the explanation is the coefficient of the local linear model, i.e.,~the estimated local effect of each feature. 

\textbf{CEM}, Contrastive Explanations Method~\cite{dhurandhar2018explanations}, is a local, post-hoc and model-specific explanation method, tailored for neural networks which outputs \textit{contrastive explanations}. 
\cem{} has two components: \textit{Pertinent Positives (PP)}, which can be seen as prototypes, and are the minimal and sufficient factors that have to be present to obtain the output $y$; and \textit{Pertinent Negatives (PN)}, which are counterfactuals factors, that should be minimally and necessarily absent. 
\cem{} is formulated as an optimization problem over the perturbation variable $\delta$. 
In particular, given $x$ to explain, \cem{} considers $x_1 = x + \delta$, where $\delta$ is a perturbation applied to $x$. 
During the process, there are two values of $\delta$ to minimize: $\delta^{p}$ for the pertinent positives, and $\delta^{n}$ for the pertinent negatives. 
\cem{} solves the optimization problem with a variant that employs an autoencoder to evaluate the closeness of $x_1$ to the data manifold. 
\ceml{}~\cite{ceml} is also a Python toolbox for generating counterfactual explanations, suitable for ML models designed in Tensorflow, Keras, and PyTorch.

\textbf{DICE}, Diverse Counterfactual Explanations~\cite{mothilal2020explaining} is a local, post-hoc and agnostic method which solves an optimization problem with several constraints to ensure \textit{feasibility} and \textit{diversity} when returning counterfactuals. 
Feasibility is critical in the context of counterfactual since it allows avoiding examples that are unfeasible. 
As an example, consider the case of a classifier that determines whether to grant loans. 
If the classifier denies the loan to an applicant with a low salary, the cause may be low income. 
However, a counterfactual such as ``You have to double your salary'' may be unfeasible, and hence it is not a satisfactory explanation. 
The feasibility is achieved by imposing some constraints on the optimization problem: the proximity constraint, from~\cite{wachter2017counterfactual}, the sparsity constraint, and then user-defined constraints.
Besides feasibility, another essential factor is diversity, which provides different ways of changing the outcome class. 

\textbf{FACE}, Feasible and Actionable Counterfactual Explanations~\cite{poyiadzi2020face} is a local, post-hoc agnostic explanation method that focuses on returning ``achievable'' counterfactuals.
Indeed, \face{} uncovers ``feasible paths'' for generating counterfactual. 
These feasible paths are the shortest path distances defined via density-weighted metrics. 
It can extract counterfactuals that are \textit{coherent} with the input data distribution.
\face{} generates a graph over the data points, and the user can select the prediction, the density, also the weights, and a conditions function. 
\face{} updates the graph accordingly to these constraints and applies the shortest path algorithm to find all the data points that satisfy the requirements. 

\textbf{CFX}~\cite{albini2020relation} is a local, post-hoc, and model-specific method that generates counterfactuals explanations for Bayesian Network Classifiers. The explanations are built from relations of influence between variables, indicating the reasons for the classification. In particular, this method's main achievement is that it can find pivotal factors for the classification task: these factors, if removed, would give rise to a different classification.

\subsection{Transparent methods}
In this section we present some transparent methods, tailored for tabular data. In particular, we first present some models which output feature importance, then methods which outputs rules. 

\textbf{EBM}, Explainable Boosting Machine~\cite{nori2019interpretml} is an interpretable ML algorithm. 
Technically, \ebm{} is a variant of a Generalized Additive Model ({\scshape gam})~\cite{hastie1990generalized}, i.e.,~a generalized linear model that incorporates nonlinear forms of the predictors. 
For each feature, \ebm{} uses a boosting procedure to train the generalized linear model: it cycles over the features, in a round-robin fashion, to train one feature function at a time and mitigate the effects of co-linearity. 
In this way, the model learns the best set of feature functions, which can be exploited to understand how each feature contributes to the final prediction. 
\ebm{} is implemented by the {\scshape interpretml} Python Library\footnote{\url{https://github.com/interpretml/interpret}}. 
We trained an \ebm{} on \adult{}. 
In Figure~\ref{fig:ebm} we show a global explanation reporting the importance for each feature used by \ebm{}. 
We observe that \textit{Maritial Status} is the most important feature, followed by \textit{Relationship} and \textit{Age}. 
In Figure~\ref{fig:ebm} we show an inspection of the feature \textit{Education Number} illustrating  how the prediction score changes depending on the value of the feature.
In Figure~\ref{fig:ebm}, are also reported two examples of local explanations for \ebm{}. 
For the first record, predicted as $>50k$, the most important feature is \textit{Education Num}, which is Master for this record. 
For the second record, predicted as $<50k$, the most important feature is \textit{Relationship}. 
This feature is important for both records: in the first (husband) is pushing the value higher, while in the second (own-child) lower. 

\textbf{TED}~\cite{hind2019ted} is an intrinsically transparent approach that requires in input a training dataset in which its explanation correlates each record. Explanations can be of any type, such as rules or feature importance. 
For the training phase, the framework allows using any ML model capable of dealing with multilabel classification. 
In this way, the model can classify the record in input and correlate it with its explanation. A possible limitation of this approach is the creation of the explanations to feed during the training phase. 
{\scshape ted} is implemented in {\scshape aix360}.

\textbf{SLIPPER}~\cite{cohen1999simple} is a transparent rule learner based on a modified version of Adaboost. 
It outputs compact and comprehensible rules by imposing constraints on the rule builder. 

\textbf{LRI}~\cite{weiss2000lightweight} is a transparent rule learner that achieves good performance while giving interpretable rules as explanations. 
In {\scshape lri}, each class of the training is represented by a set of rules, without ordering. 
The rules are obtained by an induction method that weights the cumulative error adaptively without pruning. 
When a new record is considered, all the available rules are tested on it. 
The output class is the one that has the most satisfying set of rules for the record under analysis. 

\textbf{MlRules}~\cite{dembczynski2008maximum} is a transparent rule induction algorithm solving classification tasks through probability estimation. 
Rule induction is done with boosting strategies, but a maximum likelihood estimation is applied for rule generation.

\textbf{RuleFit}~\cite{Friedman2008PREDICTIVELV} is a transparent rule learner that exploits an ensemble of trees. 
As a first step, it creates an ensemble model by using gradient boosting. 
The rules are then extracted from the ensemble: each path in each tree is a rule. 
After the rules' extraction, they are weighted according to an optimization problem based on L1 regularization.

\textbf{IDS}, Interpretable Decision Sets~\cite{lakkaraju2016interpretable}, is a transparent and highly accurate model based on \textit{decision sets}.
Decision sets are sets of independent, short, accurate, and non-overlapping if-then rules. 
Hence, they can be applied independently.

\begin{table}[t]
    \footnotesize
    \caption{Comparison on the fidelity and the faithfulness metrics of different explanation methods. For every evaluation we report the mean and the standard deviation over a subset of $50$ test set records. }
    \label{tab:fidelity-faith}
    \centering
    %\scriptsize
    \begin{tabular}{|c|c||c|c||c|c||c|c|c|}
         %\hline
         %\multicolumn{5}{|c|}{\textbf{ADULT Dataset}}\\
        \hline
        \rowcolor[gray]{0.95}
         & & \multicolumn{4}{c||}{\textbf{Fidelity}} & \multicolumn{2}{c|}{\textbf{Faithfulness}}\\
        \cline{3-8}
        \multirow{-2}{*}{\cellcolor[gray]{0.95} \textbf{Dataset}} &
        \multirow{-2}{*}{\cellcolor[gray]{0.95} \textbf{Black-Box}} & \cellcolor[gray]{0.95} \lime{} & \cellcolor[gray]{0.95} \shap{} & \cellcolor[gray]{0.95} \anchor{}  & \cellcolor[gray]{0.95} \lore{} & \cellcolor[gray]{0.95} \lime{} & \cellcolor[gray]{0.95} \shap{}\\
         %\hline
         %\multicolumn{5}{|c|}{\textbf{GERMAN Dataset}}\\
         \hline
        \multicolumn{1}{|c|}{\multirow{3}{*}{\adult{}}} & LG & 0.979 & 0.613 & \textbf{0.989} & 0.984 & 0.099 (0.30) & \textbf{0.38} (0.37)\\
         & XGB & 0.977 & 0.877 & 0.978  & \textbf{0.982} & 0.030 (0.32) & \textbf{0.36} (0.49)\\ 
         & CAT & 0.96 & 0.777 & 0.988 & \textbf{0.989} & 0.077 (0.32) & \textbf{0.44} (0.37)\\
         \hline
         \multicolumn{1}{|c|}{\multirow{3}{*}{\german{}}} & LG & \textbf{0.984} & 0.910 &0.730  & 0.983 & \textbf{0.23} (0.60) & 0.19 (0.63)\\
         & XGB & \textbf{0.999} & 0.821 & 0.802 & 0.982 & 0.16 (0.26) & \textbf{0.44} (0.21)\\ 
         & CAT & 0.979 & 0.670 & 0.620 & \textbf{0.981} & 0.34 (0.33) & \textbf{0.43} (0.32)\\
         \hline
    \end{tabular}

\end{table}

\begin{table}[t]
    \footnotesize
    \centering
    \caption{Comparison on the stability metric. We report the mean and the standard deviation over a subset of $30$ test records.}
    \label{tab:lipwitch}
    %\scriptsize
    \begin{tabular}{|c|c||c|c||c|c|}
         %\hline
         %\multicolumn{5}{|c|}{\textbf{ADULT Dataset}}\\
        \hline
        \cellcolor[gray]{0.95} \textbf{Dataset} &
        \cellcolor[gray]{0.95} \textbf{Black-Box} & \cellcolor[gray]{0.95} \lime{} & \cellcolor[gray]{0.95} \shap{} & \cellcolor[gray]{0.95} \anchor{} & \cellcolor[gray]{0.95} \lore{}\\
         %\hline
         %\multicolumn{5}{|c|}{\textbf{GERMAN Dataset}}\\
         \hline
        \multicolumn{1}{|c|}{\multirow{3}{*}{\adult{}}} & LG & 24.37 (2.74) & 1.52 (4.49) & 22.36 (8.37)  & 21.76 (11.80)\\
         & XGB & 10.16 (6.48) & 2.17 (2.18) & 26.53 (13.08) &  30.01 (20.52) \\ 
         & CAT & 0.35 (0.43)  & 0.03 (0.01) & 6.51 (4.40) & 27.80 (70.05)\\
         \hline
         \multicolumn{1}{|c|}{\multirow{3}{*}{\german{}}} & LG & 18.87 (0.73) & 19.01 (23.44) & 101.07 (62.75)  & 622.12 (256.70)\\
         & XGB & 26.08 (14.50) & 38.43 (30.66) & 121.40 (98.43) &  725.81 (337.26) \\ 
         & CAT & 2.49 (9.91) &  15.92 (10.71) & 123.79 (76.86) & 756.70 (348.21)\\
         \hline
    \end{tabular}
    
\end{table}

\subsection{Quantitative Comparison}
\label{sec:tabular_experiment}
We validated explanation models by considering the two most important metrics in the context of tabular data: \textit{fidelity}, and the \textit{stability}. 
In particular, we evaluated \lime{}, \shap{}\footnote{Since \shap{} is not training a local surrogate, we evaluate the fidelity of \shap{} by learning a classifier on the sum of the \shap{}'s values.}, \anchor{} and \lore{}. 
The results of the fidelity are reported in Table~\ref{tab:fidelity-faith}. 
The fidelity values are relatively high for all the methods highlighting that the local surrogate models are good at mimicking their black-box models.
Regarding the feature importance-based models, \lime{} shows higher values of fidelity w.r.t.~\shap{}, especially for \adult{}. In particular, \shap{} has lower values for the CAT models (both \german{} and \adult{}), suggesting that it may be not good in explaining this kind of ensemble models. 
Concerning the rule-based models, the fidelity is high for both of them. However, we remark that \anchor{} shows lower values of fidelity for the CAT model for \german{}, a behavior which is similar to the one of \shap{}.
We compared \lime{} and \shap{} on faithfulness and monotonicity. 
Overall, we did not find any model to be monotonic, and hence we do not report any results. 
The results for the faithfulness are reported in Table~\ref{tab:fidelity-faith}. 
For \adult{}, the faithfulness is quite low, especially for \lime{}. 
The model with the highest faithfulness is CAT explained by \shap{}. 
Regarding \german{}, instead, the values are higher, highlighting a better faithfulness overall. 
However, also for this dataset \shap{} has a better faithfulness w.r.t.~\lime{}. 
In Table~\ref{tab:lipwitch} are reported the results obtained from the analysis on the stability. 
For this metric, a high value means that the model presents high instability, meaning that we can have quite different explanations for similar inputs. 
None of the methods is remarkably stable according to this metric.

\begin{table}[t]
    \footnotesize
    \caption{Explanation runtime expressed in seconds for explainers of tabular classifiers approximated as order of magnitude.}
    \label{tab:runtime_tabular}
    \centering
    %\scriptsize
    \begin{tabular}{|c|c|c|c|c|c|c|c|c|}
        \hline
        \cellcolor[gray]{0.95} \textbf{Dataset} 
        & \cellcolor[gray]{0.95} \textbf{Black-Box} 
        & \cellcolor[gray]{0.95} \lime{} 
        & \cellcolor[gray]{0.95} \shap{} 
        & \cellcolor[gray]{0.95} \dalex{} 
        & \cellcolor[gray]{0.95} \anchor{} 
        & \cellcolor[gray]{0.95} \lore{}
        & \cellcolor[gray]{0.95} {\scshape skoperule}\\
         \hline
        \multicolumn{1}{|c|}{\multirow{3}{*}{\adult{}}} 
         & LG & 0.1 & \textbf{0.001} & 90 & 2 & 15 & 100 \\
         & XGB & \textbf{0.1} & 0.2 & 108 & 5 & 50 & - \\
         & CAT & \textbf{0.2} & 3 & 110 3 & 35 & - & -\\
         \hline
         \multicolumn{1}{|c|}{\multirow{3}{*}{\german{}}} 
         & LG & 0.007 & \textbf{0.0008} & 0.8 & 2 & 2 & 70 \\
         & XGB & 0.03 & \textbf{0.002} & 2 & 2 & 4 & - \\
         & CAT & 0.03 & \textbf{0.002} & 1 & 2 & 6 & - \\
         \hline
    \end{tabular}
    
\end{table}

\subsubsection*{\textbf{Runtime Analysis}}
Table~\ref{tab:runtime_tabular} shows the explanation runtime approximated as order of magnitude.
Overall, feature importance explanation algorithms are faster w.r.t. the rule-based ones. In particular, \shap{} is the most efficient, followed by \lime{}. We remark that the computation time of \lore{} depends on the number of neighbors to generate exploiting a genetic algorithm (in this case, we considered $1000$ samples). \anchor{}, instead, requires a minimum precision as well as {\scshape skoperule} (we selected min precision of $0.40$).

\subsubsection*{\textbf{Discussion}}
In the context of tabular data, many explainable methods have been proposed. In particular, the most explored area is feature importance-based explanators, such as \lime{} and \shap{}. These methods provide an importance value for each feature in the input. It is suitable for domain experts who know the meaning of the features employed. However, it may be too difficult for a common end-user to understand, especially when obtaining such importance values is complex. In contrast, rule-based explanations, prototypes, and counterfactuals are more suitable for the common end-user due to their logical structure and the similarity by example they exploit. This is particularly true in decision rules correlated by counterfactual ones, like in \lore{}. The end-user can understand why she received that outcome, but she also has a suggestion about what to change to achieve another classification. However, fewer methods are proposed in this context w.r.t.~feature importance explanations. 
In particular, the majority of rule and prototype-based explanators are intrinsic. 
For the few post-hoc ones, on average, they require more time to provide an explanation w.r.t.~feature importance ones.
Regarding the post-hoc prototype-based models, there are some interesting approaches, but there is no code for them, highlighting that they are still in an early stage of development.
During the past few years, counterfactuals have witnessed a particularly great interest.
Overall, even if the rules, prototypes, and counterfactuals seem to be the best solution, there are still several open questions and challenges in this research area such as improving the efficiency and the accuracy of these explanation algorithms as well as considering the constraints of the domain in which the model is being employed.

%% file: 5_image.tex
\begin{table}[t]
    \small
    \caption{Explainers for black-boxes classifying image data sorted by explanation type: Saliency Maps (SM), Concept Attributions (CA), Counterfactuals (CF), and Prototypes (PR). For every method is indicated if is possible it for images (IMG) only, or for ANY type of data, if it is an Intrinsic (IN) or a Post-Hoc (PH) model, Local (L) or Global (G), and if it is model Agnostic (A) or model-Specific (S).}
   % \end{varwidth}
    \label{tab:image_summary}
    \centering
    \setlength{\tabcolsep}{1mm}
  \begin{tabular}{cccccccccc} 
    \hline
    \rotatebox[origin=c]{0}{\textbf{Type}} & \rotatebox[origin=c]{0}{\textbf{Name}}       & \rotatebox[origin=c]{0}{\textbf{Ref.}}  & \rotatebox[origin=c]{0}{\textbf{Authors}} & \rotatebox[origin=c]{0}{\textbf{Year}} & \rotatebox[origin=c]{0}{\textbf{Data Type}}  & \rotatebox[origin=c]{0}{\textbf{IN/PH}} & \rotatebox[origin=c]{0}{\textbf{G/L}} & \rotatebox[origin=c]{0}{\textbf{A/S}}  &
    \rotatebox[origin=c]{0}{\textbf{Code}}  \\ 
    
    \hline
                         & \shap 	             & \cite{lundberg2017unified} 		& Lundberg et al.           & 2007  & ANY & PH & L & A & \href{https://github.com/slundberg/shap}{link}\\
    \rowcolor{gray!15}   
    \cellcolor{white}    & \lime                 & \cite{ribeiro2016should} 		& Ribeiro et al.            & 2016  & ANY & PH & L & A & \href{https://github.com/marcotcr/lime}{link}\\
                         & \elrp  	             & \cite{bach2015pixel} 			& Bach et al.               & 2015  & ANY & PH & L & S & \href{https://github.com/marcoancona/DeepExplain}{link}\\
    \rowcolor{gray!15}
    \cellcolor{white}    & \intgrad              & \cite{sundararajan2017axiomatic} & Sundararajan et al.       & 2017  & ANY & PH & L & S & \href{github.com/marcoancona/DeepExplain}{link}\\
                         & \deeplift             & \cite{shrikumar2017learning} 	& Shrikumar et al.          & 2017  & ANY &	PH & L & S & \href{https://github.com/marcoancona/DeepExplain}{link}\\
    \rowcolor{gray!15}
    \cellcolor{white}    & \smoothgrad           & \cite{smilkov2017smoothgrad} 	& Smilkov et al.            & 2017  & IMG & PH & L & S & \href{https://github.com/PAIR-code/saliency}{link}\\
                         & \xrai                 & \cite{kapishnikov2019xrai} 		& Kapishnikov et al.        & 2019  & ANY & PH & L & S & \href{https://github.com/PAIR-code/saliency}{link}\\
    \rowcolor{gray!15}
    \cellcolor{white}    & \gradcam              & \cite{selvaraju2017grad} 		& Selvaraju et al.          & 2017  & IMG &	PH & L & S & \href{https://github.com/adityac94/Grad_CAM_plus_plus}{link}\\
                         & \gradcamplus          & \cite{chattopadhay2018grad} 		& Chattopadhay et al.       & 2018  & IMG &	PH & L & S & \href{https://github.com/adityac94/Grad_CAM_plus_plus}{link}\\
    \rowcolor{gray!15}
    \multirow{-10}{*}{SM}
    \cellcolor{white}    & \rise                 & \cite{petsiuk2018rise} 		    & Petsiuk et al.            & 2018  & IMG & PH & L & S & \href{https://github.com/eclique/RISE}{link}\\
%                         & {\scshape kl-lime}    & \cite{peltola2018local}       	& Peltola et al.            & 2018  & IMG & PH & L & A &  -\\
%    \rowcolor{gray!15}
%    \cellcolor{white}    & {\scshape d-lime}     & \cite{zafar2019dlime}     		& Zafar et al.              & 2019  & IMG & PH & L & A &  \href{https://github.com/rehmanzafar/dlime_experiments}{link}\\
%                         & {\scshape q-lime}     & \cite{bramhall2020qlime}         & Bramhall et al.           & 2020  & IMG & PH & L & A &  -\\
%    \rowcolor{gray!15}
    
%    \cellcolor{white}    & {\scshape mps-lime}   & \cite{shi2020modified}     	    & Shi et al.                & 2020  & IMG & PH & L & A &  - \\
    
    \cellcolor{gray!15}  & \tcav                 & \cite{kim2018interpretability} 	& Kim et al.                & 2018  & IMG & PH & L & A &  \href{https://github.com/tensorflow/tcav}{link}\\
    \rowcolor{gray!15}
                         & \ace                  & \cite{ghorbani2019towards}		& Ghorbani et al.           & 2019  & IMG & PH & G & A &  \href{https://github.com/amiratag/ACE}{link}\\
    \cellcolor{gray!15}  & {\scshape conceptshap}& \cite{yeh2020completeness}       & Yeh et al.                & 2020  & IMG & PH & G & A &  -\\
    \rowcolor{gray!15}
    \multirow{-4}{*}{\cellcolor{gray!15}CA}
                         & {\scshape cace}       & \cite{goyal2019explaining} 	    & Goyal et al.              & 2019  & IMG & IN & G & A &  -\\
    \cellcolor{white}    & \cem                  & \cite{dhurandhar2018explanations}& Dhurandhar, Amit, et al.  & 2018  & IMG & PH & L & A &  \href{https://github.com/SeldonIO/alibi}{link}\\
    \rowcolor{gray!15}
    \cellcolor{white}    & \abele                & \cite{guidotti2020explaining}    & Guidotti et al.           & 2020  & IMG & PH & L & A &  \href{https://github.com/riccotti/ABELE}{link} \\
    \cellcolor{white}
                         & {\scshape l2x} 		 & \cite{l2x} 				        & Chen et al. 		        & 2018  & ANY & PH & L & A & \href{https://github.com/Jianbo-Lab/L2X}{link} \\
    \rowcolor{gray!15}
    \multirow{-3}{*}{CF}
    \cellcolor{white}    & {\scshape guided proto}                     & \cite{van2019interpretable} 	    & Van Looveren et al.       & 2019  & IMG & PH & L & A &  \href{https://github.com/SeldonIO/alibi}{link}\\
    \cellcolor{gray!15}  & \mmdcritic            & \cite{mmdcritic} 			    & Kim et al.                & 2016  & ANY & IN & G & A &  \href{https://github.com/BeenKim/MMD-critic}{link}\\
    \rowcolor{gray!15}
    \cellcolor{gray!15}  & - & \cite{koh2017understanding} & Koh et al.            & 2017  & ANY & PH & L & A &  \href{https://github.com/kohpangwei/influence-release}{link}\\
    \multirow{-3}{*}{\cellcolor{gray!15}PR} 
                         & {\scshape protonet}   & \cite{chen2019looks}		        & Chen et al.               & 2019  & IMG & IN & G & S & \href{https://github.com/cfchen-duke/ProtoPNet}{link}\\
    \hline
    \end{tabular}%
\end{table}

\section{Explanations for Image Data}
\label{sec:image}
This section presents the solutions in the state of the art, proposing explanations for decision systems acting on image data. 
In particular, we distinguish the following types of explanations: 
\textit{Saliency Maps} (SM, Section~\ref{sec:saliency}, \textit{Concept Attribution} (CA, Section~\ref{sec:concept_attribution}), \textit{Prototypes} (PR, Section~\ref{sec:proto_image}) and \textit{Counterfactuals} (CF, Section~\ref{sec:counter_image}). 
Table~\ref{tab:image_summary} summarizes and categorizes the explanation methods acting on image data. 
For the experiments, we considered three datasets\footnote{
    \mnist{}: \url{http://yann.lecun.com/exdb/mnist/},
    \cifar{}: \url{https://www.cs.toronto.edu/~kriz/cifar.html}, and
    \imagenet{}: \url{http://image-net.org/}: 
}: \mnist{}, \cifar{} in its 10 class flavor and \imagenet{}. 
We choose these datasets because they are the most utilized, and we have different types of classes with various image dimensions. 
On these three datasets, we trained the models most used in literature to evaluate the explanation methods: for \mnist{} and \cifar{} we a CNN with two convolutions and two linear layers, while for \imagenet{} the VGG16 network~\cite{simonyan2014very}.

\subsection{Saliency Maps}
\label{sec:saliency}
A \textit{Saliency Map (SM)} is an image in which a pixel's brightness represents how salient the pixel is. 
Formally, a SM is modeled as a matrix $S$ which dimensions are the sizes of the image we want to explain, and the values $s_{ij}$ are the saliency values of the pixels $ij$. 
The greater the value of $s_{ij}$ the bigger is the saliency of that pixel. 
To visualize SM, we can use a divergent color map for example, ranging from red to blue. 
A positive value (red) means that the pixel $ij$ has contributed positively to the classification, while a negative one (blue) means that it has contributed negatively. 
There are two methods for creating SMs.
The first one assigns to every pixel a saliency value. The second one segments the image into different pixel groups and then assign a saliency value for each group. 

\begin{figure}[t]
    \begin{center}
        \centerline{\includegraphics[width=.85\textwidth]{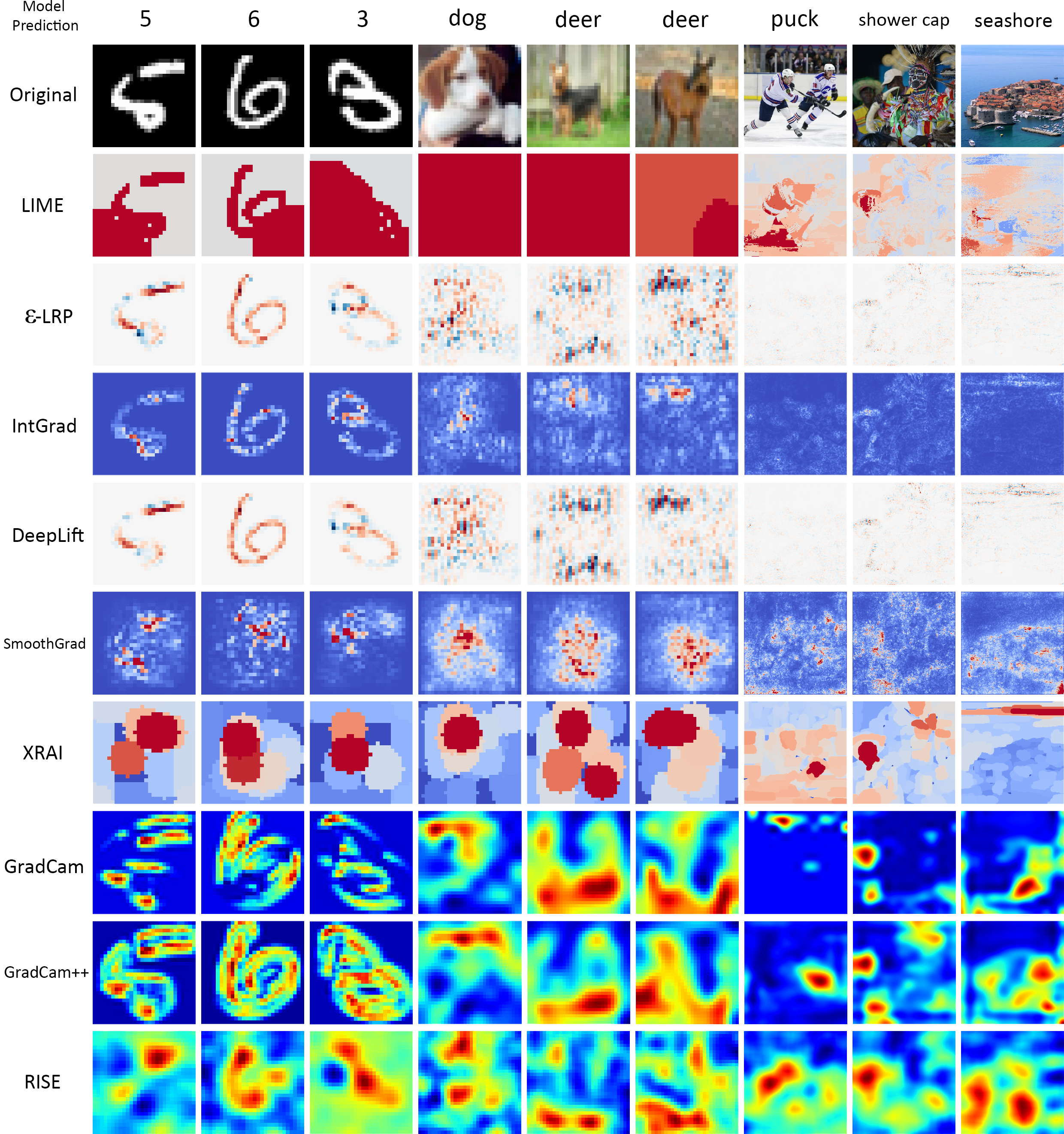}}
        \caption{Examples of saliency maps obtained with the algorithm exposed in Section 5.1 on various datasets. The first row are the original images of the dataset and on top of them we have the predicted class from the original model.}
        \label{fig:saliency_visual_comparison}
    \end{center}
\end{figure}

\textbf{LIME}, already presented in Section~\ref{sec:tabular}, can also be used to retrieve SM for classifiers working on images. 
For images, the perturbation is done by segmentation. 
More in detail, \lime{} divides the input image into segments called \textit{superpixels}. 
Then it creates the neighborhood by randomly substituting the super-pixels with a uniform, possibly neutral, color. 
This neighborhood is then fed into the black-box, and a sparse linear model is learned on top. 
An example of such a super-pixel explanation is shown in Figure~\ref{fig:saliency_visual_comparison}. 
The super-pixel segmentation is critical to obtain a good explanation. 
For small resolution images, the segmentation in \lime{} does not work out of the box, resulting in the algorithm selecting all the image as a super-pixel. 
To obtain a decent result, the user needs to tune the segmentation parameters. 
Recently, many research improved and extended \lime{}~\cite{shi2020modified,peltola2018local,zafar2019dlime,bramhall2020qlime}~\footnote{{\scshape dlime}: \url{https://github.com/rehmanzafar/dlime_experiments}}

\textbf{$\epsilon$-LRP}, Layer-wise Relevance Propagation~\cite{bach2015pixel} is a model specific method which produce post-hoc local explanations for any type of data.
\elrp{} explains the classifier’s decisions by decomposition. 
The \elrp{} redistribution process was introduced for feed-forward neural networks~\cite{arras2017explaining}. 
Mathematically, it redistributes the prediction $y$ backwards using local redistribution rules until it assigns a relevance score $R_i$ to each pixel value.
Let $a_i$ be the neuron activations at layer $l$, $R_j$ be the relevance scores associated to the neurons at layer $l + 1$ and $w_{ij}$ be the weight connecting neuron $i$ to neuron $j$.
The simple \elrp{} rule redistributes relevance from layer $l + 1$ to layer $l$ is: $R_i = \sum_j{\frac{a_iw_{ij}}{\sum_i{a_iw_{ij}}+\epsilon}}R_j$ 
where a small stabilization term $\epsilon$ is added to prevent division by zero. 
Intuitively, this rule redistributes relevance proportionally from layer $l + 1$ to each neuron in $l$ based on the connection weights.
The final explanation is the relevance of the input layer.
Figure~\ref{fig:saliency_visual_comparison} shows some examples of \elrp{} in the third row. 
As with all the pixel-wise explanation method, the algorithm works very well on \mnist{} while it is difficult to address larger images.
A variant of \elrp{} is {\scshape spray}~\cite{lapuschkin2019unmasking} which builds a specrtal clustering on top of the local instance-based \elrp{} explanations. 
Similar work is done in~\cite{li2019beyond}: it starts with the \elrp{} of the input instance and finds the LRP attribution relevance for a single input of interest $x$. 

\textbf{INTGRAD}, Integrated Gradient~\cite{sundararajan2017axiomatic}, is a model-specific method that produces post-hoc local explanations for any type of data.
\intgrad{} utilizes the gradients of a black-box along with the sensitivity techniques like \elrp{}. 
For this reason, it can be applied only on differentiable models. 
Formally, given $b$ and $x$, and let $x'$ be the baseline input.\footnote{The baseline $x'$ is generally chosen as a zero matrix or vector. 
For example, for the image domain, the baseline is generally a black or a white image.},
\intgrad{} constructs a path from $x'$ to $x$ and computes the gradients of points along the path.
For example, with images, the points are taken by overlapping $x$ on $x'$ and gradually modifying the opacity of $x$.
Integrated gradients are obtained by cumulating the gradients of these points. 
Formally, the integrated gradient along the $i^{th}$ dimension for an input $x$ and baseline $x'$ is defined as follows. 
Here, $\partial b(x)/\partial x_i$ is the gradient of $b(x)$ along the $i^{th}$ dimension. 
The equation for computing the scores is: $e_i(x) = (x_i-x'_i) \int_{\alpha=0}^{1} \tfrac{\partial b(x' + \alpha(x-x'))}{\partial x_i  }~d\alpha$.
An example of \intgrad{} explanations is in Figure~\ref{fig:saliency_visual_comparison}. 
The saliency maps obtained tend to have uniform pixels than \elrp{}. 
As shown before, \elrp{} highlights that when predicting the ``deer'' the most salient regions are in the background.
However, an arbitrary choice of baselines could cause issues. 
For example, a black baseline image could cause the method to lower the importance of black pixels in the source image. 
This problem is due to the difference between the image's pixel and the baseline ($x_i-x_i'$) present in the integral equation.
\textit{Expected Gradients}~\cite{erion2019learning} tries to overcome this problem by averaging \intgrad{} to different baselines.

\textbf{DEEPLIFT}~\cite{shrikumar2017learning}, is a model-specific and data-agnostic explainer which produces post-hoc local explanations. 
It computes SMs in a backward fashion similarly to \elrp{}, but it uses a baseline reference like in \intgrad{}. 
\deeplift{} uses the slope, instead of the gradients, which describes how the output $y=b(x)$ changes as the input $x$ differs from a baseline $x'$. 
Like \elrp{}, an attribution value $r$ is assigned to each unit $i$ of the neural network going backward from the output $y$. 
This attribution represents the relative effect of the unit activated at the original network input $x$ compared to the activation at the baseline reference $x'$. 
\deeplift{} computes the starting values of the last layer $L$ by the difference between the output of the input and baseline. 
Then, it uses the following recursive equation to compute the attribution values of layer $l$ using the attributions of layer $l+1$ to obtain the values of the starting layer:
$
    r_i^{(l)} = \sum_j \frac{a_{ji} - a'_{ji}}{\sum_{i} a_{ji} - \sum_{i} a'_{ji}} r_j^{(l+1)}, 
    a_{ji}  = w^{(l+1,l)}_{ji} x_i^{(l)}, 
    a'_{ji} = w^{(l+1,l)}_{ji} x_i^{'(l)}
$
where $w_{ij}^{l+1,l}$ are the weights of the network between the layer $l$ and the layer $l+1$, and $a$ are the activation values.
As for \intgrad{}, picking a baseline is not trivial and might require domain experts.
The SMs obtained with \deeplift{} are very similar to those obtained with \elrp{} (Figure~\ref{fig:saliency_visual_comparison}). 

\begin{figure}[t]
    \begin{center}
        \centerline{\includegraphics[trim={0 0 0 2.5cm},clip,width=0.9\textwidth]{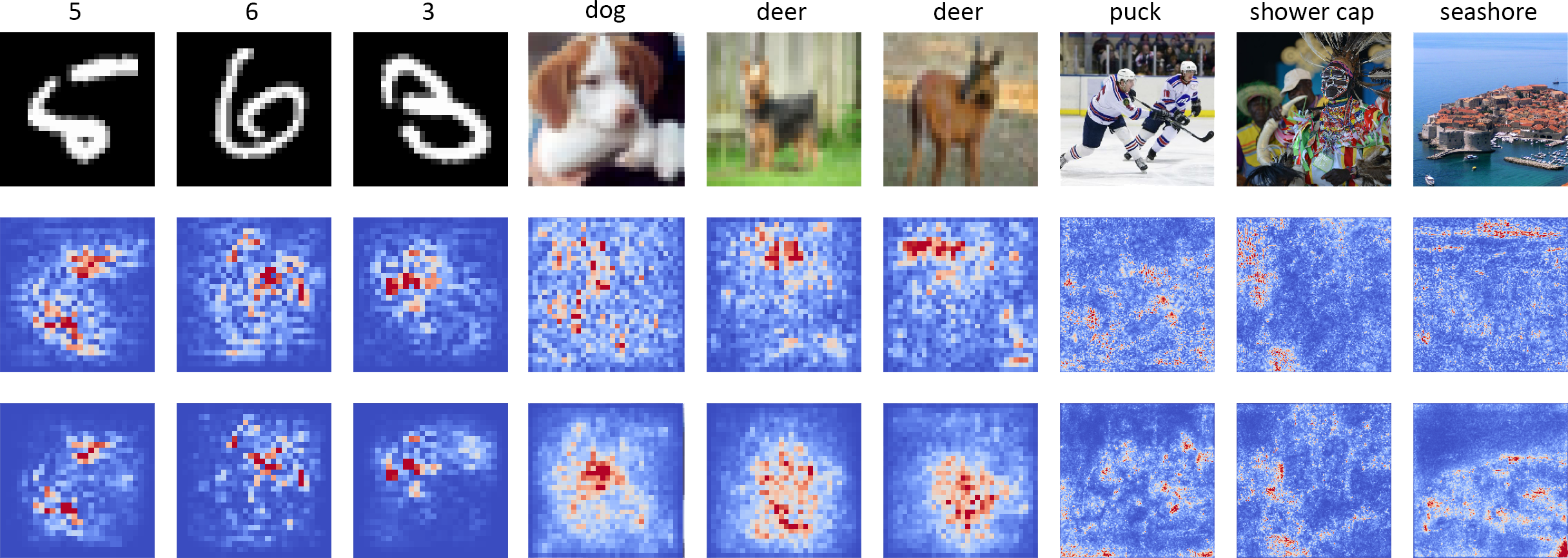}}
        \caption{Visual Comparison of saliency maps obtained by taking the gradient of the output $y$ w.r.t.~the input image $x$ (center) and \smoothgrad{} (bottom). On the three image in the center the saliency map changes drastically. On all three cases is focusing to the subject of the image completely changing original values. This is true also for the seashore image on the far right.}
        \label{fig:smoothgrad}
    \end{center}
\end{figure}

\textbf{SMOOTHGRAD}~\cite{smilkov2017smoothgrad} is a post-hoc model-specific and data-agnostic explanation method. 
A SM tends to be noisy, especially for pixel-wise saliency maps. 
\smoothgrad{} tries to overcome this problem by \textit{smoothing} the noisiness in the SMs. 
Usually, a SM is created directly on the gradient of the model's output signal w.r.t.~the input $\partial y/\partial x$. 
\smoothgrad{} augments this process by smoothing the gradients with a Gaussian noise kernel. 
It takes $x$, applies Gaussian noise to it, and retrieve the SM for every perturbed image, using the gradient. 
The final SM is an average of these.
Formally, given a saliency method $f(x)$ which produces a saliency map $s$, its smoothed version $\hat{f}$ can be expressed as: $ \hat{f} = \frac{1}{n} \sum_1^n f(x + \mathcal{N}(0, \sigma^2))$ where $n$ is the number of samples, and $\mathcal{N}(0, \sigma^2)$ is the Gaussian noise. 
In~\cite{adebayo2018sanity,adebayo2020debugging} are shown some weaknesses of \smoothgrad{}: people tend to evaluate SMs on what they are expected to see. 
For example, in a bird image, we want to see the shape of a bird. 
However, this does not mean that this is what the network is looking at. 
Figure~\ref{fig:smoothgrad} highlights this problem.
We obtained the SMs taking the gradient of the output w.r.t.~the input, and then we used \smoothgrad{}. We observe that the SMs completely changed their behavior, moving in direction of the subject. 

\begin{figure}[t]
    \centering
       \includegraphics[width=.85\linewidth]{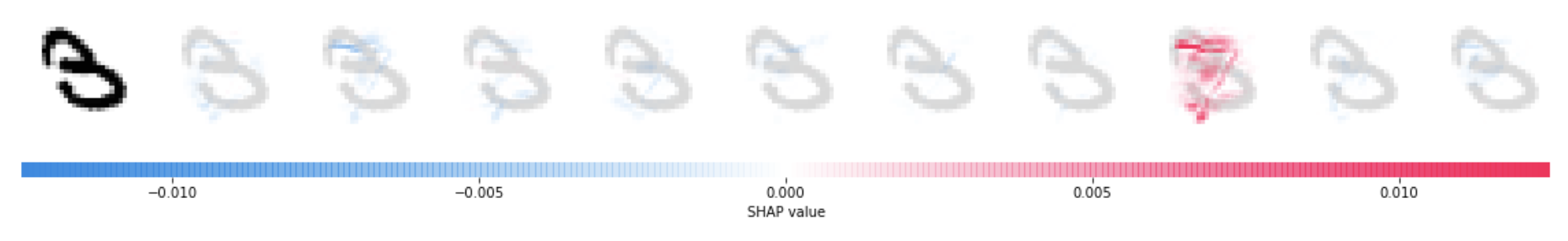}
       \includegraphics[width=.3\linewidth]{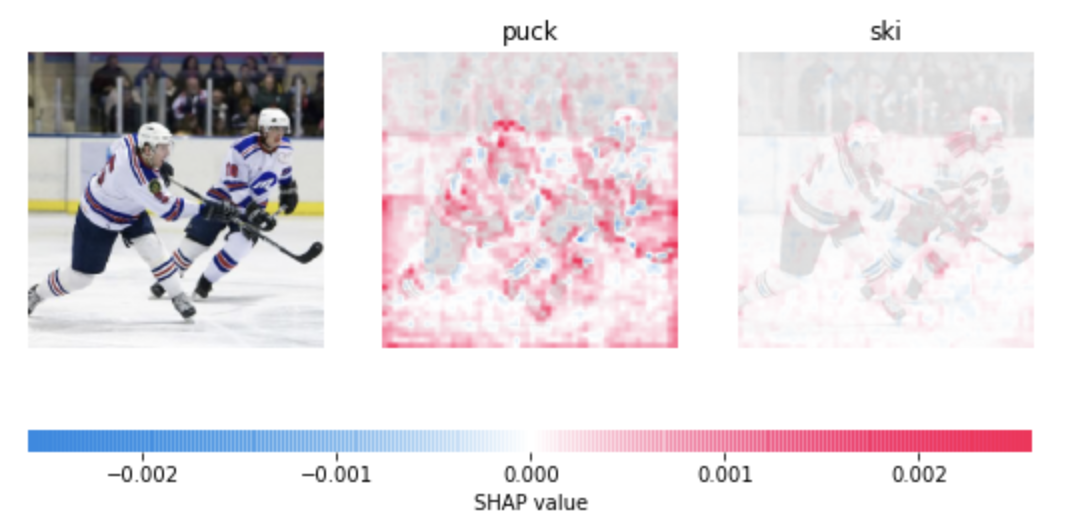}
       \includegraphics[width=.3\linewidth]{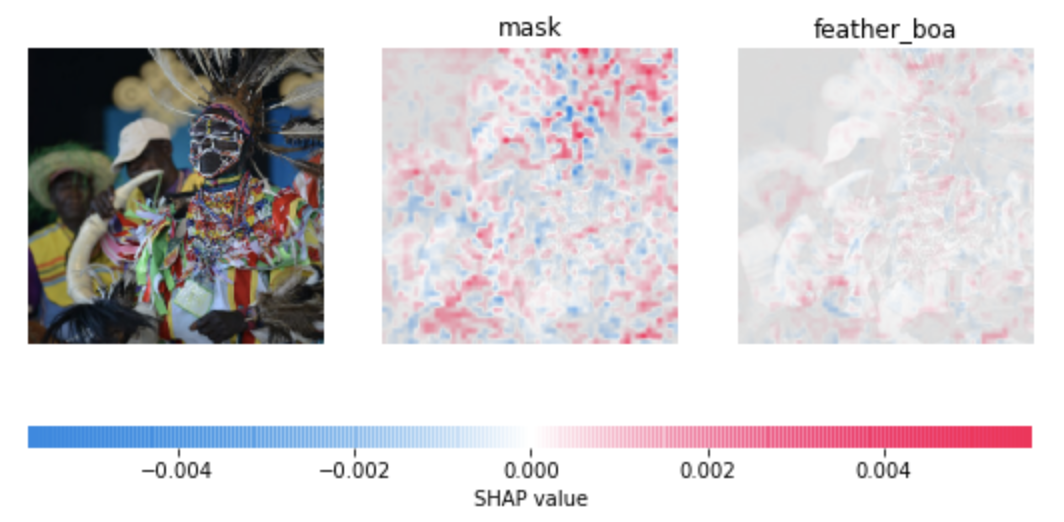}
       \includegraphics[width=.3\linewidth]{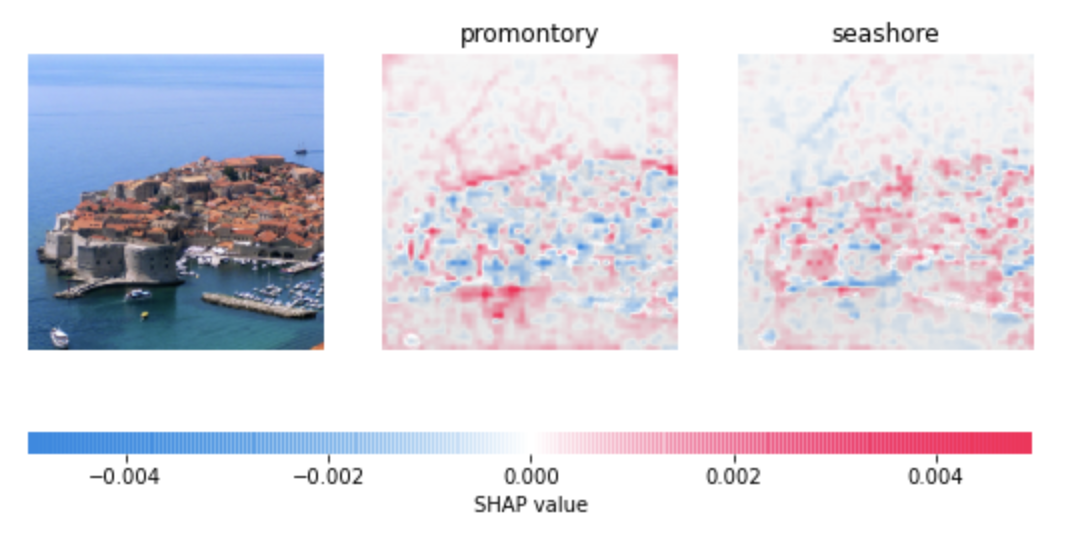}
    \caption{(Top) Explanations of \deepshap{} on \mnist{}. (Bottom) Explanations of \gradshap{} on \imagenet{}.}
\label{fig:saliency_shap}
\end{figure}

\textbf{SHAP}, presented in Section \ref{sec:tabular}, has two explanators that can be employed for deep networks tailored for image classification: \deepshap{} and \gradshap{}. 
\deepshap{} is a high-speed approximation algorithm for \shap{} values in deep learning models that builds on a connection with \deeplift{}. 
The implementation is different from the original \deeplift{} by using as baseline a distribution of background samples instead of a single value and using shapley equations to linearise non-linear components of the black-box such as max, softmax, products, divisions, etc. 
\gradshap{}, instead, is based on \intgrad{} and \smoothgrad{}~\cite{sundararajan2017axiomatic,smilkov2017smoothgrad}. 
\intgrad{} values are a bit different from \shap{} values, and require a single reference value to integrate from. As an adaptation to make them approximate \shap{} values, \gradshap{} reformulates the integral as an expectation and combines that expectation with sampling reference values from the background dataset as done in \smoothgrad{}. 
We tested both \deepshap{} and \gradshap{} experimentally and the results are shown in Figure~\ref{fig:saliency_shap}. 
\deepshap{} outputs a saliency map explaining every class of the input image. 
\gradshap{} instead produce a pixel-wise saliency map similar to those shown before. 

\textbf{XRAI}~\cite{kapishnikov2019xrai} is based on \intgrad{} and inherits its properties. 
Differently from \intgrad{}, \xrai{} first over-segments the image. It iteratively tests each region's importance, fusing smaller regions into larger segments based on attribution scores. 
It is divided into three steps: \textit{segmentation}, \textit{get attribution}, and \textit{selecting regions}.
The segmentation is repeated several times with different segments to reduce the dependency on image segmentation. 
For attribution, \xrai{} uses \intgrad{} with black and white baselines averaged. 
Finally, to select regions, \xrai{} leverages the fact that, given two regions, the one that sums to the more positive value should be more important to the classifier. 
From this observation, \xrai{} starts with an empty mask, then selectively adds the regions that yield the maximum gain in the total attributions per area. 
The saliency maps obtained from \xrai{} are very different from those already presented. 
Figure~\ref{fig:saliency_visual_comparison} shows some examples. 
As all the segmentation methods \xrai{} performs at its best when having high-resolution images. 
However, it still obtains good results on low-resolution images.

\textbf{GRADCAM}~\cite{selvaraju2017grad} is a model-specific post-hoc local explainer for image data. 
It uses the gradient information flowing into the last convolutional layer of a CNN to assign saliency values to each neuron for a particular decision. 
Convolutional layers naturally retain spatial information in fully-connected layers, so we can expect the last convolutional layers to have the best compromise between high-level semantics and detailed spatial information. 
To create the SM, \gradcam{} takes the feature maps created at the last layer of the convolutional network $a$. 
Then, it computes the gradient of an output of a particular class $y^c$ for every feature map activations $k$, i.e.,~ $\partial y^c/\partial a^k$. 
This equation returns a tensor of dimensions $[k,v,u]$ where $k$ is the number of features maps and $u,v$ are height and width of the image. 
\gradcam{} compute the saliency value for every feature maps by pooling the dimensions of the image. The final heatmap is calculated as a weighted sum of these values.
Notice that this results in a coarse heatmap of the same size as the convolutional feature maps. 
An up-sampling technique is applied to the final result to produce a map of the initial image dimension. 
From Figure~\ref{fig:saliency_visual_comparison} is clear that these coarse grain heatmap style are very characteristic of \gradcam{}. 
These heat maps highlight very different parts of the image compared to the other methods. 

\textbf{GRADCAM++}~\cite{chattopadhay2018grad} extends \gradcam{} solving some related issues.
The spatial footprint in an image is essential for \gradcam{}'s visualizations to be robust. 
Hence, if there are multiple objects with slightly different orientations or views, different feature maps may be activated with differing spatial footprints and the one with lesser footprints fade away in the final sum.
\gradcamplus{} fix this problem by taking a weighted average of the pixel-wise gradients.
In particular, \gradcamplus{} reformulates \gradcam{} by explicitly coding the structure of the weights $\alpha_k^c$ as: $\alpha_k^c = \sum_{i}\sum_{j}w_{ij}^{kc}\cdot\text{ReLU}\left(\partial y^c/\partial a_{ij}^k\right) \label{weighted version}$ where \textit{ReLU} is the Rectified Linear Unit activation function, and $w_{ij}^{kc}$ are the weighting co-efficients for the pixel-wise gradients for class $c$ and convolutional feature map $a^k$. 
The idea is that $w_k^c$ captures the importance of a particular activation map $a^k$, and positive gradients are preferred to indicate visual features that increase the output neuron's activation rather than those that suppress the output neuron's activation.

\textbf{RISE}~\cite{petsiuk2018rise} is a model-agnostic method which produces post-hoc local explanations on image data. 
To produce a saliency map for an image $x$, \rise{} generate $N$ random mask $M_i \in [0,1]$ from Gaussian noise.
The input image $x$ is element-wise multiplied with these masks $M_i$, and the result is fed to the base model. The saliency map is obtained as a linear combination of the masks $M_i$ with the predictions from the black box corresponding to the respective masked inputs.
The intuition behind this is that $b(x \odot M_i)$ is high when pixels preserved by mask $M_i$ are essential.

\subsubsection*{\textbf{Qualitative and Quantitative Comparison of Saliency Maps}}
In Figure~\ref{fig:saliency_visual_comparison}, we report the SMs obtained for every method tested.
The segmentation used by \lime{} is very poor with small images as it results in super-pixels big as the whole image in some cases. 
On the other hand, those produced by \xrai{} are much more clear. 
For the majority of images, the SMs are very similar among those returned by the various explainers but we can observe conflicts. 
For instance, in \cifar{} we can assume that the background is useful to predict the class deer, but we do not know-how. 
Some explainers highlight the top background while other the bottom background, so it is difficult to understand. 
Moving to bigger images, these conflicts become more evident. 
Let us look at the ice hockey image. 
The class in the dataset here is ``puck'': the hockey disk. 
\lime{} highlights the ice as important, while other methods (\xrai{} and \gradcamplus{}) highlight the stick of the player. 
\gradcam{} highlights the fans while \rise{} the hockey player. 
Thus, for the same image, we can obtain very different explanations.  
When moving to the second image from \imagenet{} (the mask), we can observe that all the methods capture the same pattern. 
A straw hat in the background triggered the class ``shower cap'' while the correct one was ``mask''. 
In the ``seashore'' of \imagenet{}, we have an island in the sea. 
The top three predicted classes are: seashore ($0.91$), promontory ($0.04$) and cliff ($0.01$). 
Half of the tested methods like \lime{}, \smoothgrad{}, \rise{}, and \gradcam{} was fooled that the promontory is important to the class ``seashore''. 
We can conclude that SMs are very fragile when we have multiple classes in the image, even if these classes has very low predicted probability.

\begin{figure}[t]
    \begin{center}
        \centerline{\includegraphics[width=.9\textwidth]{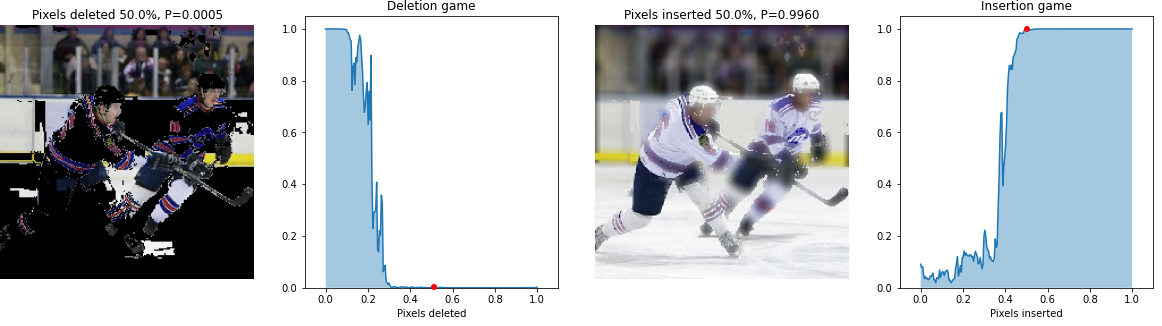}}
        \caption{Example of Insertion (on the left) and Deletion (on the right) metric computation performed on \lime{} and the hockey image. The area under the curve is 0.2156 for deletion and 0.5941 for Insertion.}
        \label{fig:insertion_deletion_single}
    \end{center}
\end{figure}

\begin{table}[t]
\centering
\caption{Insertion (left) and deletion (right) metrics expressed as AUC of accuracy vs. percentage of removed/inserted pixels.}
\label{tab:insertion_deletion_tab}
\footnotesize
\begin{minipage}{.49\linewidth}
    \begin{tabular}{|c|c|c|c|}
        \hline
        \rowcolor{gray!15}
         & \mnist{} &  \cifar{} & \imagenet{}\\
        \hline
        \lime{}        & 0.807 (0.14) & 0.41 (0.21) & 0.34 (0.25)\\
        %\hline
        \elrp{}        & 0.976 (0.02) & 0.56 (0.20) & 0.28 (0.19)\\
        %\hline
        \intgrad{}     & 0.973 (0.03) & 0.50 (0.22) & 0.27 (0.23)\\
        %\hline
        \deeplift{}    & 0.976 (0.02) & 0.57 (0.20) & 0.28 (0.19)\\
        %\hline
        \smoothgrad{}  & \textbf{0.979} (0.03) & 0.55 (0.23) & 0.34 (0.26)\\
        %\hline
        \xrai{}        & 0.956 (0.04) & 0.58 (0.21) & 0.40 (0.26)\\
        %\hline
        \gradcam{}     & 0.941 (0.04) & 0.57 (0.20) & 0.21 (0.19)\\
        %\hline
        \gradcamplus{} & 0.941 (0.04) & 0.52 (0.22) & 0.32 (0.26)\\
        %\hline
        \rise{}        & 0.978 (0.03) & \textbf{0.61} (0.21) & \textbf{0.50} (0.26)\\
        \hline
    \end{tabular}
\end{minipage}
\begin{minipage}{.49\linewidth}
    \begin{tabular}{|c|c|c|c|}
        \hline
        \rowcolor{gray!15}
          & \mnist{} &  \cifar{} & \imagenet{}\\
        \hline    
        \lime{}        & 0.388 (0.21)           & 0.221 (0.19) & 0.051 (0.05)\\
        %\hline    
        \elrp{}        & \textbf{0.120} (0.01)  & 0.127 (0.11) & \textbf{0.014} (0.02)\\
        %\hline    
        \intgrad{}     & 0.126 (0.01)           & 0.148 (0.17) & 0.029 (0.04)\\
        %\hline    
        \deeplift{}    & \textbf{0.120} (0.01)  & 0.127 (0.11) & \textbf{0.014} (0.02)\\
        %\hline    
        \smoothgrad{}  & 0.135 (0.04)           & 0.153 (0.13) & 0.033 (0.05)\\
        %\hline    
        \xrai{}        & 0.151 (0.04)           & 0.144 (0.07) & 0.086 (0.11)\\
        %\hline    
        \gradcam{}     & 0.297 (0.20)           & 0.153 (0.12) & 0.139 (0.12)\\
        %\hline    
        \gradcamplus{} & 0.252 (0.13)           & 0.283 (0.24) & 0.081 (0.10)\\
        %\hline    
        \rise{}        & \textbf{0.120} (0.01)  & \textbf{0.124} (0.07) & 0.044 (0.05)\\
        \hline
    \end{tabular}
\end{minipage}
\end{table}

To investigate more the performance of the methods analyzed we computed the \textit{deletion} and the \textit{insertion} metric, discussed in Section~\ref{sec:comparison_theory}. 
For a query image, we substitute pixels in order of importance scores given by the explanation method. 
For \textit{insertion}, we blurred the image and then slowly inserted pixels while substituting with black pixels for deletion. 
For every substitution we made, we query the image to the black-box, obtaining an accuracy. 
The final score is obtained by taking the area under the curve (AUC)~\cite{hand2001simple} of accuracy as a function of the percentage of removed pixels. 
In Figure~\ref{fig:insertion_deletion_single} we have an example of this metric computed on the hockey figure of \imagenet{}. 
For every dataset, we performed this metric calculation for a set of 100 samples, and then we averaged. 
The results are shown in Table~\ref{tab:insertion_deletion_tab}.
Insertion scores decrease while augmenting the dataset image dimension because we have higher information and more pixels have to be inserted to higher the accuracy. 
On the other hand, deletion scores decrease. This fact could be because since we have greater information, it is easier to decrease accuracy. The best methods are highlighted in bold, and we can see that \rise{} is the best in three out of five experiments. 
\rise{} is followed by \deeplift{}, and \elrp{}. 
Segmentation based methods (\lime{}, \xrai{}, \gradcam{}, \gradcamplus{}) struggles when using low-resolution images.

\begin{figure}[t]
    \centering
    \includegraphics[width=.9\linewidth]{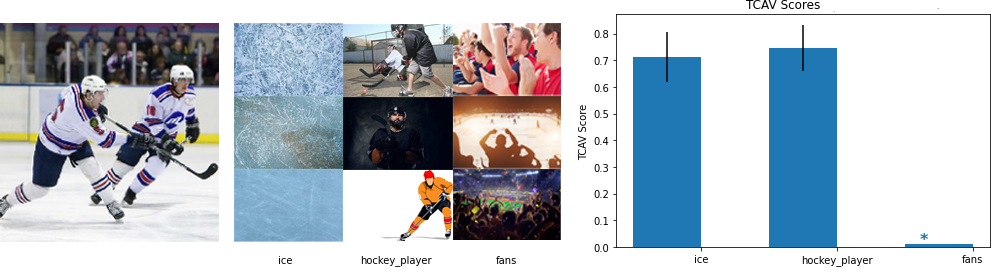}
    \caption{\tcav{} scores for three concepts: \textit{ice, Hockey player, and cheering people (fans)} for the class \textit{puck} of \imagenet{}. On the left the query image; on the center some sample of the image tested in \tcav{} as concepts, and on the right the histogram of the scores with errors. The \textit{hockey players} has been classified as a \textit{puck}, but the saliency maps are very different alongside methods. Here we can see that the \textit{ice} and the \textit{hockey players} are important concepts, while the background fans are not significant.}
    \label{fig:tcav}
\end{figure}

\subsection{Concept Attribution}
\label{sec:concept_attribution}
Most ML models are designed to operate on low-level features like edges and lines in a picture that do not correspond to high-level concepts that humans can easily understand. In \cite{adebayo2018sanity,yang2019bim}, they pointed out that feature-based explanations applied to state-of-the-art complex black-box models can yield non-sensible explanations.
Concept-based explainability constructs the explanation based on human-defined concepts rather than representing the inputs based on features and internal model (activation) states. 
This idea of high-level features is more familiar to humans, that are more likely to accept it.
For example, a low-level explanation for images is to assign to every pixel a saliency value. 
Although it is possible to look at every pixel and infer their numerical values, these make no sense to humans: we do not say that the $5^{th}$ pixel of an image has a value of $28$.
Instead CA method quantifies, for example, how much the concepts ``stripes'', has contributed to the class prediction of ``zebra''.
Formally, given a set of images belonging to a concept ${[x^{(1)},x^{(2)},...,x^{(i)}] \text{with} x^{(i)} \in C}$, CA methods can be thought as a function $f: (b,[x^{(i)}]) \rightarrow e$ which assign a score $e$ to the concept $C$ basing on the predictions and the values of the black-box $b$ on the set $[x^{(i)}]$. 

\textbf{TCAV}, Testing with Concept Activation Vectors~\cite{kim2018interpretability} is a model-agnostic method that produces post-hoc global explanations for image classifiers. 
\tcav{} provides a quantitative explanation of how important is a concept for the prediction. Every concept is represented by a particular vector called \textit{Concept Activation Vectors (CAVs)} created by interpret an internal state of a neural network in terms of human-friendly concepts.
\tcav{} uses directional derivatives to quantify the degree to which, a user-defined concept, is \textit{vital} to a classification result.

For example, how sensitive a prediction of ``zebra'' is to the presence of ``stripes''.
\tcav{} requires two main ingredients: \textit{(i)} concept-containing inputs and negative samples (random inputs), and \textit{(ii)} pre-trained ML models on which the concepts are tested. 
The concept-containing and random inputs are fed into the model to obtain the predictions to test how well a trained ML model captured a particular concept.
A linear classifier is trained to distinguish the activation of the network due to concept-containing vs. random inputs. 
The result of this training is \textit{concept activation vectors (CAVs)}. 
Once CAVs are defined, the directional derivative of the class probability along CAVs can be computed for each instance that belongs to a class. 
The ``concept importance'' for a class is computed as a fraction of the class instances that get positively activated by the concept containing inputs vs. random inputs. 
In Figure~\ref{fig:tcav}, we can see an Example of \tcav{} explanation. The user must collect some images of some concept, like ``ìce'', ``hockey player'' and ``fans''. Then \tcav{} compute the score for everyone of these, telling us which one has more impact on the prediction of a query image.

\textbf{ACE}, Automated Concept-based Explanation~\cite{ghorbani2019towards}, is the evolution of \tcav{}, and it does not need any concept example. It can automatically discover them. 
It takes training images and segments them using a segmentation method.
These super-pixels are fed into the black-box model as there where input images clustered in the activation space. 
Then we can obtain like in \tcav{} how much these clusters contributed to the prediction of a class.

\textbf{ConceptSHAP}~\cite{yeh2020completeness} defines an importance score for each concept discovered. 
Similar to \ace{}, {\scshape conceptshap} aims at having concepts consistently clustered to certain coherent spatial regions. 
{\scshape conceptshap} finds the importance of each individual concepts from a set of $m$ concept vectors $C_s = \{c_1, c_2, \dots, c_m\}$ by utilizing Shapley values. 

\textbf{CaCE}, Causal Concept Effect~\cite{goyal2019explaining}, is another variation of \tcav{}. 
It looks at the causal effect of the presence or absence of high-level concepts on the deep learning model's prediction.
\tcav{} can suffer from confounding of concepts that could happen if the training data instances have multiple classes, even with a low correlation. 
{\scshape cace} can be computed exactly if the concepts of interest are changed by intervening in the counterfactual data generation. 

\subsection{Prototypes}
\label{sec:proto_image}
Another possible explanation for images is to produce prototypical images that best represent a particular class. 
Human reasoning is often prototype-based, using representative examples as a basis for categorization and decision-making. 
Similarly, prototype explanation models use representative examples to explain and cluster data.

\begin{figure}[t]
    \centering
    \includegraphics[trim=0cm 13cm 5.9cm 0cm, clip, width=0.45\linewidth]{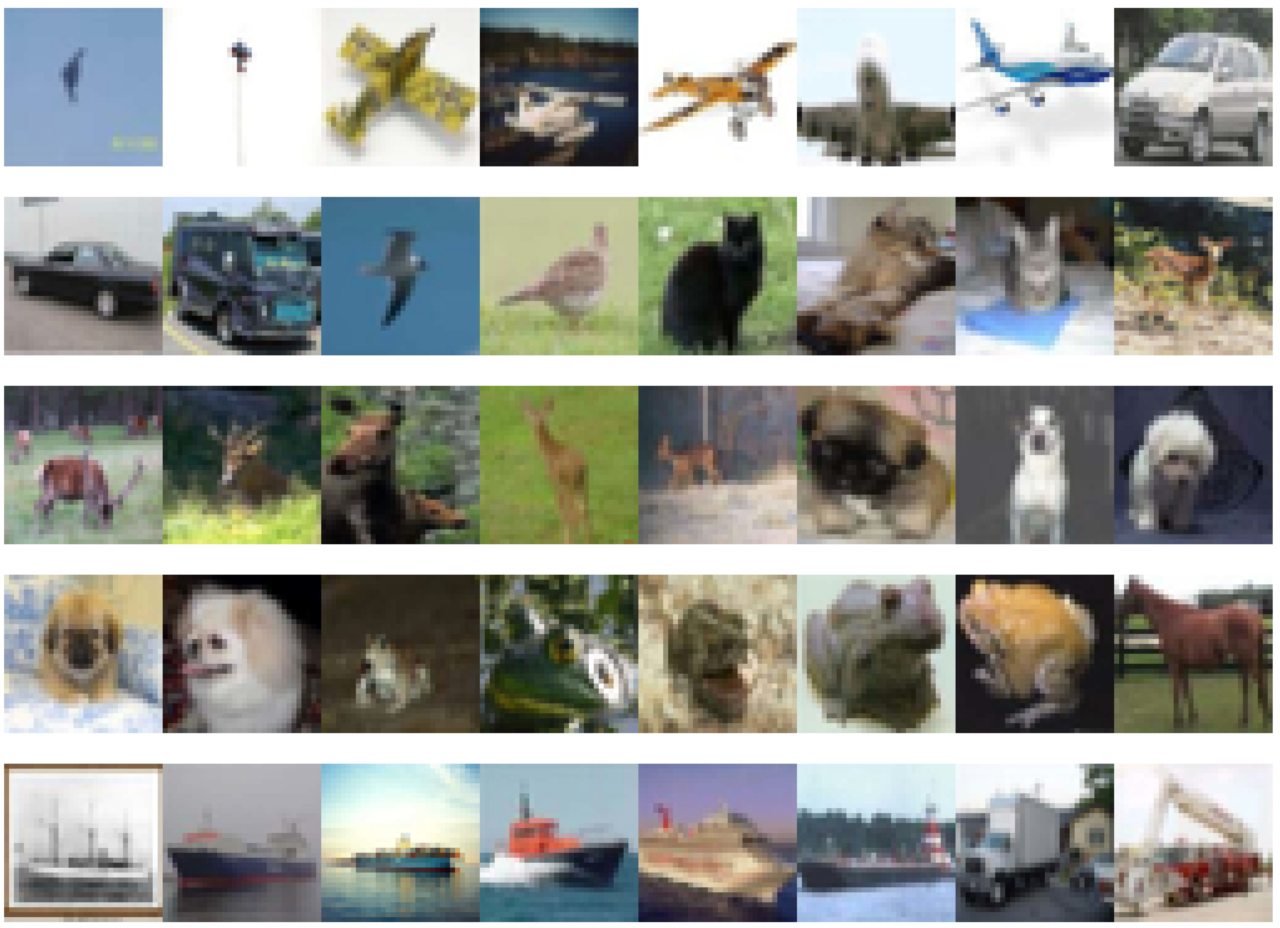}
    \includegraphics[trim=0cm 13cm 5.9cm 0cm, clip,width=0.45\linewidth]{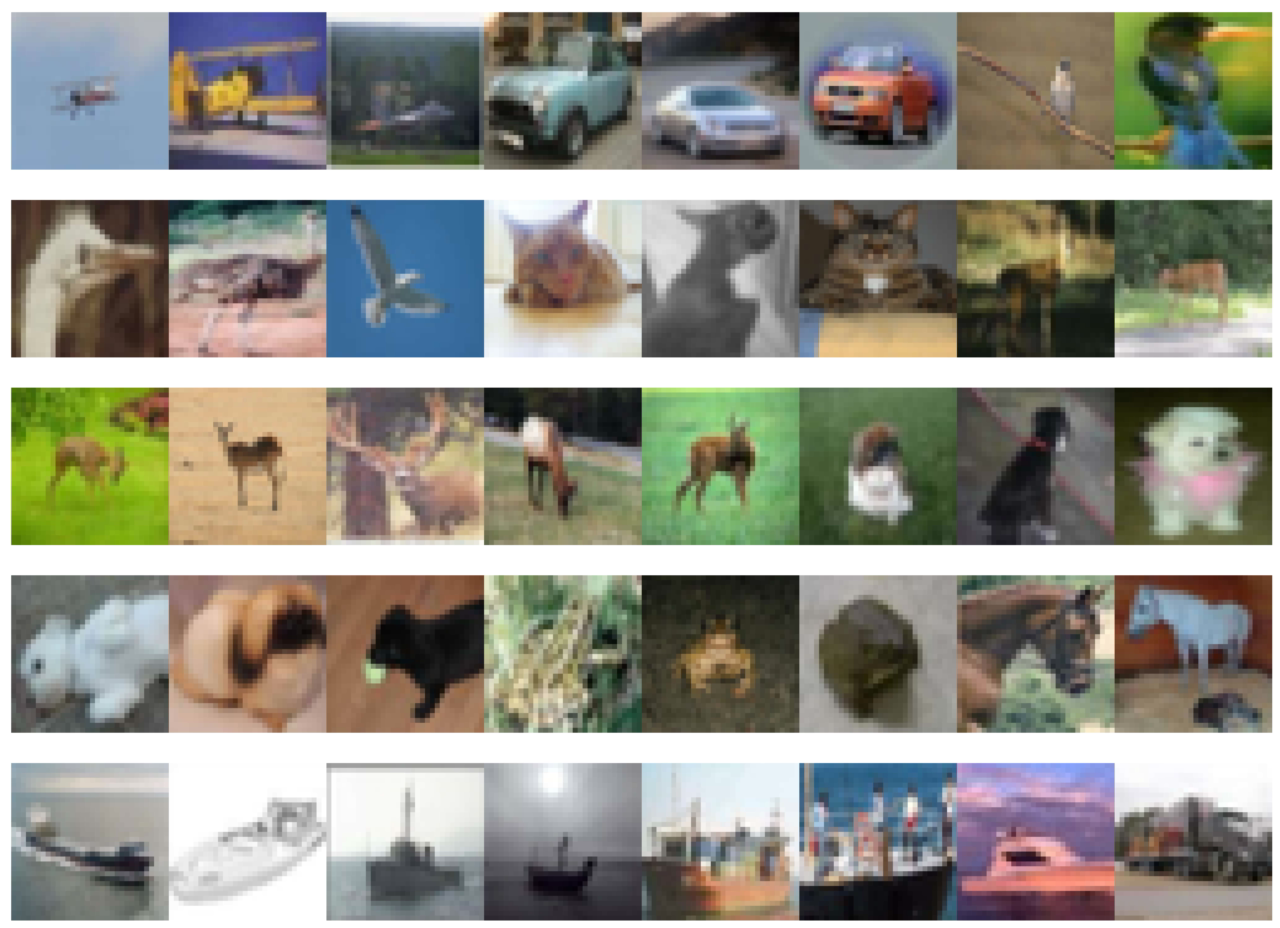}
    \caption{Criticism (on the left) and prototypes (on the right), output of \mmdcritic{} from \cifar{}. On the criticisms we have a lot of planes on white background, so the sky background is important for the plane.}
    \label{fig:mmd_critic_images}
\end{figure}  

\textbf{MMD-CRITIC}~\cite{mmdcritic}, already presented in Section~\ref{sec:tabular}, can be applied to retrieve image prototypes and criticisms. 
In Figure~\ref{fig:mmd_critic_images} is presented an application of \mmdcritic{} on \cifar{}. 
We can extract some interesting knowledge from these methods. 
For example, in the criticism images, planes are all on a white background or have a different form from the usual one. 
We can conclude that in \cifar{}, most planes are in the sky and have a passenger airplane shape.

\textbf{PROTONET}~\cite{chen2019looks} is a model-agnostic explainer that produces post-hoc global explanations on image data.
It figures out some prototypical parts of images (named prototypes) and then uses them to classify, hence making the classification process interpretable. 
A special architecture is needed to produce prototypes. 
The network learns from the training set a limited number of prototypical parts useful in classifying a new image. 
The model identifies several parts on the test image that look like some training image prototypical parts. 
Then, it makes the prediction based on a weighted combination of the similarity scores between parts of the image and the learned prototypes.
The performance is comparable to the actual state of the art but with more interpretability. 

\textbf{Influence Functions}~\cite{koh2017understanding} is another variant for building prototypes. 
Instead of building prototypical images for a class, it tries to find the most responsible images for a given prediction using influence functions. 
Influence functions is a classic technique from robust statistics to trace a model's prediction through the learning algorithm and back to its training data, thereby identifying training points most responsible for a given prediction.
Visualizing the training points most responsible for a prediction could be useful for more in-depth insights into model behavior.

\subsection{Counterfactuals}
\label{sec:counter_image}
Counterfactuals are another type of explanation for images. Its application for images is similar to the one already done for tabular data in Section \ref{sec:counter_tabular}. As output, counterfactuals methods for images produce samples of images similar to the original one but with altered prediction. Some methods output only the pixel variation, others the whole altered image.

\begin{figure}[t]
    \centering
    \includegraphics[width=\linewidth]{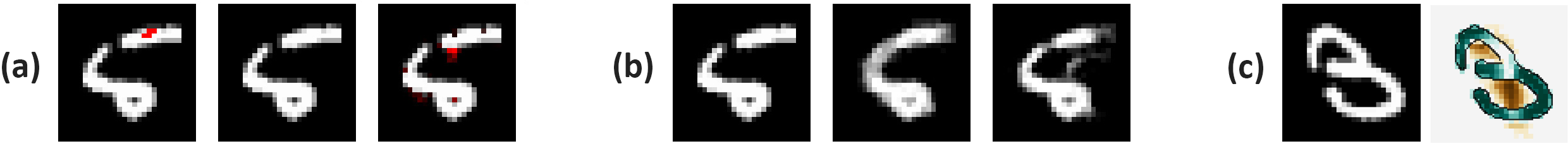}
    \caption{\textbf{(a)}:Explanation of {\scshape cem} on \mnist{}: query on the center, \textit{Pertinent Negative} left, and \textit{Pertinent Positive} right. \textbf{(b)}: Explanation of {\scshape guidedproto} on \mnist{}: left to right, the query, the closest counterfactuals labeled as 6, and 8. \textbf{(c)}: Explanation of \abele{} on \mnist{}: left query, right SM. Green/yellow areas can be exchanged without impact.}
    \label{fig:image_counterfactuals}
\end{figure}

\textbf{Guided Prototypes}, Interpretable Counterfactual Explanations Guided by Prototypes ({\scshape guidedproto})~\cite{van2019interpretable} proposes a model-agnostic method to find interpretable counterfactuals. 
{\scshape guidedproto} perturbs the input image to find the closest image to the original one but with a different classification by using an objective loss function $\mathcal{L} = cL_{pred}+\beta L_1 + L_2$ optimized using gradient descent. 
The first term, $cL_{pred}$, encourages the perturbed instance to predict another class then $x$ while the others are regularisation terms. 
In Figure~\ref{fig:image_counterfactuals} we have an example of application of {\scshape guidedproto} on \mnist{}. 
It is interesting to notice how easy it is to change the digit class with very few focused pixels.

\textbf{CEM}, Contrastive Explanation Method (\cem{})~\cite{dhurandhar2018explanations}, already presented in Section~\ref{sec:tabular}, can also be applied on image data. 
For images, Pertinent Positives (PP) or Pertinent Negatives (PN) are the pixels that lead to the same or a different class w.r.t.~the original instance. 
To create PP's and PN's, feature-wise perturbation is done by keeping the perturbations sparse and close to the original instance through an objective function that contains an elastic net $\beta L_1 + L_2$ regularizer. 
An auto-encoder is trained to reconstruct images of the training set. 
As a result, the perturbed instance lies close to the training data manifold. 
In fact, in Figure~\ref{fig:image_counterfactuals}, we can see how very few pixels are obtained as explanations on \mnist{}.

\textbf{L2X}~\cite{l2x} finds the pixels that change the classification. 
It is based on learning a function for extracting a subset of the most informative features for each given sample using Mutual Information. 
{\scshape l2x} adopts a variational approximation to efficiently compute the Mutual Information and gives a value for a group of pixels called \textit{patches}.
If the value is positive, a group contributed positively to the prediction. Otherwise, it contributed negatively.

\textbf{ABELE}, Adversarial black-box Explainer generating Latent Exemplars)~\cite{guidotti2019black}, is a local, model-agnostic explainer that produces explanations composed of: \textit{(i)} a set of exemplar and counter-exemplar images, and \textit{(ii)} a saliency map. 
The end-user can understand the classification by looking at images similar to those under analysis that received the same prediction or a different one. 
Moreover, by exploiting the SM, it is possible to understand the areas of the images that cannot be changed and varied without impacting the outcome. 
\abele{} exploits an adversarial autoencoder (AAE) to generate the record's local neighborhood to explain $x$. 
It builds the neighborhood on a latent local decision tree, which mimics the behavior of $b$. 
Finally, exemplars and counter-exemplars are selected, exploiting the rules extracted from the decision tree. 
The SM is obtained by a pixel-by-pixel difference between $x$ and the exemplars. 
In Figure~\ref{fig:image_counterfactuals} we have an example of application of \abele{} on \mnist{}. Green and yellow areas can change without impacting the black-box outcome, while the gray areas must remain the same to have the same prediction. 

\begin{table}[t]
    \centering
    \caption{Explanation runtime expressed in seconds for explainers of image classifiers approximated as order of magnitude.}
    \label{tab:runtime_image}
    \footnotesize
    %\scriptsize
    \begin{tabular}{|c|c|c|c|c|c|c|c|c|c|c|c|c|c|c|c|}
        \hline
        \cellcolor[gray]{0.95} \rotatebox[origin=c]{90}{\textbf{Dataset}} 
        & \cellcolor[gray]{0.95} \rotatebox[origin=c]{90}{\textbf{Black-Box}} 
        & \cellcolor[gray]{0.95} \rotatebox[origin=c]{90}{\lime{}} 
        & \cellcolor[gray]{0.95} \rotatebox[origin=c]{90}{\elrp{}} 
        & \cellcolor[gray]{0.95} \rotatebox[origin=c]{90}{\intgrad{}} 
        & \cellcolor[gray]{0.95} \rotatebox[origin=c]{90}{\deeplift{}} 
        & \cellcolor[gray]{0.95} \rotatebox[origin=c]{90}{\smoothgrad{}} 
        & \cellcolor[gray]{0.95} \rotatebox[origin=c]{90}{\xrai{}}
        & \cellcolor[gray]{0.95} \rotatebox[origin=c]{90}{\gradcam{}}
        & \cellcolor[gray]{0.95} \rotatebox[origin=c]{90}{\gradcamplus{}}
        & \cellcolor[gray]{0.95} \rotatebox[origin=c]{90}{\rise{}}
        & \cellcolor[gray]{0.95} \rotatebox[origin=c]{90}{\tcav{}}
        & \cellcolor[gray]{0.95} \rotatebox[origin=c]{90}{\mmdcritic{}}
        & \cellcolor[gray]{0.95} \rotatebox[origin=c]{90}{\cem{}}
        & \cellcolor[gray]{0.95} \rotatebox[origin=c]{90}{{\scshape guidedprop}}
        & \cellcolor[gray]{0.95} \rotatebox[origin=c]{90}{\abele{}}\\
         \hline
                                                                     % lime|lrp|intgrad|dlift|smooth|xrai |gcam |gcam++|rise |tcav |mmd  |cem |guid|abele
        \multicolumn{1}{|c|}{\multirow{1}{*}{\mnist{}}}      & CNN   & 1  & 1  & 0.03  & 2   & 0.04 & 1  & 0.1  & 0.1  & 0.5 & -  & -   & 580 & 11 & 2000\\
        \multicolumn{1}{|c|}{\multirow{1}{*}{\cifar{}}}      & CNN   & 10 & 1  & 0.06  & 1   & 0.07 & 1.5& 0.15 & 0.15 & 2   & -  & 277 & -   & -  & 1800\\
        \multicolumn{1}{|c|}{\multirow{1}{*}{\imagenet{}{}}} & VGG16 & 50 & 2  & 5     & 3   & 0.8  & 18 & 0.25 & 0.25 & 21  & 300& -   & -   & -  & -\\
         \hline
    \end{tabular}
\end{table}

\subsubsection*{\textbf{Runtime Analysis}}
Table~\ref{tab:runtime_image} shows the explanation runtime approximated as order of magnitude.
We notice that \gradcam{} and \gradcamplus{} are the fastest methods, especially for big models like the VGG network. In general, pixel-wise Saliency explanations are more comfortable to obtain, while segmentation slows a lot, especially for high-resolution images. CA, CF, and PR methods are very slow compared to the SM. This problem is because these algorithms need additional training or use some searching algorithm.

\subsubsection*{\textbf{Discussion}}
When dealing with images, the most diffused explanations are Saliency Maps (Section~\ref{sec:saliency}). The literature presents a multitude of methods that are capable of producing such type of explanation. 
The problem with saliency maps is the confirmation bias~\cite{adebayo2018sanity}. Also, humans do not think in terms of pixels. The explanation of Saliency Maps is provided in terms of pixels, which are low level features that are useful only for an expert user who wants to check the robustness of the black-box. For a general audience, there is the need to build an explanation in terms of higher features called concepts. This is the goal of Concept Attributions based explanations (Section~\ref{sec:concept_attribution}). For a concept selected by a human team, these types of methods compute a score that evaluates the probability that the selected concept has influenced the prediction. Concept based explanations are a very recent type of explanation for images, and they have potential improvements. It is a first step in the direction of human-like explanations. Human-friendly concepts make it possible to build straightforward and useful explanations. Humans still need to map images to concepts, but it is a small price to pay to augment the human-machine interaction.
Other approaches are based on the concept of producing examples to support the explanation. Prototypes and Counterfactual (Sections~\ref{sec:proto_image} and~\ref{sec:counter_image}) are two types of similar explanations but with very different meaning. The goal of prototypes is to produce an example that reflects the common proprieties of a class, while the goal of counterfactual is to produce examples similar to the input, but with a different predicted class. The first one is useful for model inspection, while the ladder for the user experience. In particular, counterfactuals are more user-friendly since they highlight the changes to make to obtain the desired prediction. 

%% file: 6_text.tex
\begin{table}[t]
    \small
    \centering
    \setlength{\tabcolsep}{1mm}
    \caption{Summary of methods for opening and explaining black-boxes. }
  \begin{tabular}{cccccccccc}
    \hline
    \rotatebox[origin=c]{0}{\textbf{Type}}  & \rotatebox[origin=c]{0}{\textbf{Name}}    & \rotatebox[origin=c]{0}{\textbf{Ref.}}  & \rotatebox[origin=c]{0}{\textbf{Authors}} & \rotatebox[origin=c]{0}{\textbf{Year}} & \rotatebox[origin=c]{0}{\textbf{Data Type}}  &
    \rotatebox[origin=c]{0}{\textbf{IN/PH}}  & \rotatebox[origin=c]{0}{\textbf{G/L}}    &
    \rotatebox[origin=c]{0}{\textbf{A/S}}  &
    \rotatebox[origin=c]{0}{\textbf{Code}} \\
    
    \hline
    \cellcolor{white}       & \lime{}                   & \cite{ribeiro2016should} 		    & Ribeiro et al. 	    & 2016  & ANY & PH & L & A & \href{https://github.com/marcotcr/lime}{link} \\
    \rowcolor{gray!15}
    \cellcolor{white}       & \intgrad{}                & \cite{sundararajan2017axiomatic} 	& Sundararajan et al.   & 2017  & ANY & PH & L & S & \href{github.com/marcoancona/DeepExplain}{link}\\
    \cellcolor{white}
                            & {\scshape l2x} 		    & \cite{l2x} 				        & Chen et al. 		    & 2018  & ANY & PH & L & A & \href{https://github.com/Jianbo-Lab/L2X}{link} \\
    \rowcolor{gray!15}
    \cellcolor{white}       & \deeplift{}               & \cite{shrikumar2017learning} 		& Shrikumar et al. 	    & 2017  & ANY & PH & L & S & \href{https://github.com/marcoancona/DeepExplain}{link} \\
    \multirow{-4}{*}{SH} 
    \cellcolor{white}       & {\scshape lionets}        & \cite{LioNets} 		& Mollas et al. 	    & 2019  & ANY & PH & L & S & \href{https://github.com/iamollas/LioNets}{link} \\
    \rowcolor{gray!15}
    \cellcolor{gray!15}     & -                         & \cite{li2016understanding}		& Li et al.	            & 2014  & TXT & PH & L & S & - \\
    \cellcolor{gray!15}     & {\scshape exbert}         & \cite{hoover2019exbert}           & Hoover et al.         & 2019  & TXT & PH & L & S & \href{https://exbert.net/}{link}\\
    \rowcolor{gray!15}
    \multirow{-3}{*}{AB}
    \cellcolor{gray!15}     & - & \cite{vaswani2017attention}		& Vaswani et al.	    & 2017  & TXT & PH & L & S & - \\
    \cellcolor{white}     & \anchor{}                 & \cite{ribeiro2018anchors}		    & Ribeiro et al.	    & 2018  & TXT & PH & L & A & \href{https://github.com/marcotcr/anchor}{link} \\
    \rowcolor{gray!15}
    \cellcolor{white}     & {\scshape quint}          & \cite{abujabal2017quint}	    	& Abujabal et al.	    & 2017  & TXT & PH & L & S & - \\
    
    \cellcolor{white}     & {\scshape criage}         & \cite{pezeshkpour2019investigating}& Pezeshkpour et al.   & 2019  & TXT & PH & L & S & \href{https://github.com/pouyapez/criage}{link} \\
    \rowcolor{gray!15}
    \cellcolor{white}     & {\scshape lasts}          & \cite{guidotti2020lasts}	    	& Guidotti et al.	        & 2020  & TXT & PH & L & S & - \\
    
    \cellcolor{white}     & {\scshape xspells}        & \cite{lampridis2020explaining}	& Lampridis et al.	    & 2020  & TXT & PH & L & S & \href{https://github.com/orestislampridis/X-SPELLS}{link}\\
    \rowcolor{gray!15}
    \cellcolor{white}     & {\scshape -}              & \cite{rajani2019explain}	    	& Rajani et al.	        & 2019  & TXT & PH & L & S & - \\
    \multirow{-7}{*}{Other}
    \cellcolor{white}     & {\scshape doctorxai}      & \cite{panigutti2020doctor}	    	& Panigutti et al.	    & 2020  & ANY & PH & L & S & - \\
    
    \hline
    \end{tabular}%
   % \end{varwidth}
    \label{tab:text_summary}
\end{table}

\section{Text}
\label{sec:text}
For text data, we can distinguish the following types of explanations: 
\textit{Saliency Maps (SM)}, described in Section~\ref{sec:Sentence_highlighting}, \textit{Attention-Based methods (AB)}, described in Section~\ref{sec:Attention_Based}, \textit{Other Methods}, detailed in Section~\ref{sec:text_others}.
Additional details available \cite{danilevsky2020survey}.
Table~\ref{tab:text_summary} summarizes the explanation methods acting on text data.
Text, unlike tabular and image data, does not have a structure.
The variety and complexity of tasks related to text are enormous and in literature is known as \textit{Natural Language Processing (NLP)}~\cite{chowdhary2020natural}. 
In the following, we analyze text classification in detail because, among information retrieval, machine translation, and question answering.
Text classification is the main topic where XAI methods exist in literature. 
Examples of usage in text classification are sentiment analysis, topic labeling, and spam.
Text classification is the process of assigning tags or categories to text according to its content. 
Using labeled examples as training data, a ML model can learn the different associations between pieces of text and a particular output called tags. 
Tags can be thought of as labels which distinguish different type of text. 
For sentiment analysis, it is possible to have tags as positive, negative, or neutral.
XAI techniques are generally applied to understand what words are the most relevant for a specific tag assignment.
We experimented on three datasets: \sst{}, \imdb{}, and \yelp{}. 
We selected these datasets\footnote{\sst{}: \url{https://nlp.stanford.edu/sentiment/index.html}, \imdb{}: \url{https://ai.stanford.edu/~amaas/data/sentiment/}, \yelp{}: \url{https://www.kaggle.com/yelp-dataset/yelp-dataset}}, because they are the most used on sentiment classification and have different dimensions. 
On these datasets we trained different black-box models.  
For every explainer we present an example of an application on one or more datasets.

\begin{figure}[t]
    \centering
    \includegraphics[width=0.9\linewidth]{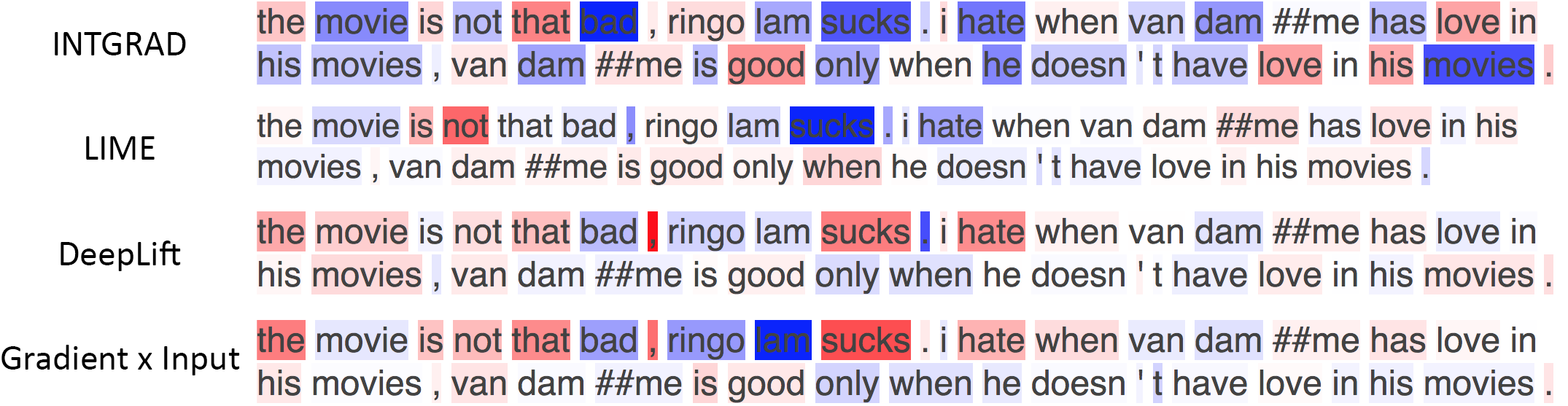}
    \caption{Example of sentence highlighting, on top we have the score produce by IntGrad and below we have in order, LIME, DeepLift and the baseline which consists of multiplying the input with the gradient w.r.t.~input. The sentence is taken from \imdb{}}
    \label{fig:sentence_highlighting}
\end{figure}

\subsection{Sentence Highlighting} 
\label{sec:Sentence_highlighting}
As seen in Section~\ref{sec:saliency}, saliency-based explanations are prevalent because they present visually perceptive explanations. 
\textit{Saliency highlighting} is saliency maps applied to text and consists of assigning to every word a score based on the importance that that word had in the final prediction. 
Formally, a Sentence Highlighting (SH) is modeled as a vector $s$ who explain a classification $y = b(x)$ of a black-box $b$ on $x$. 
The dimensions of $s$ are the words present in the sentence $x$ we want to explain, and the value $s_{i}$ is the saliency value of the word $i$. 
The greater the value of $s_{i}$ the bigger is the importance of that word. 
A positive value indicates a positive contribution towards $y$, while a negative one means that the word has contributed negatively. 
Some examples are reported in Figure~\ref{fig:sentence_highlighting}.
To obtain such an explanation, it is possible to adapt some of the saliency maps methods presented in Section~\ref{sec:saliency}.

\textbf{LIME}~\cite{ribeiro2016should}, presented in Section~\ref{sec:tabular}, can be applied to text with a modification to the perturbation of the original input. 
Given an input sentence $x$, \lime{} creates a neighborhood of sentences by replacing one or multiple words with spaces. 
Another possible variation is to insert a similar word instead of removing them.% to maintain the meaning of the sentence.

\textbf{INTGRAD}~\cite{sundararajan2017axiomatic}, presented in Section~\ref{sec:tabular}, can also be exploited to explain text classifiers.
Indeed, gradient-based methods are challenging to apply to NLP models because the vector representing every word is usually averaged into a single sentence vector. 
Since it does not exist a mean operation gradient, the explainer cannot redistribute the signal back to the original vectors.
On the other hand, \intgrad{} is immune to this problem because the saliency values are computed as a difference with a baseline value. 
\intgrad{} computes the saliency value of a single word as a difference from the sentence without it. 
For a fair comparison, we substituted the words with spaces as done for \lime{}.% also for \intgrad{}.

\textbf{DEEPLIFT}~\cite{shrikumar2017learning}, presented in Section~\ref{sec:tabular}, can also be applied on text following the same principle of \intgrad{}.
For the experiments, we adopt the same preprocessing used for \lime{} and \intgrad{}.

\textbf{L2X}~\cite{l2x} can produce a SH explanation for text. In particular, for text, the patches are now a group of words. 

\begin{table}[t]
    \centering
    \caption{Deletion (right) and Insertion (left) metrics and computed on Sentence Highlighting for different datasets.}
    \label{tab:deletion_insertion_text}
    \footnotesize
    \begin{minipage}{.49\linewidth}
        \centering
        \begin{tabular}{|c|c|c|c|}
            \hline
            \rowcolor{gray!15}
             & \sst{} &  \imdb{} & \yelp{}\\
            \hline
            \intgrad{}  & 0.6447 (0.21) & 0.647 (0.21) & 0.7595 (0.25)\\
            %\hline
            \lime{}     & 0.6199 (0.23) & 0.648 (0.21) & 0.7712 (0.25)\\
            %\hline
            \deeplift{} & 0.6297 (0.23) & 0.600 (0.15) & 0.7565 (0.31)\\
            %\hline
            %\shortstack[c]{{\scshape gradient} \\ {\scshape x input}} & 0.6287 (0.23) & 0.630 (0.16) & 0.7590 (0.28)\\
            \makecell{{\scshape gradient} \\ {\scshape x input}} & 0.6287 (0.23) & 0.630 (0.16) & 0.7590 (0.28)\\
            \hline
        \end{tabular}
    \end{minipage}
    \begin{minipage}{.49\linewidth}
        \centering
        \begin{tabular}{|c|c|c|c|}
            \hline
            \rowcolor{gray!15}
             & \sst{} &  \imdb{} & \yelp{}\\
            \hline
            \intgrad{}  & 0.6107 (0.23) & 0.616 (0.16) & 0.7625 (0.33)\\
            %\hline
            \lime{}     & 0.6337 (0.23) & 0.599 (0.17) & 0.7513 (0.33)\\
            %\hline
            \deeplift{} & 0.6137 (0.21) & 0.645 (0.16) & 0.7524 (0.30)\\
            %\hline
            \makecell{{\scshape gradient} \\ {\scshape x input}} & 0.5852 (0.22) & 0.632 (0.16) & 0.7479 (0.31)\\
            \hline
        \end{tabular}
    \end{minipage}
\end{table}

\subsubsection*{\textbf{Qualitative and Quantitative Comparison of Sentence Highlighting}}
Besides the methods exposed above we tested also a baseline method. 
This baseline named Gradient $\times$ Input takes the black-box gradient of the input w.r.t to the output and multiply these value by the input values. 
The results are shown in Figure~\ref{fig:sentence_highlighting}. 
The highlighted words are very different among the various methods. 
\intgrad{} and \lime{} are the ones who output meaningful explanations, while \deeplift{} struggles a lot to diversify from the baseline. 
We also measured the \textit{deletion/insertion} and report the results in Table~\ref{tab:deletion_insertion_text}. 
For both metrics, we have very poor performance among all the methods. 
However removing a single word barely changes the meaning of the sentence.

\subsection{Attention-based Methods}
\label{sec:Attention_Based}
\textit{Attention} was proposed in~\cite{xu2015show} to improve the model performance.
The authors managed to show through an attention layer which parts of the images contributed most to realize the caption.
Attention is a layer to put on top of the model that, for each pixel, $ij$ of the image $x$, generates a positive weight $\alpha_{ij}$, i.e.,~the \textit{attention} weight. 
This value can be interpreted as the probability that a pixel $ij$ is in the right place to focus on producing the next word in the caption. 
Attention mechanisms allow models to look over all the information the original sentence holds and learn the context~\cite{wu2020context,bahdanau2014neural}.
Therefore, it has caught the interest of XAI researchers who started using these weights as an explanation. 
The explanation $e$ of the instance $x$ is composed by the set of attention values ($\alpha$), one for each feature $x_i$. 
Attention is nowadays a delicate argument, and while it is clear that it augments the performance of models, it is less clear if it helps gain interpretability and what are the relationship with model outputs~\cite{jain2019attention}.

\textbf{Attention Based Sentence Highlighting}~\cite{li2016understanding} is an AB mechanism to produce a heatmap explanation similar to the one used for SMs.
The scores are computed for every word of the sentence by using the attention layer of the black-box. 
The weights $\alpha_{ij}$ of the attention layer are used as a score. 
The higher the score, the redder highlighting.

\begin{figure}[t]
    \centering
    \begin{minipage}{0.48\linewidth}
        \includegraphics[width=\linewidth]{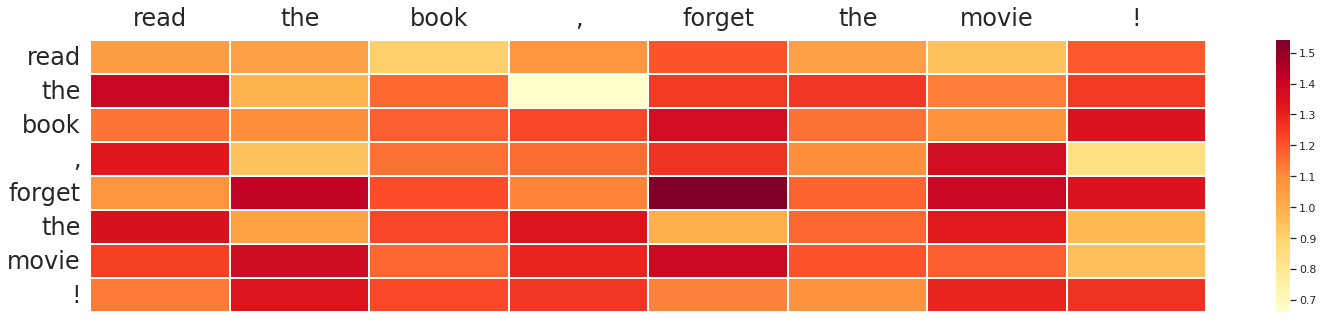}
        \caption{Saliency heat-map matrix generated from the method presented in~\cite{cheng2016long}. The row and the columns of the matrix correspond to the words in the sentence `Read the book, forget the movie!''. Each value of the matrix shows the attention weight $\alpha_{ij}$ of the annotation of the $i$-th word w.r.t.~the $j$-th.}
        \label{fig:attention_matrix}
    \end{minipage}
    \quad
    \begin{minipage}{0.48\linewidth}
        \includegraphics[trim={0 0 0 3.45cm},clip,width=\linewidth]{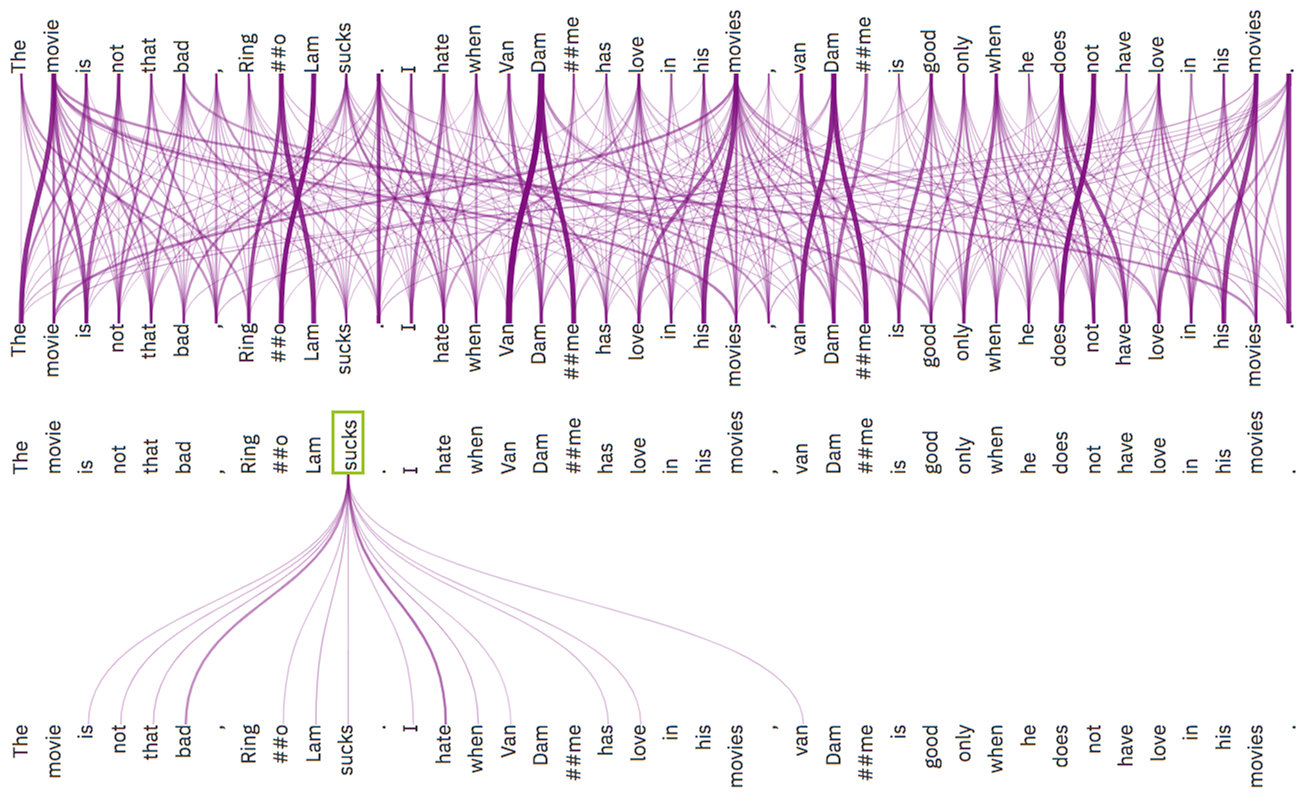}    
        \caption{Representation of the attention in BERT for a sentence taken from \imdb{} using the visualization of \cite{hoover2019exbert}. The greater the attention between two words, the bigger the line. Here is selected only the attention related to the word ``sucks''.}
        \label{fig:self_attention}
    \end{minipage}
\end{figure}

\textbf{Attention Matrix}~\cite{cheng2016long} looks at the dependencies between words for producing explanations.
It is a self-attention method, sometimes called \textit{intra-attention}. 
{\scshape attentionmatrix} relates different positions of a single sequence to compute its internal representation. 
The attention of a sentence $x$ composed of $N$ words can be understood as an $N \times N$ matrix, where each row and columns represent a word in the input sentence. 
The values of the matrix are the attention values of every possible combination of the tokens. 
This matrix is a representation of values pointing from each word to every other word~\cite{vaswani2017attention} (see Figure~\ref{fig:attention_matrix}). 
We can also visualize this matrix with a focus on the connection between words~\cite{hoover2019exbert} as in Figure~\ref{fig:self_attention}, where the thickness of the lines is the self-attention value between two tokens.

\subsubsection*{\textbf{Runtime Analysis}}
NLP models are usually very large resulting in poor performance in terms of runtime.
Apart from Attention Matrix methods which are instant, we notice that for all the datasets, the time are pretty much the same in the order of magnitude of ten seconds. 
The time of the methods is independent from the dataset size. 

\subsubsection*{\textbf{Discussion}}
Explanations of text data are at the very early stages compared to tabular data and images. The majority of the methods focus on low feature explanation by giving a score to words that make up the sentence. As said for image type of explanations in Section~\ref{sec:image}, these low feature explanations are useful to check the model's robustness, not to give a useful explanation for the final inexpert user. Natural Language processing is a very complex field, and find a human-friendly explanation is challenging. Researchers are working in the direction of creating explanation with high concept~\cite{srivastava2017joint}, and using humans to augment these type of concept~\cite{rajani2019explain}, as done for Concept Attribution.

\subsection{Miscellanea of Other Methods}
\label{sec:text_others}
There other methods that are important to mention when we talking about XAI using text or sequential data. 

\textbf{ANCHOR}, presented in Section~\ref{sec:tabular}, can be adapted to text by using as perturbation the word UNK. 
It consists of perturbing a sentence by substituting words with UNK (unknown). 
For example, It shows how ``sucks'' contributed to the negative prediction of the sentence, but when coupled with ``love'' then the sentence prediction switches to positive.

\textbf{Natural Language Explanation} verbalizes explanations in natural human language. 
Natural language can be generated with complex \textit{deep learning models}, e.g.,~by training a model with natural language explanations and coupling with a generative model~\cite{rajani2019explain}. 
Besides, it can also be generated using a simple \textit{template-based approach}~\cite{abujabal2017quint}. 

\textbf{XSPELLS}~\cite{lampridis2020explaining} is a model-agnostic explainer returning exemplars and counterexamples sentences as explanation.
It re-implements {\scshape abele} for text data by using LSTM layers in the autoencoder. 
Exemplars and counterexemplars are selected, exploiting the rules extracted from the decision tree learned in the latent space. 

\textbf{LASTS}, Local Agnostic Shapelet-based Time Series explainer ({\scshape lasts})~\cite{guidotti2020lasts}, is a variation of {\scshape abele} for time series. 
Since a text could be interpreted as a time series, we report here this work. 
As explanation {\scshape lasts} returns exemplars and counterexamples time series and shapelet-based rules. 
Shapelets are locally discriminative subsequences characterizing the classification.
An example of a rule is: ``If these shapelets are present and these others not, then $x$ is classified as $y$''.

\textbf{DOCTORXAI}~\cite{panigutti2020doctor} is a local post-hoc model-agnostic explainer acting on sequential data in the medical setting.
In particular, it exploits a medical ontology to perturb the data and to generate neighbors.
{\scshape doctorxai} is designed on healthcare data, but it can theoretically be applied to every type of sequential data with an ontology.

%% file: 7_Toolboxes.tex
\section{Explanation Toolboxes}
\label{expl-tools}
A significant number of toolboxes for the ML explanation have been proposed during the last few years. 
In the following, we report the most popular Python toolkits with a brief description of the explanation models they provide\footnote{
    AIX360: \url{https://github.com/Trusted-AI/AIX360},
    CaptumAI: \url{https://captum.ai/},
    InterpretML: \url{https://github.com/interpretml/interpret},
    Alibi~\url{https://github.com/SeldonIO/alibi},
    FAT-Forensics: \url{https://github.com/fat-forensics/fat-forensics},
    What-If Tool: \url{https://github.com/pair-code/what-if-tool}.
}. 

%\textbf{AIX360}\footnote{\url{https://github.com/Trusted-AI/AIX360}}
\textbf{AIX360}~\cite{aix360-sept-2019} contains both intrinsic, post-hoc, local, and global explainer and it can be used with every kind of input dataset. 
Regarding the local post-hoc explanations, different methods are implemented, such as \lime{}~\cite{ribeiro2016should}, \shap{}~\cite{lundberg2017unified}, \cem{}~\cite{dhurandhar2018explanations}, {\scshape cem-maf}~\cite{luss2019generating} and {\scshape protodash}~\cite{gurumoorthy2019efficient}). 
Another interesting method proposed in this toolkit is {\scshape ted}~\cite{hind2019ted,dash2018boolean}, which provides intrinsic local explanations and provides global explanations based on rules. 
%\textbf{CaptumAI}\footnote{https://captum.ai/} 
\textbf{CaptumAI} is a library built for \texttt{PyTorch} models.
%It allows both for model explanation and inspection. 
CaptumAI divides the available algorithms into three categories: 
\textit{Primary Attribution}, in which there are methods able to evaluate the contribution of each input feature to the output of a model: \intgrad{}~\cite{sundararajan2017axiomatic}, \gradshap{}~\cite{lundberg2017unified}, \deeplift{}~\cite{shrikumar2017learning}, \lime{}~\cite{ribeiro2016should}, \gradcam{}~\cite{selvaraju2017grad}.
\textit{Layer Attribution}, in which the focus is on the contribution of each neuron: e.g.~\gradcam{}~\cite{selvaraju2017grad} and {\scshape layer-\deeplift{}~\cite{shrikumar2017learning}}.
\textit{Neuron Attribution}, in which is analyzed the contribution of each input feature on the activation of a particular hidden neuron: e.g.~{\scshape neuron-\intgrad{}~\cite{sundararajan2017axiomatic}}, {\scshape neuron-\gradshap{}~\cite{lundberg2017unified}}.
%\textbf{InterpretML}\footnote{\url{https://github.com/interpretml/interpret}} 
\textbf{InterpretML}~\cite{nori2019interpretml} contains intrinsic and post-hoc methods for Python and R. 
InterpretML is particularly interesting due to the \textit{intrinsic} methods it provides: Explainable Boosting Machine ({\scshape ebm}), Decision Tree, and Decision Rule List.
These methods offer a user-friendly visualization of the explanations, with several local and global charts. 
InterpretML also contains the most popular methods, such as \lime{} and \shap{}. 
\textbf{DALEX}~\cite{biecekexplanatory} is an R and Python package that provides post-hoc and model-agnostic explainers that allow local and global explanations. It is tailored for tabular data and is able to produce different kinds of visualization plots.
%\textbf{Alibi}~\footnote{\url{https://github.com/SeldonIO/alibi}} 
\textbf{Alibi} provides intrinsic and post-hoc models. 
It can be used with \textit{any type of input dataset} and \textit{both for classification and regression} tasks.
Alibi provides a set of counterfactual explanations, such as \cem{}, and, interestingly, an implementation of \anchor{}~\cite{ribeiro2018anchors}. 
Regarding global explanation methods, Alibi contains {\scshape ale} (Accumulated Local Effects)~\cite{apley2016visualizing}, which is a method based on partial dependence plots~\cite{guidotti2018survey}.
%\textbf{FAT-Forensics}\footnote{\url{https://github.com/fat-forensics/fat-forensics}} 
\textbf{FAT-Forensics} takes into account \textit{fairness, accountability and transparency}. 
Regarding intrinsic explainability, it provides methods to assess explainability under three perspectives: data, models, and predictions.
For \textit{accountability}, it offers a set of techniques that \textit{assesses privacy, security, and robustness}. 
For \textit{fairness}, it contains methods for \textit{bias detection}.
%\textbf{What-If Tool}\footnote{\url{https://github.com/pair-code/what-if-tool}}
\textbf{What-If Tool} is a toolkit providing a visual interface from which it is possible to play without coding. 
Moreover, it can work directly with ML models built on \textit{Cloud AI Platform} (\url{https://cloud.google.com/ai-platform}).
It contains a variety of approaches to get feature attribution values such as \shap{}~\cite{lundberg2017unified}, \intgrad{}~\cite{sundararajan2017axiomatic}, and \smoothgrad{}~\cite{selvaraju2017grad}.

%% file: 8_conclusion.tex
\section{Conclusion}
\label{sec:conclusion}
This paper has presented a survey of the last advances on XAI methods, following a categorization based on the data types and explanation strategies. 
We measured and evaluated a set of benchmarks for each explanation technique for a comparison from both the quantitative and qualitative point of view. 

Our literature review revealed interesting trends in the strategies proposed for an explanation. For tabular data, feature importance is the most widely adopted strategy, particularly for Explainable-by-Design solutions and model agnostic black box explanations. Rule-based explanations are gaining attention since their logic formalization enables a deeper understanding of the AI model's internal decisions. Recently, methods that explain in terms of counterfactuals are yielding interesting results.
For image data, the most considerable adopted technique is based on the creation of Saliency Maps, which translate to the image domain the feature relevance approach for tabular data, highlighting the portions of the relevant images for the AI model outcome. However, other approaches, like Concept Attribution, Prototypes, and Counterfactual, are rising in recent years.
The explanation techniques are still limited for text data, but it is still possible to highlight a few trends. We recall the Sentence Highlight that, similarly to feature importance for tabular data, provides a weight to the portion of the input that contributed, positively or negatively, to the outcome.
Across the different data types, different approaches tend to use similar strategies. This is also evident if we look at the internals of these algorithms. For example, several methods exploit the generation of a synthetic neighborhood around an instance to reconstruct the local distribution of data around the point to investigate. This stochastic generation is the base of several methods, and it also explains the low performance on the stability measure (see Table~\ref{tab:fidelity-faith}). Another frequent strategy consists of learning a surrogate model from partial training data (sometimes created from the neighborhood generation). This approach tries to bring the benefit of intrinsic methods in the context of black box explanation.

In recent years the contributions on the Explainable AI topics are constantly growing, particularly in AI and ML. However, there are still a restricted number of contributions focusing on the comparison of these methods. A definition of a unifying metric for measuring the efficacy of explanation strategies is difficult, particularly when human-grounded evaluations are addressed. We believe that next year of research will focus more on the human side, emphasizing the human-machine interactions and aligning the generation of the explanation with the cognitive model of the final user. Some preliminary results of this direction are presented in~\cite{guidotti2020evaluating,jeyakumar2020can,hase2020evaluating}.
We believe that XAI must be addressed more in the development of AI applications in the future, and we hope that this work could help in its development.